\documentclass[journal]{IEEEtai}

\usepackage[colorlinks,urlcolor=blue,linkcolor=blue,citecolor=blue]{hyperref}
\usepackage{stfloats}
\usepackage{array}
\usepackage{color,array}
\usepackage[numbers,sort&compress]{natbib}
\usepackage{graphicx}
\usepackage{amsmath,amsfonts}

\begin{document}

\title{IGDNet: Zero-Shot Robust Underexposed Image Enhancement via Illumination-Guided and Denoising} 

\author{Hailong Yan, Junjian Huang, and Tingwen Huang, \IEEEmembership{Fellow, IEEE}
}
\markboth{Journal of IEEE Transactions on Artificial Intelligence, Vol. 00, No. 0, Month 2020}
{First A. Author \MakeLowercase{\textit{et al.}}: Bare Demo of IEEEtai.cls for IEEE Journals of IEEE Transactions on Artificial Intelligence}

\maketitle

\begin{abstract}
Current methods for restoring underexposed images typically rely on supervised learning with paired underexposed and well-illuminated images. However, collecting such datasets is often impractical in real-world scenarios. Moreover, these methods can lead to over-enhancement, distorting well-illuminated regions. To address these issues, we propose IGDNet, a Zero-Shot enhancement method that operates solely on a single test image, without requiring guiding priors or training data. IGDNet exhibits strong generalization ability and effectively suppresses noise while restoring illumination. The framework comprises a decomposition module and a denoising module. The former separates the image into illumination and reflection components via a dense connection network, while the latter enhances non-uniformly illuminated regions using an illumination-guided pixel adaptive correction method. A noise pair is generated through downsampling and refined iteratively to produce the final result. Extensive experiments on four public datasets demonstrate that IGDNet significantly improves visual quality under complex lighting conditions. Quantitative results on metrics like PSNR (20.41dB) and SSIM (0.860dB) show that it outperforms 14 state-of-the-art unsupervised methods. The code will be released soon.
\end{abstract}

\begin{IEEEImpStatement}
This paper presents IGDNet, a zero-shot approach for enhancing underexposed images, specifically designed to overcome the limitations of existing supervised and unsupervised methods. Unlike previous techniques that rely on extensive paired datasets or normal-light references, IGDNet requires no training data, making it highly adaptable to complex real-world low-light conditions. It incorporates a decomposition-based architecture and illumination-guided pixel correction to effectively restore image details and suppress noise. Extensive experiments validate IGDNet’s performance and robustness, suggesting its potential to inspire new directions in zero-shot learning for underexposure image enhancement.
\end{IEEEImpStatement}


\begin{figure*}
	\centering
\includegraphics[width=\linewidth]{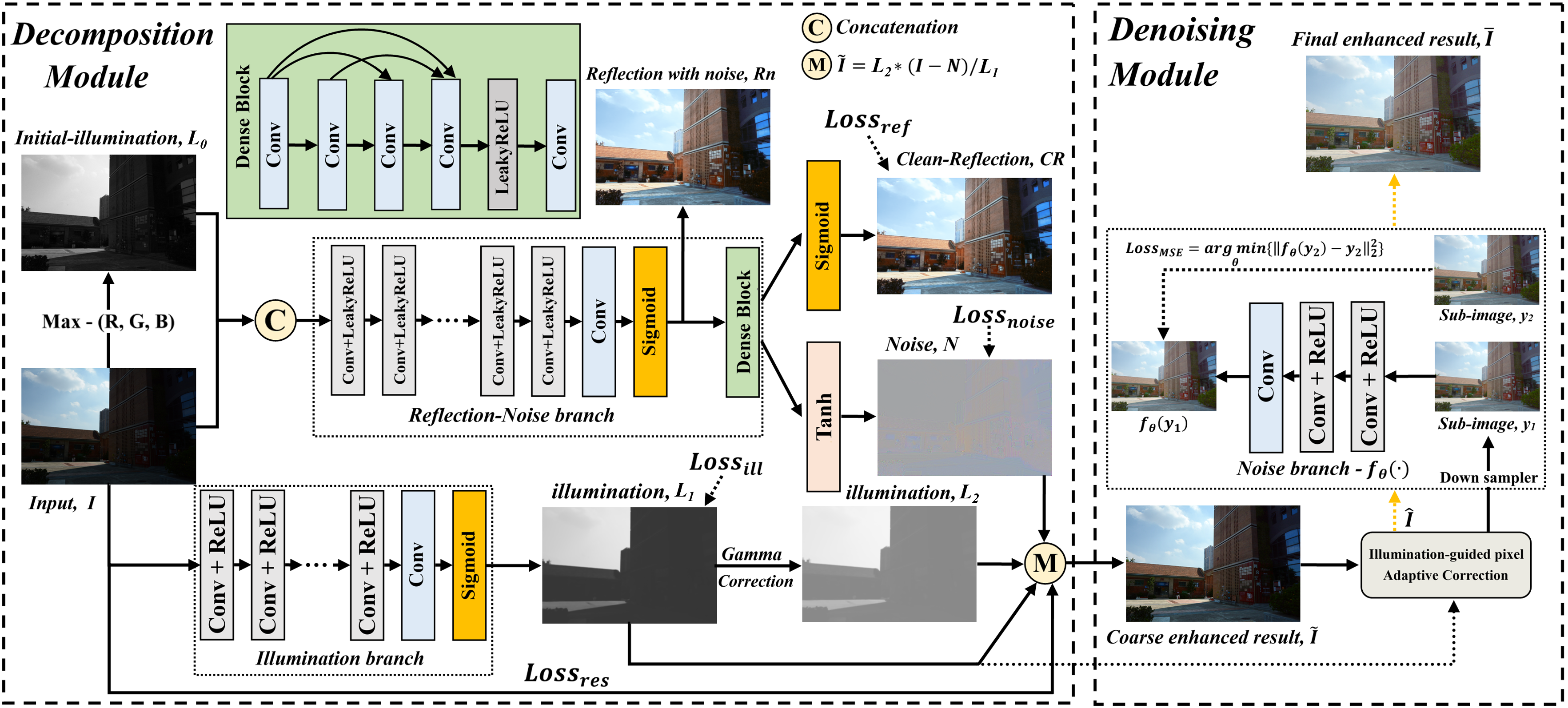}
	\caption{Illustration of the IGDNet framework. It includes a Decomposition Module and a Denoising Module. The Decomposition Module features an Illumination branch and a Reflection-Noise branch. The Denoising Module uses downsampling to divide the coarse enhancement results into two sub-images, which are then denoised using a simple network.}
	\label{fig1}
\end{figure*}

\IEEEPARstart{D}{ue} to limitations in environmental lighting and camera hardware, underexposed images captured under complex illumination conditions often suffer from severe quality degradation. This substantially impairs both visual perception and the performance of downstream high-level vision tasks~\cite{ref1,li2021low}.

Enhancing underexposed images entails several challenges, including correcting non-uniform illumination, suppressing noise, and preserving natural image fidelity. Non-uniform illumination often stems from uneven lighting or improper exposure, such as backlighting or low-light conditions. Such images typically suffer from two main issues: low global brightness, leading to reduced contrast and detail loss; and low local brightness, with uneven regions and shadows hindering information visibility. Additionally, many existing methods amplify noise during enhancement~\cite{ref27,ref30}, further degrading image quality and obscuring fine details.

Traditional methods such as Histogram Equalization~\cite{ref2}, the Retinex model~\cite{ref3}, and Gamma Correction~\cite{ref4} enhance overall brightness by leveraging global illumination statistics. However, due to limited modeling of illumination characteristics in underexposed images and inherent design constraints, these methods often lead to underexposure, color desaturation, and visible artifacts or noise. The problem is exacerbated in images with non-uniform lighting, where local regions may be either insufficiently enhanced or overexposed. To address complex lighting conditions, several studies have proposed enhancement strategies, such as wavelet decomposition with fractional-order diffusion for noise suppression and detail preservation~\cite{rahman2021structure}, exposure ratio adjustment with nonlinear stretching for correcting uneven illumination~\cite{rahman2022diverse}, and multi-stage fusion tailored to low-light industrial scenarios to improve visibility and structure~\cite{rahman2023efficient}. These methods effectively restore structure and reduce noise under challenging illumination.

To address the limitations of traditional methods, deep learning has facilitated the development of heuristic supervised networks that model the mapping between underexposed inputs and their enhanced counterparts. While effective, these approaches are heavily reliant on large-scale training data. Supervised methods~\cite{ref10,ref19,ref20} often exhibit limited generalization due to the inherent constraints of data-driven learning. Existing paired datasets frequently fail to capture the full complexity of real-world underexposed scenes. Moreover, constructing diverse and representative datasets is labor-intensive, and reliance on labeled data can significantly hinder performance, particularly in scenarios involving unfamiliar content.

Paired-data training schemes inherently suffer from limited generalization, and inaccuracies within the data further hinder model effectiveness. To reduce dependency on paired data, unsupervised approaches have been explored. Among them, Generative Adversarial Networks (GANs)~\cite{ref27} leverage unpaired data and have achieved notable success via adversarial learning. However, GANs are susceptible to mode collapse, leading to training instability and potentially unrealistic outputs. Furthermore, despite reducing reliance on paired data, GAN-based methods still require normal-light reference images during training, limiting their flexibility.

To eliminate reliance on reference images, Zero-DCE~\cite{ref30} was proposed based on curve estimation. While effective, it applies global enhancement directly, often causing color distortion and ignoring latent noise. SCI~\cite{ref34} simplifies the enhancement process using a self-calibration module with shared illumination learning. However, it lacks precision in handling varying brightness regions and may fail to differentiate between underexposed and well-lit areas.

While unsupervised methods show strong potential, they still face challenges related to data dependency. To ensure adaptability, these models must be trained on low-light images from diverse scenarios. However, if the training data lacks variety in lighting and scene conditions, model performance may degrade when applied to unseen cases. For example, noisy or poor-quality inputs can lead the model to learn suboptimal enhancement strategies. Thus, the diversity and quality of training data remain critical to model effectiveness.

Enhancing underexposed images under complex lighting conditions without a real reference remains a significant challenge. Inspired by Retinex theory~\cite{ref3,ref5,ref29}, and aiming to address issues such as uneven brightness and noise amplification, we propose a novel zero-shot method named IGDNet, which performs enhancement using only a single underexposed image. Unlike supervised methods that require extensive paired datasets or unsupervised methods dependent on low-light training data, IGDNet operates entirely without training data. This zero-shot design ensures robust generalization across diverse low-light scenarios, including previously unseen conditions. By integrating illumination guidance and noise-aware processing, IGDNet enables precise non-uniform illumination enhancement while suppressing noise amplification. As a result, it effectively overcomes data dependency and achieves superior performance under challenging lighting environments. The main contributions are as follows:

\begin{itemize}
    \item We propose IGDNet, a novel zero-shot image enhancement approach comprising a decomposition module and a denoising module. Notably, IGDNet operates without prior knowledge or training data.
    \item We design an illumination-guided pixel correction module that selectively enhances underexposed regions while maintaining consistency with human visual perception.
    \item Extensive experiments demonstrate that IGDNet outperforms existing unsupervised methods and achieves performance comparable to leading supervised approaches.
\end{itemize}

\section{Related Works}
\textbf{The conventional method} mainly processes each pixel to enhance brightness and reduce noise, typically falling into Histogram Equalization-based (HE)~\cite{ref2} and Retinex-based methods~\cite{ref5,ref6,ref7,ref9}. HE expands the image's global gray level range but may cause local overexposure due to abnormal stretching in specific regions.

Retinex-based approaches decompose images into reflection and illumination components, with reflection assumed to be invariant under varying lighting. Guo~\cite{ref5} initialized illumination using the maximum RGB value, refined with structural priors. Li~\cite{ref6} proposed a noise-aware Retinex model with optimization-based illumination estimation. Ren~\cite{ref7} incorporated low-rank priors to reduce reflection noise, while Xu~\cite{ref9} introduced a structure and texture-aware Retinex framework.

Learning-based methods are categorized into supervised and unsupervised learning. \textbf{Supervised learning} utilizes neural networks to learn the relationship between low-light images and their normal-light counterparts \cite{yan2025mobileie}. Rahman~\cite{rahman2020efficient} proposed a method that combines CNN-based image classification with illumination estimation to handle various exposure types. Wei~\cite{ref10} introduced Retinex-Net, while Wang~\cite{ref11} devised a progressive network for Retinex decomposition. Zhang~\cite{ref12} proposed the KinD network, further improved by KinD++~\cite{ref13}. Wang~\cite{ref14} trained on diverse exposure samples for image enhancement, while Fan~\cite{ref16} integrated semantic segmentation into Retinex. Ma~\cite{ref17} developed a lightweight residual network for multi-scale Retinex transformations, and Wang~\cite{ref18} proposed a flow regularization model. URetinex-Net~\cite{ref19} introduced implicit prior regularization. Additionally, techniques leveraging Transformer~\cite{ref20}, Diffusion~\cite{ref25} have advanced enhancement.

\textbf{Unsupervised learning} in image enhancement operates without a reference image of normal illumination. Methods like EnlightenGAN~\cite{ref27} and NeRCo~\cite{ref28} utilize unpaired sample data, employing adversarial generation with discriminators guiding enhancement. Kandula~\cite{kandula2023illumination} has developed an unsupervised network for low-light image enhancement, employing a multi-stage approach based on a context-guided illumination-adaptive norm. Zhu~\cite{ref29} refines RRD-Net iteratively using a non-reference loss function. Guo~\cite{ref30} adjusts underexposed images by estimating brightness enhancement curves~\cite{ref31}. Xia~\cite{ref32} enhances illumination correction in RRD-Net with an illumination estimation network. Liu~\cite{ref33} combines Retinex with neural network structure search. Ma~\cite{ref34} introduces the SCI method, incorporating a self-calibration module. CLIP-LIT~\cite{ref35}, SGZ~\cite{ref36}, and PairLIE~\cite{ref37} learn cues from images under varying illuminations. Additionally, Wang~\cite{ref38} designs a zero-reference frame combined with a diffusion model, achieving outstanding performance with training solely on normal-light images. Zhao~\cite{zhao2021retinexdip} proposes a generative method for Retinex decomposition that enhances low-light images by decoupling latent components and adjusting illumination.

\section{Proposed Method}
\subsection{Retinex Theory}
The Retinex theory~\cite{ref3} elucidates how illumination influences image acquisition: light enters the camera lens through object reflection, influenced by the external environment and determined by the object's reflectivity. In the realm of image enhancement, Retinex posits that underexposed images arise from insufficient illumination, underscoring the importance of lighting adjustment for enhanced quality. However, it often overlooks noise originating from diverse sources within the image. Thus, Retinex can be conceptualized as:
\begin{equation}\label{(0)}
Input=Reflection{\odot}Illumination+Noise
\end{equation}
where $\odot$ denotes pixel-wise multiplication.

\subsection{Network Architecture}
The framework of the IGDNet is depicted in Fig. \ref{fig1}. Rooted in the Retinex theory~\cite{ref3}, IGDNet decomposes underexposed images into illumination and reflectance components. Extracting noise directly from such degraded images poses a challenge. However, reflectance retains texture and color information, facilitating accurate noise extraction. Therefore, we devised Reflectance-Noise and Illumination branches within our decomposition module to estimate clean reflectance, illumination, and noise components.

\textbf{The Reflection-Noise branch} isolates reflection-noise components for a cleaner image representation. It consists of seven convolutional layers and a Dense Block. Recognizing that the reflection encapsulates the texture and color details of the underexposed image, we have deepened the branch to thoroughly capture its features. While ReLU is commonly used for its strong nonlinear properties, it can hinder the transmission of feature information backward. Therefore, the first six convolutional layers employ LeakyReLU, enabling the flow of negative information~\cite{ref40}, followed by sigmoid activation to obtain the reflection ($R_{n}$) with noise. Dense Block enhances information transmission by reusing features. Introducing a Tanh layer confines the noise value ($N$) within [-1,1]. To enhance the utilization of information from the underexposed image, we merge it with its initial illumination ($L_{0}$), resulting in a four-channel input. $L_{0}$ is typically estimated using the maximum intensity of the $R, G, B$ channels from the underexposed image, as illustrated in Equation \eqref{(1)}.
\begin{equation}\label{(1)}
L_{0}=max_{{\operatornamewithlimits{{c\in\{R,G,B\}}}}}I(x)
\end{equation}

Specifically, the output of the Reflection-Noise Branch is composed of a reflectance map and a noise map, produced through Sigmoid and Tanh activations, respectively. The Sigmoid function ensures that the reflectance output remains bounded within [0,1], conforming to standard image intensity ranges. The Tanh function, in contrast, allows the noise component to express both positive and negative deviations from the ideal image signal, enabling the model to learn a richer and more flexible noise representation. In this way, the branch serves not only to recover fine-grained reflectance details but also to perform preliminary suppression of structured and background noise embedded in the underexposed input.

\textbf{The Illumination branch} extracts global illumination to restore overall brightness in underexposed images. It consists of five convolutional layers. The first four layers utilize ReLU activation, followed by Sigmoid activation in the final layer, with pixel values normalized to [0, 1]. Assuming uniform illumination across the three color channels of the underexposed image~\cite{ref39}, the branch's output channel is set to 1. The extracted illumination, labeled as $L_{1}$, is then refined through Gamma-Transformation: $L_{2}=(L_{1})^{\gamma}$.

Once the decomposition module is finally determined, following Equation \eqref{(0)}, we substitute the adjusted illumination $L_{2}$ for the estimated $L_{1}$ from the illumination branch and remove noise $(I-N)$ to produce a coarse enhanced result $\Tilde{I}$:
\begin{equation}\label{(2)}
\Tilde{I}={L_{2}}{\ast}[(I-N)/{L_{1}}]
\end{equation}

Since the coarse enhanced result may not yield satisfactory visual quality, simple Gamma Correction is insufficient for handling non-uniform illumination. Conventional techniques often over-enhance locally bright regions, amplifying noise. To address this, we introduce an illumination-guided pixel adaptive correction module and a noise branch within the denoising module to refine illumination and effectively suppress noise.

\textbf{The Illumination-Guided Pixel Adaptive Correction Module} refines the coarse enhancement by applying pixel-wise curve adjustments. Two conditions are required for effective correction under non-uniform illumination: (1) contrast between original pixel values must be preserved, retaining brightness hierarchy; and (2) adjusted pixel values must remain within [0,1] to prevent overflow and information loss.

The illumination component, following Retinex, depicts the overall brightness of objects in the image, guiding the intensity mapping of pixels. Using these conditions, we formulate a correction function outlined in Equation \eqref{(3)}.
\begin{equation}\label{(3)}
\sigma(p(x,y)) =\left\{
	\begin{aligned}
	0.5+(2&^{p(x,y)-0.5}-1)  &\quad 0.5 {\leq} p(x,y) {\leq} 1\\
	0.5 & &\quad 0 {\leq} p(x,y) {\leq} 0.5\\
	\end{aligned}
	\right
	.
\end{equation}
where $\sigma(\cdot)$, $p(x,y)$ denote the correction factor and a normalized pixel value in the illumination component, respectively.

Pixels with optimal exposure tend to have a normalized mean approaching 0.5~\cite{ref41}. The illumination component, derived from the illumination branch, acts as a guiding threshold for determining the brightness coefficient of the image and dynamically adjusting the image's pixel values. If the illumination component corresponds one-to-one with pixel value pairs in the image, denoted as $[P_{ill}(x,y), P_{rgb}(x,y)]$, then the pixel-by-pixel adaptive adjustment process proceeds as follows:
\begin{equation}\label{(4)}
P_{rgb}(x,y)=P_{ill}(x,y)^{\sigma(P_{ill}(x,y))}
\end{equation}

\textbf{The Noise branch} further reduces the amplified noise present in underexposed images, enhancing visual clarity. It achieves denoising by combining downsampling and iterative optimization with Mean Squared Error (MSE)~\cite{ref42}. It consists of three convolutional layers, with the first two employing ReLU activation.

Initially, the downsampling decomposes the image from the pixel correction module into two sub-images ($y_{1},y_{2}$) with comparable noise levels. Subsequently, a sub-image is mapped to another sub-image after noise branch processing. This process can be described as:
\begin{equation}\label{(5)}
\left\{\begin{aligned}
&y_{1} = ds_{1} \otimes \hat{I},\quad y_{2} = ds_{2} \otimes \hat{I}\\
&\mathcal{L}_{MSE}= \mathop{arg\,min}\limits_{\theta}\{{\mid\mid} f_{\theta}(y_{1})-y_{2} {\mid\mid}^{2}_{2}\}\\
	\end{aligned}
	\right
	.
\end{equation}
where $\hat{I}$ is the input image; $ds_{1}$ and $ds_{2}$ denote downsampling kernels; $\otimes$ and $f_{\theta}(\cdot)$ represent the convolution operation and the Noise branch, respectively.

The noise branch produces the final restoration result $\Bar{I}$:
\begin{equation}\label{(6)}
\Bar{I}=\hat{I}-f_{\theta}(\hat{I})
\end{equation}

The Noise Branch plays a critical role in refining the image after coarse enhancement. While the Reflection-Noise Branch addresses early-stage noise based on structural decomposition, the Noise Branch focuses on residual high-frequency noise, such as texture jitter and edge artifacts, that may be amplified during the illumination correction process. By downsampling the enhanced image into a pseudo-paired structure and using a residual learning mechanism, the network learns a noise-specific mapping that is subtracted from the input to produce a cleaner final output. Together, the two branches form a coarse-to-fine denoising pipeline, significantly improving the robustness and visual fidelity of the enhancement results.

\subsection{Non-Reference Loss Function}
1) The loss function of the decomposition module comprises three components, with $\lambda_{n}$ serving as the weight factor:
\begin{equation}\label{(7)}
\mathcal{L}_{decom.}= \mathcal{L}_{recon} + \mathcal{L}_{tv} + \lambda_{n}\cdot\mathcal{L}_{noise}
\end{equation}

\textbf{Reconstruction loss} maintains consistency among illumination, reflection, and reconstruction outcomes by leveraging information from the underexposed image. This relationship can be expressed as:
\begin{align}
\begin{aligned}
\label{(7)}
&\mathcal{L}_{recon} = \mathcal{L}_{res} + \mathcal{L}_{ill} + \mathcal{L}_{ref} \\
&= \|I-(L_{1} * CR+N)\|_{1} + \|L_{1}-L_{0}\|_{1} + \|\frac{I}{L_{0}}-CR\|_{1}
\end{aligned}
\end{align}
where $\mathcal{L}{res}$, $\mathcal{L}_{ill}$, and $\mathcal{L}_{ref}$ represent the resulting loss, illumination loss, and reflection loss after Retinex reconstruction, respectively. $I$ denotes the underexposed image, while $L_{1}$, $CR$, and $N$ stand for the illumination, reflection, and noise components extracted by the decomposition module. $L_{0}$ is estimated using Equation \eqref{(1)}, and $\|\cdot\|_{1}$ denotes the $L_{1}$-norm.

\textbf{Total variation loss} is employed to quantify the disparity between neighboring pixels within the image~\cite{ref44}. This constraint on gradients ensures the reflection's constancy and enhances the illumination's smoothness.

\begin{equation}\label{(8)}
\mathcal{L}_{tv}=\sum_{h,w}(\lambda_{ref}^{-1}\|{\nabla^{(h,w)}}CR\|_{1}+\lambda_{ill}^{-1}\|{\nabla^{(h,w)}}{L_{1}}\|_{1})
\end{equation}

\begin{equation}\label{(9)}
\left\{\begin{aligned}
&\lambda_{ref} =G{\otimes}({\nabla^{(h,w)}L_{1}}_{gray})^{2}   \\
&\lambda_{ill} =normalize({L_{1}}\cdot{(\nabla^{(h,w)}CR)^{2}+\tau)} \\
	\end{aligned}
	\right
	.
\end{equation}
where $\lambda_{ref}$ and $\lambda_{ill}$ are utilized as weights to estimate the gradients of reflection and illumination; $\nabla^{(h,w)}$ is employed to compute both vertical and horizontal gradients; $G$ and $\otimes$ denote the Gaussian kernel and convolution operation, respectively. ${I_{1}}_{gray}$ is the grayscale version of the light, while $\tau$ denotes a very small equilibrium constant.

\textbf{Noise loss} is evaluated based on the illumination of the underexposed image. As the image contrast is augmented, potential noise is magnified, and the loss serves to constrain the noise ($N$)~\cite{ref29}.
\begin{equation}\label{(10)}
\mathcal{L}_{noise}=\|{L_{1}}\cdot{N}\|_{2}
\end{equation}

2) The loss functions of the denoising module consist of two components:
\begin{equation}\label{(14)}
\mathcal{L}_{denoi.}=\mathcal{L}_{sr} +\mathcal{L}_{sc}
\end{equation}

MSE measures the average deviation between the network output and the input by squaring the difference between predicted values for each sample and averaging them. Its definition is as follows:
\begin{equation}\label{(11)}
\displaystyle\mathcal{L}_{MSE}={\|f_{\theta}(y_{1})-y_{2}\|_{2}^{2}}
\end{equation}
where $w$ and $h$ represent the width and height of the image respectively; $y_{1}$ and $y_{2}$ represent two sub-images obtained by down-sampling; $f_{\theta}$ represents the noise branch.

Residual learning~\cite{ref45} and symmetric learning~\cite{ref46} offer avenues to optimize the denoising network. To better capture the essence of the noise, the MSE loss can be redefined to be more adaptive to the noise itself:
\begin{equation}\label{(12)}
\displaystyle\mathcal{L}_{sr}=\frac{1}{2}({\|y_{1}-f_{\theta}(y_{1})-y_{2}\|_{2}^{2}} + {\|y_{2}-f_{\theta}(y_{2})-y_{1}\|_{2}^{2}})
\end{equation}
where residual learning prioritizes capturing the noise in the image rather than the image itself. Meanwhile, symmetric learning encourages the network to generate more consistent output across various input samples.

To counteract the loss of high-resolution detail caused by downsampling, we introduce the global consistency optimization MSE loss, incorporating the original resolution image into the loss function.
\begin{align}
\begin{aligned}
\label{(13)}
\mathcal{L}_{sc}=&\frac{1}{2}({\|y_{1}-f_{\theta}(y_{1})-ds_{2} \otimes (\hat{I}-f_{\theta}(\hat{I}))\|_{2}^{2}} \\
& + {\|y_{2}-f_{\theta}(y_{2})-ds_{1} \otimes (\hat{I}-f_{\theta}(\hat{I}))\|_{2}^{2}})
\end{aligned}
\end{align}

\begin{table}[htbp]
  \centering
  \caption{Descriptions of various test datasets.}
\resizebox{1\columnwidth}{!}{
    \begin{tabular}{cccc}
    \hline
    Datasets & \multicolumn{1}{c}{Number (Test)} & Type  & Size \\
    \hline
    LOLv1~\cite{ref10} & 15 & Low overall brightness & 600$\times$400 \\
    LOLv2-Real~\cite{ref47} & 100 & Low overall brightness & 600$\times$400 \\
    LOLv2-Synthetic~\cite{ref47} & 100 & Low local brightness  & 384$\times$384 \\
    NTIRE LLE~\cite{ref48} & 140 & Low local brightness & 600$\times$400 \\
    Dark Face~\cite{ref51} & 100 & Low overall brightness & 1080$\times$720 \\
    \hline
    \end{tabular}}
  \label{tab:1}
\end{table}

\section{Experiments}
\subsection{Experimental Settings}
\textbf{Implementation Details.} IGDNet is implemented in PyTorch with key parameters set as follows: $\gamma=0.4$, $\lambda_{n}=5,000$. All evaluation experiments were conducted on a mobile workstation equipped with a 13th Gen Intel(R) Core(TM) i9-13900H 2.60GHz CPU and NVIDIA GeForce GTX 4080 Laptop GPU. The decomposition and denoising modules were respectively iterated 1,000 and 2,000 times. The model was optimized using Adam with a learning rate of 0.01.

\textbf{Dataset and Metrics.} We use the test sets of LOLv1 and LOLv2-Real to represent real-world indoor scenes with low overall brightness, while LOLv2-Synthetic simulates non-uniform illumination. The NTIRE LLE dataset contains both uniform and non-uniform low-light scenes from indoor and outdoor environments; to specifically evaluate performance under non-uniform illumination, we retained only the images exhibiting uneven lighting conditions. Additionally, the Dark Face dataset is employed to assess the effectiveness of enhancement methods in higher-level visual tasks. Detailed descriptions are provided in Table~\ref{tab:1}.

To evaluate enhancement performance, we adopt five image quality metrics: PSNR, SSIM, LPIPS~\cite{ref50}, LOE~\cite{ref49}, and MAE~\cite{ref52}. Higher PSNR and SSIM indicate closer alignment with the ground truth, while lower LPIPS, LOE, and MAE reflect better perceptual quality and fidelity. Additionally, to assess the denoising module’s effectiveness, we employ NES~\cite{chen2015efficient}, which estimates noise levels by analyzing the statistical relationship between the eigenvalues of image patch covariance matrices and noise variance.

\begin{table*}[b]
  \centering
  \caption{Quantitative results. In unsupervised methods, we highlight the best scores in red and the next best in blue. We also specify the training set for each method, with "LOL+" indicating fusion with other datasets, and "$\textbf{-}$" denoting no reliance on a training set.}
  \resizebox{\textwidth}{!}{
    \begin{tabular}{r|c|c|ccccc|ccccc|ccccc}
    \hline
    \multicolumn{2}{c|}{Datasets} & Train Set & \multicolumn{5}{c|}{LOLv1+LOLv2-Real} & \multicolumn{5}{c|}{LOLv2-Synthetic}  & \multicolumn{5}{c}{NITRE LLE} \\
    \hline
    \multicolumn{3}{c|}{Metrics} & \multicolumn{1}{c}{PSNR$\uparrow$} & \multicolumn{1}{c}{SSIM$\uparrow$} & \multicolumn{1}{c}{LPIPS$\downarrow$} & \multicolumn{1}{c}{LOE$\downarrow$} & \multicolumn{1}{c|}{MAE$\downarrow$} & \multicolumn{1}{c}{PSNR$\uparrow$} & \multicolumn{1}{c}{SSIM$\uparrow$} & \multicolumn{1}{c}{LPIPS$\downarrow$} & \multicolumn{1}{c}{LOE$\downarrow$} & \multicolumn{1}{c|}{MAE$\downarrow$} & \multicolumn{1}{c}{PSNR$\uparrow$} & \multicolumn{1}{c}{SSIM$\uparrow$} & \multicolumn{1}{c}{LPIPS$\downarrow$} & \multicolumn{1}{c}{LOE$\downarrow$} & \multicolumn{1}{c}{MAE$\downarrow$} \\
    \hline
          & Retinex-Net~\cite{ref10} & LOL   & 16.19  & 0.403  & 0.534  & 0.346  & 0.172 & 17.14  & 0.762  & 0.255  & 0.210  & 0.130  & 17.56  & 0.715  & 0.217  & 0.283  & 0.122  \\
          & KinD~\cite{ref12}  & LOL   & 20.21  & 0.814  & 0.147  & 0.245  & 0.131 & 17.28  & 0.759  & 0.252  & 0.163  & 0.157 & 20.05  & 0.836  & 0.129  & 0.210  & 0.103  \\
          & KinD++~\cite{ref13} & LOL   & 16.64  & 0.662  & 0.410  & 0.288  & 0.148 & 17.48  & 0.789  & 0.232  & 0.159  & 0.137 & 19.68  & 0.840  & 0.131  & 0.211  & 0.099  \\
          & URetinex-Net~\cite{ref19} & LOL   & 20.93  & 0.854  & 0.104  & 0.245  & 0.117 & 18.25  & 0.824  & 0.196  & 0.168  & 0.121 & 19.62  & 0.838  & 0.125  & 0.227  & 0.098  \\
    \multicolumn{1}{c|}{Supervised} & LLFlow~\cite{ref18} & LOL   & 25.29  & 0.906  & 0.084  & 0.207  & 0.063 & 17.11  & 0.812  & 0.224  & 0.148  & 0.138 & 18.63  & 0.844  & 0.116  & 0.202  & 0.114  \\
          & RetinexFormer~\cite{ref20} & LOL   & 28.48  & 0.877  & 0.117  & 0.256  & 0.045 & 25.67  & 0.930  & 0.059  & 0.142  & 0.058 & 16.98  & 0.808  & 0.158  & 0.246  & 0.134  \\
          & RetinexFormer~\cite{ref20} & MIT~\cite{ref57}   & 13.02  & 0.426  & 0.365  & 0.280  & 0.454 & 13.18  & 0.555  & 0.300  & 0.172  & 0.291 & 13.47  & 0.562  & 0.229  & 0.222  & 0.264  \\
          & DiffLL~\cite{ref25} & LOL+  & 28.54  & 0.870  & 0.102  & 0.253  & 0.063 & 29.45  & 0.908  & 0.093  & 0.153  & 0.027 & 18.15  & 0.778  & 0.172  & 0.228  & 0.124  \\
    \hline
          & EnlightenGAN~\cite{ref27} & Own data & 18.57  & 0.700  & 0.302  & 0.291  & 0.161 & 16.57  & 0.775  & 0.212  & 0.156  & 0.158 & 18.28  & 0.803  & 0.161  & 0.225  & 0.117  \\
          & PairLIE~\cite{ref37} & LOL & 19.70  & 0.774  & 0.235  & \textcolor{blue}{0.278}  & 0.132 & \textcolor{blue}{19.07}  & 0.797  & 0.230  & 0.180  & \textcolor{blue}{0.109} & 19.84  & 0.819  & 0.156  & 0.247  & \textcolor{blue}{0.094}  \\
          & NeRCo~\cite{ref28} & LSRW~\cite{ref55}  & 19.67  & 0.720  & 0.266  & 0.310  & \textcolor{red}{0.069} & 17.59  & 0.734  & 0.301  & 0.225  & 0.129 & 17.93  & 0.780  & 0.212  & 0.258  & 0.119  \\
          & CLIP-LIT~\cite{ref35} & Own data & 14.82  & 0.524  & 0.371  & 0.320  & 0.302 & 16.19  & 0.775  & 0.204  & 0.161  & 0.187 & 16.78  & 0.767  & 0.141  & 0.233  & 0.171  \\
          & Zero\text{-}DCE~\cite{ref30} & Own data & 17.64  & 0.572  & 0.316  & 0.296  & 0.225 & 17.76  & 0.816  & \textcolor{blue}{0.168}  & 0.138  & 0.147 & \textcolor{blue}{20.26}  & \textcolor{blue}{0.832}  & \textcolor{red}{0.103}  & 0.208  & 0.102  \\
          & Zero\text{-}DCE++~\cite{ref31} & Own data & 17.03  & 0.445  & 0.314  & 0.391  & 0.220  & 17.58  & 0.812  & 0.187  & 0.124  & 0.140  & 17.48  & 0.605  & 0.144  & 0.324  & 0.123  \\
          & RUAS~\cite{ref33}  & LOL   & 15.47  & 0.490  & 0.305  & 0.330  & 0.237 & 13.40  & 0.644  & 0.364  & 0.274  & 0.186 & 12.60  & 0.671  & 0.304  & 0.312  & 0.194  \\
    \multicolumn{1}{c|}{Unsupervised} & RUAS~\cite{ref33}  & MIT~\cite{ref57}   & 13.62  & 0.462  & 0.346  & 0.292  & 0.434 & 13.77  & 0.638  & 0.305  & 0.172  & 0.221 & 13.73  & 0.655  & 0.238  & 0.244  & 0.227  \\
          & RUAS~\cite{ref33}  & FACE~\cite{ref51}  & 15.05  & 0.456  & 0.371  & 0.292  & 0.235 & 13.19  & 0.673  & 0.309  & 0.234  & 0.190  & 12.36  & 0.698  & 0.274  & 0.290  & 0.194  \\
          & SGZ~\cite{ref36}& Own data & 17.07  & 0.417  & 0.324  & 0.369  & 0.209 & 17.94  & \textcolor{blue}{0.823}  & 0.178  & 0.123  & 0.130  & 17.33  & 0.606  & 0.146  & 0.324  & 0.120  \\
          & SCI~\cite{ref34}& LOL+  & 16.97  & 0.532  & 0.312  & 0.289  & 0.262 & 15.43  & 0.748  & 0.233  & 0.136  & 0.168 & 15.56  & 0.770  & 0.178  & 0.219  & 0.161  \\
          & SCI~\cite{ref34}& MIT~\cite{ref57}   & 11.67  & 0.395  & 0.361  & 0.286  & 0.544 & 13.92  & 0.626  & 0.276  & 0.120  & 0.254 & 13.88  & 0.622  & 0.208  & 0.203  & 0.256  \\
          & SCI~\cite{ref34}& FACE~\cite{ref51}  & 16.80  & 0.543  & 0.322  & 0.297  & 0.282 & 16.69  & 0.743  & 0.242  & 0.144  & 0.158 & 17.99  & 0.782  & 0.159  & 0.218  & 0.134  \\
          & ZR-PQR~\cite{ref38}& COCO~\cite{ref56}  & \textcolor{blue}{20.31}  & \textcolor{red}{0.808}  & \textcolor{blue}{0.202}  & 0.281  & \textcolor{blue}{0.119} & 16.11  & 0.759  & 0.250  & 0.188  & 0.167 & 18.37  & 0.793  & 0.157  & 0.230  & 0.128  \\
\cline{2-18}          & LIME~\cite{ref5}& -     & 17.01  & 0.477  & 0.386  & 0.293  & 0.177 & 16.37  & 0.747  & 0.236  & 0.170  & 0.140  & 17.66  & 0.761  & 0.174  & 0.224  & 0.124  \\
          & RRD-Net~\cite{ref29} & -     & 13.60  & 0.480  & 0.323  & 0.271  & 0.412 & 14.84  & 0.655  & 0.247  & \textcolor{blue}{0.119}  & 0.240  & 14.66  & 0.659  & 0.177  & \textcolor{red}{0.197}  & 0.231  \\
          & RDHCE~\cite{ref32} & -     & 17.11  & 0.482  & 0.360  & 0.279  & 0.191 & 15.94  & 0.747  & 0.225  & 0.148  & 0.184 & 17.92  & 0.763  & 0.153  & 0.210  & 0.140  \\
          & COLIE~\cite{chobola2024fast} & - & 16.24  & 0.498  & 0.349  & 0.273 & 0.234 & 16.12  & 0.743  & 0.207  & 0.142 & 0.185 & 16.32  & 0.748  & 0.162  & 0.217  & 0.176  \\
          & OUR   & -     & \textcolor{red}{20.41}  & \textcolor{blue}{0.803}  & \textcolor{red}{0.172}  & \textcolor{red}{0.239}  & 0.136 & \textcolor{red}{19.75}  & \textcolor{red}{0.860}  & \textcolor{red}{0.157}  & \textcolor{red}{0.119}  & \textcolor{red}{0.104} & \textcolor{red}{20.47}  & \textcolor{red}{0.845}  & \textcolor{blue}{0.110}  & \textcolor{blue}{0.198}  & \textcolor{red}{0.092}  \\
    \hline
    \end{tabular}}
  \label{tab:2}%
\end{table*}%

\textbf{Compared Methods.} IGDNet is compared with 20 representative underexposure image enhancement methods, including 13 unsupervised and 7 supervised approaches. The inclusion of supervised methods helps illustrate generalization performance and establish an upper-bound reference. For fairness and consistency, we use the official open-source implementations and adhere to the original parameter settings reported in each method.

\begin{figure*}[b]
\centering
 \begin{minipage}[htbp]{0.1\linewidth}
\centerline{\includegraphics[width=\textwidth]{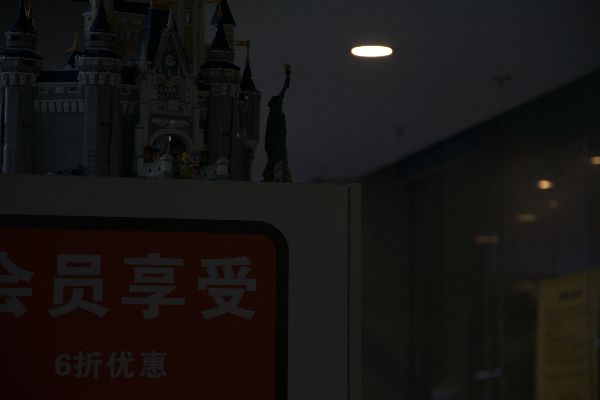}}
\centerline{\includegraphics[width=\textwidth]{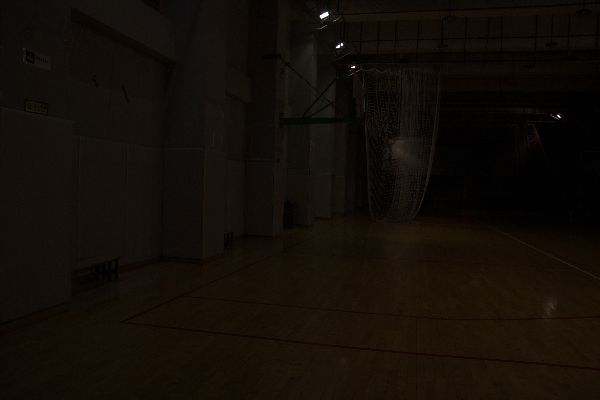}}
\centerline{\includegraphics[width=\textwidth]{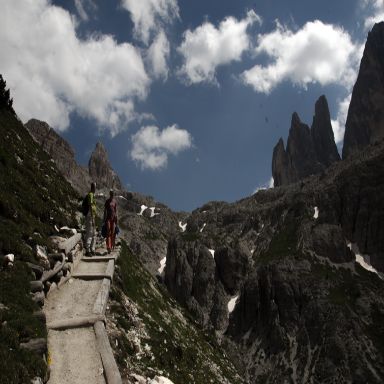}}
\centerline{\includegraphics[width=\textwidth]{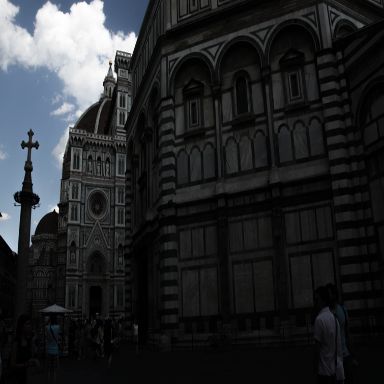}}
\centerline{\includegraphics[width=\textwidth]{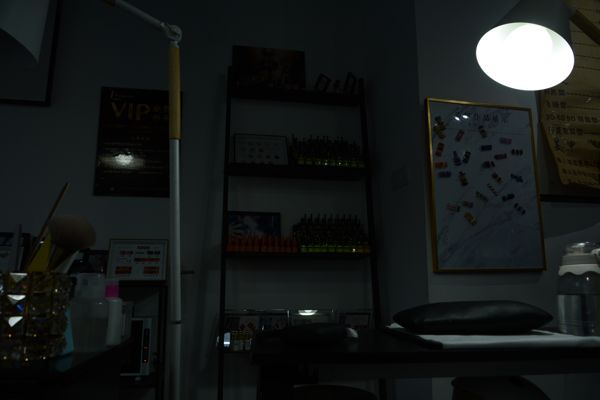}}
\centerline{\includegraphics[width=\textwidth]{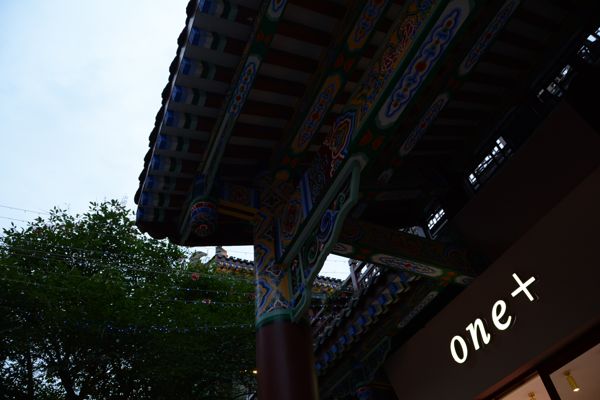}}
\centerline{\scriptsize\textbf{(a)}}
\end{minipage}%
\begin{minipage}[htbp]{0.1\linewidth}
\centerline{\includegraphics[width=\textwidth]{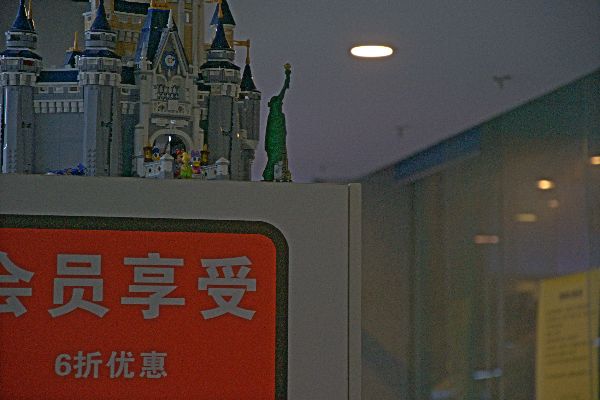}}
\centerline{\includegraphics[width=\textwidth]{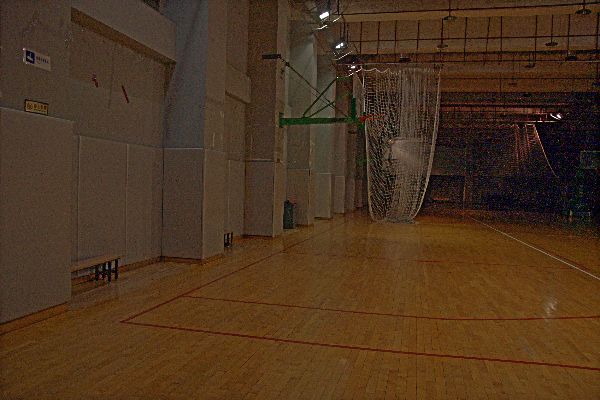}}
\centerline{\includegraphics[width=\textwidth]{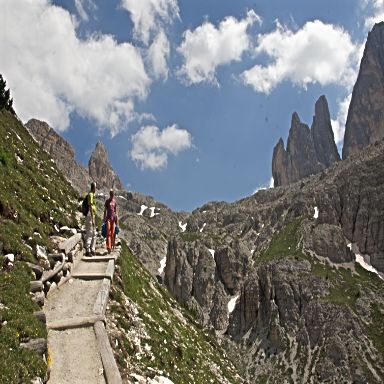}}
\centerline{\includegraphics[width=\textwidth]{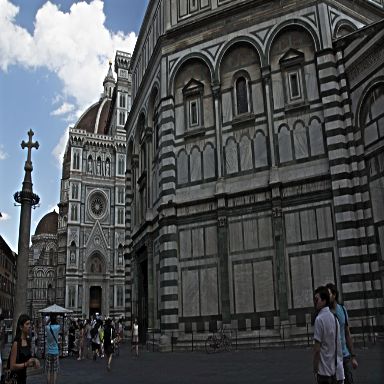}}
\centerline{\includegraphics[width=\textwidth]{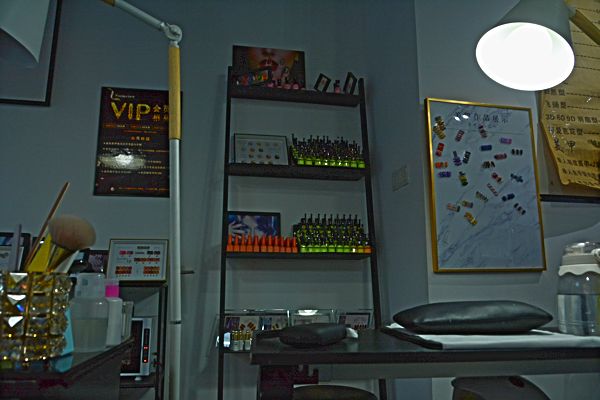}}
\centerline{\includegraphics[width=\textwidth]{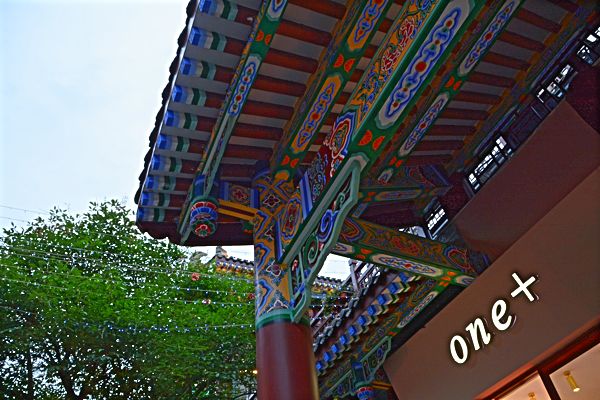}}
\centerline{\scriptsize\textbf{(b)}}
\end{minipage}%
\begin{minipage}[htbp]{0.1\linewidth}
\centerline{\includegraphics[width=\textwidth]{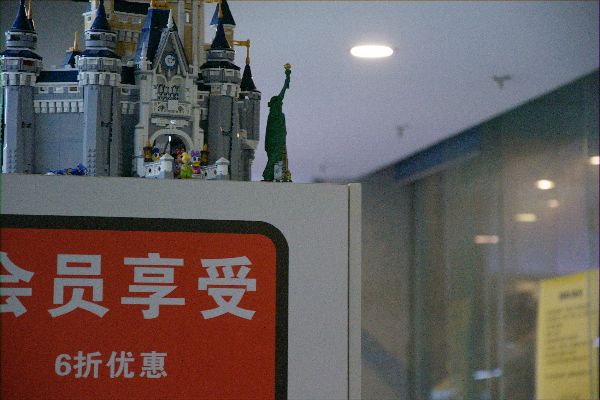}}
\centerline{\includegraphics[width=\textwidth]{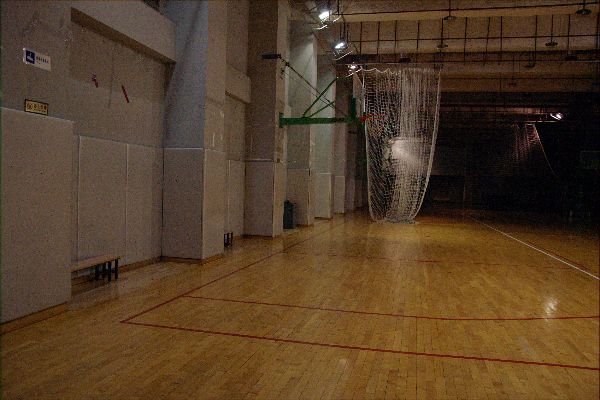}}
\centerline{\includegraphics[width=\textwidth]{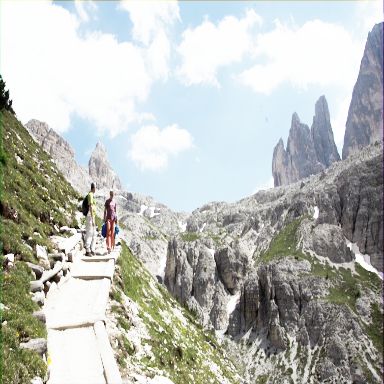}}
\centerline{\includegraphics[width=\textwidth]{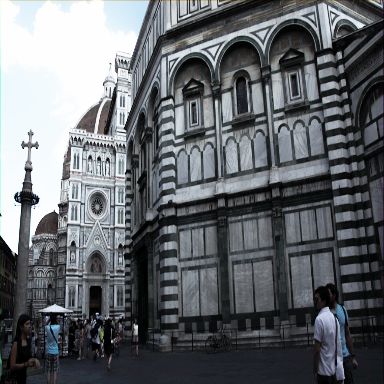}}
\centerline{\includegraphics[width=\textwidth]{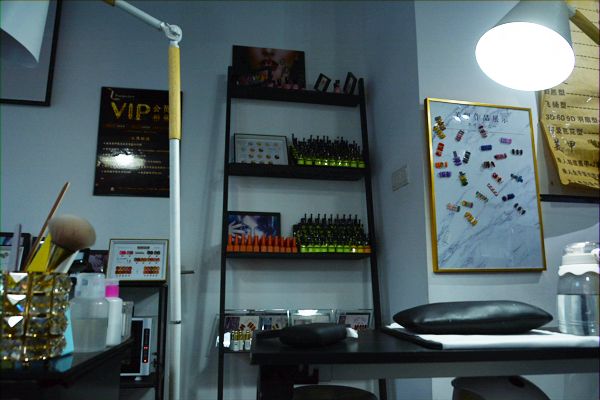}}
\centerline{\includegraphics[width=\textwidth]{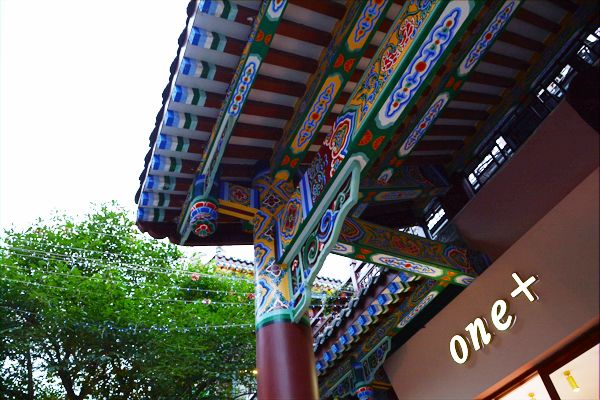}}
\centerline{\scriptsize\textbf{(c)}}
\end{minipage}%
\begin{minipage}[htbp]{0.1\linewidth}
\centerline{\includegraphics[width=\textwidth]{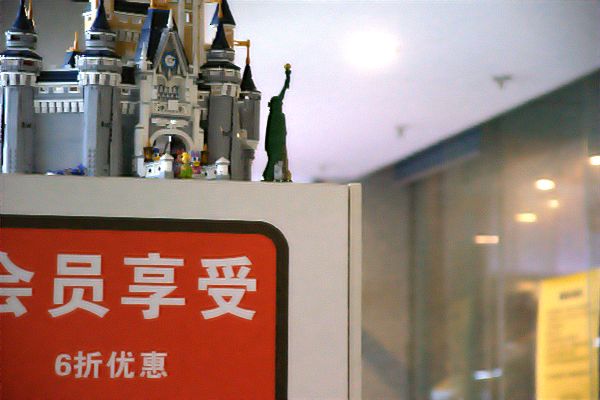}}
\centerline{\includegraphics[width=\textwidth]{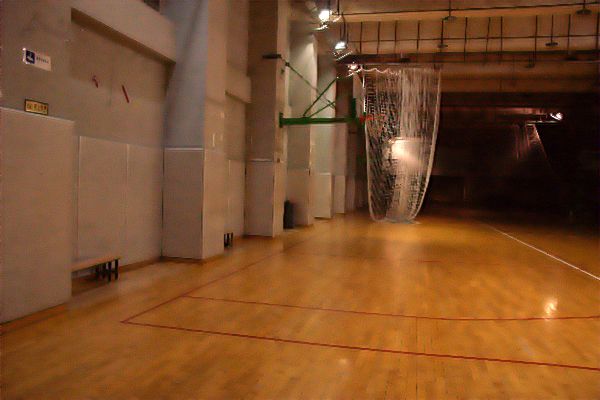}}
\centerline{\includegraphics[width=\textwidth]{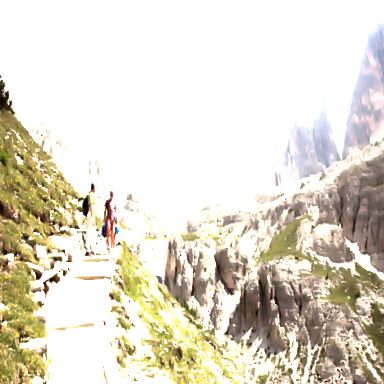}}
\centerline{\includegraphics[width=\textwidth]{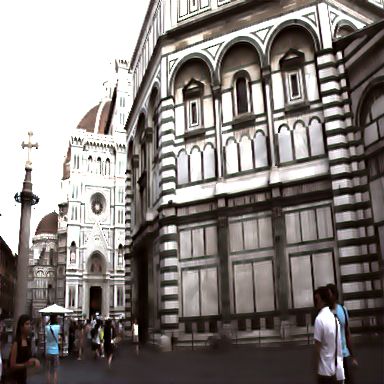}}
\centerline{\includegraphics[width=\textwidth]{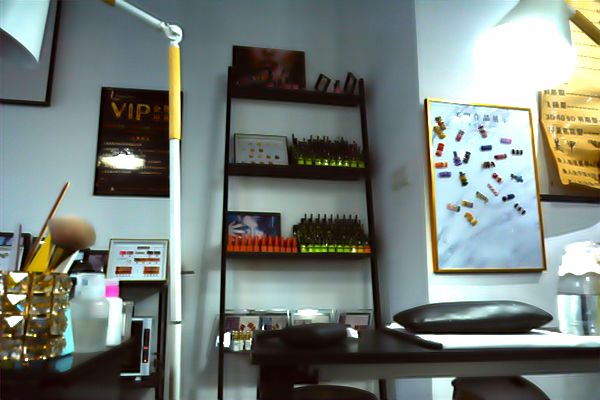}}
\centerline{\includegraphics[width=\textwidth]{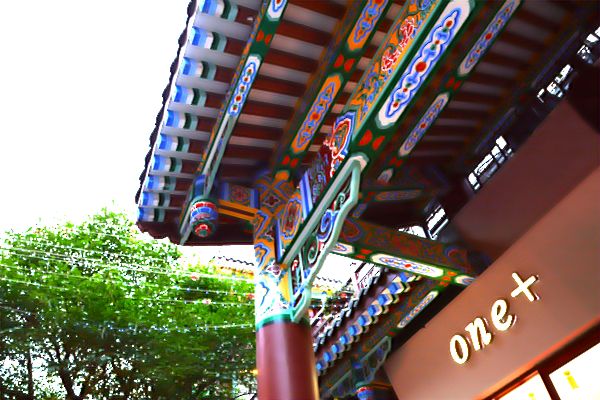}}
\centerline{\scriptsize\textbf{(d)}}
\end{minipage}%
\begin{minipage}[htbp]{0.1\linewidth}
\centerline{\includegraphics[width=\textwidth]{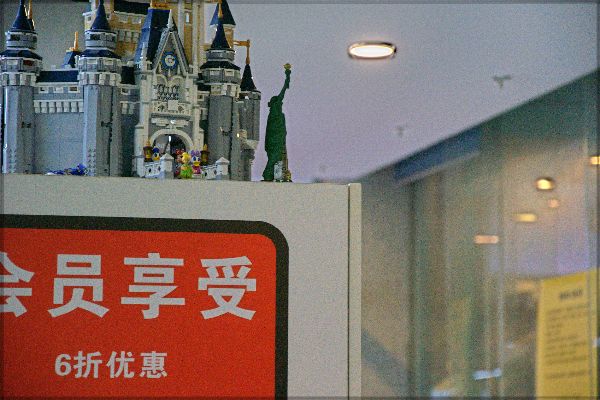}}
\centerline{\includegraphics[width=\textwidth]{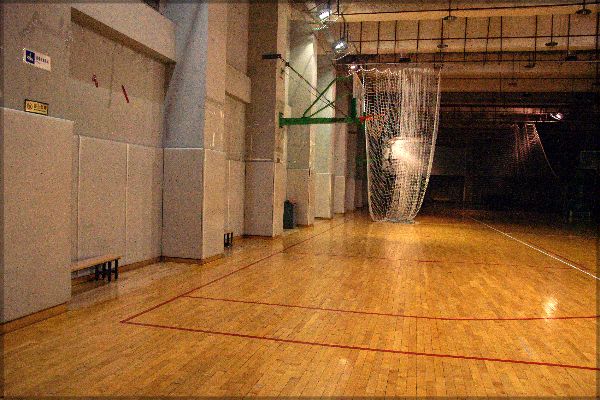}}
\centerline{\includegraphics[width=\textwidth]{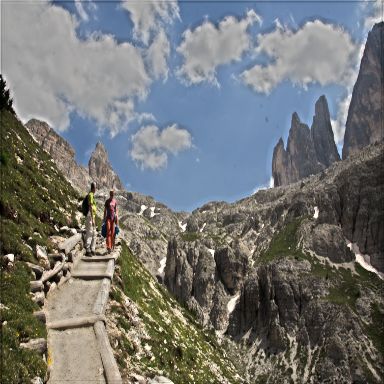}}
\centerline{\includegraphics[width=\textwidth]{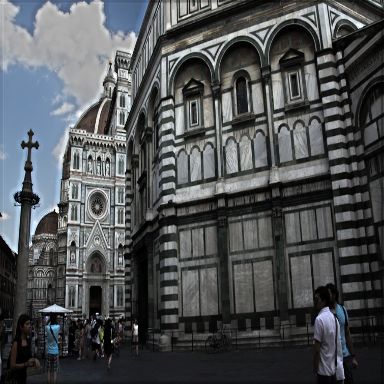}}
\centerline{\includegraphics[width=\textwidth]{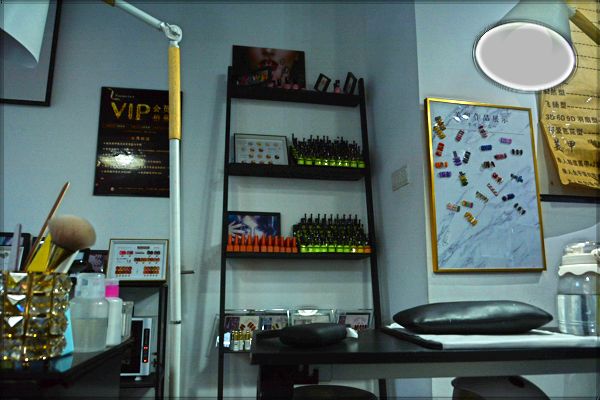}}
\centerline{\includegraphics[width=\textwidth]{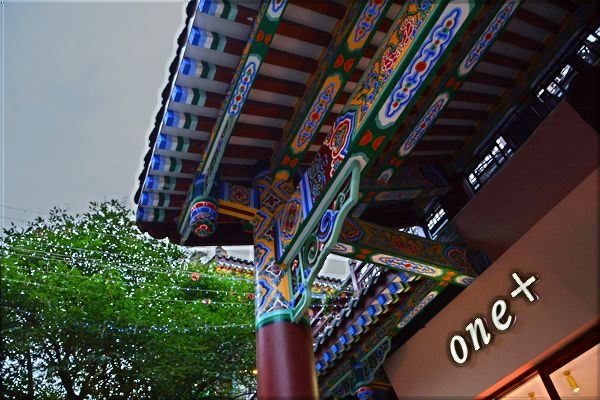}}
\centerline{\scriptsize\textbf{(e)}}
\end{minipage}%
\begin{minipage}[htbp]{0.1\linewidth}
\centerline{\includegraphics[width=\textwidth]{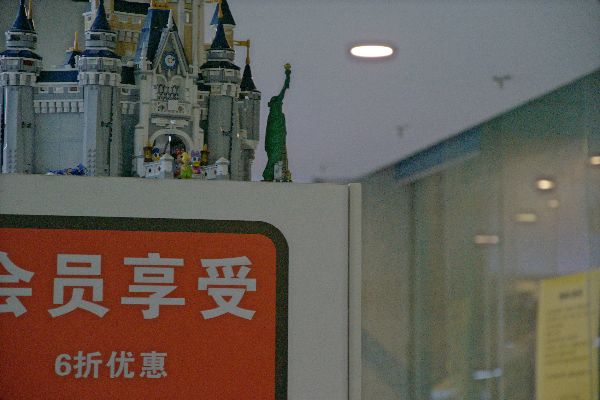}}
\centerline{\includegraphics[width=\textwidth]{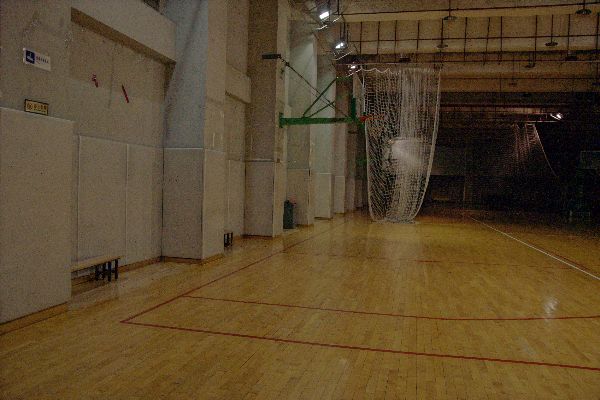}}
\centerline{\includegraphics[width=\textwidth]{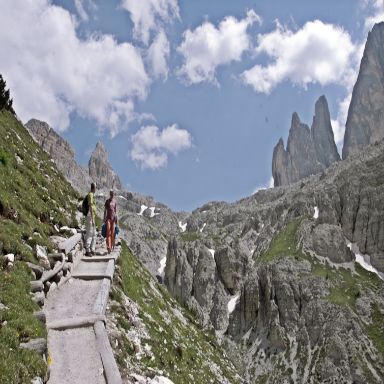}}
\centerline{\includegraphics[width=\textwidth]{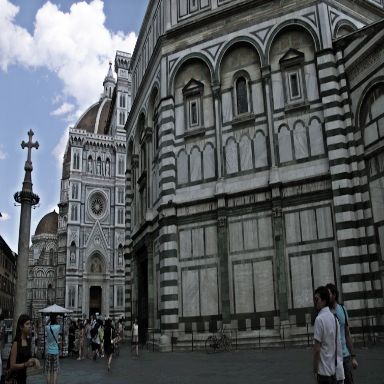}}
\centerline{\includegraphics[width=\textwidth]{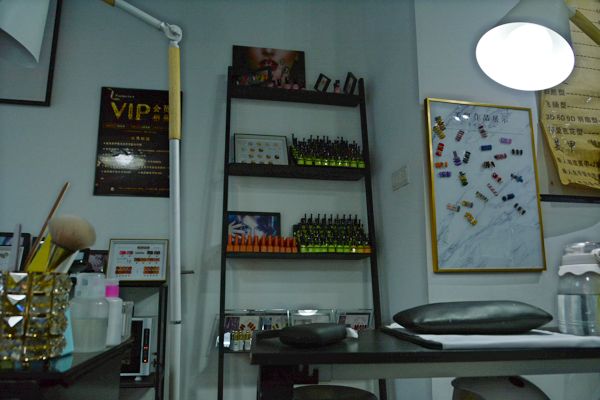}}
\centerline{\includegraphics[width=\textwidth]{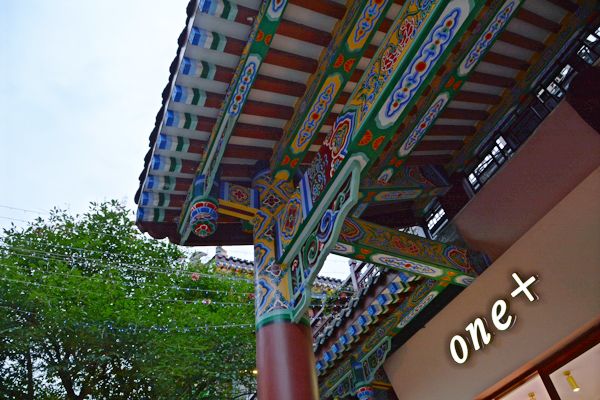}}
\centerline{\scriptsize\textbf{(f)}}
\end{minipage}%
\begin{minipage}[htbp]{0.1\linewidth}
\centerline{\includegraphics[width=\textwidth]{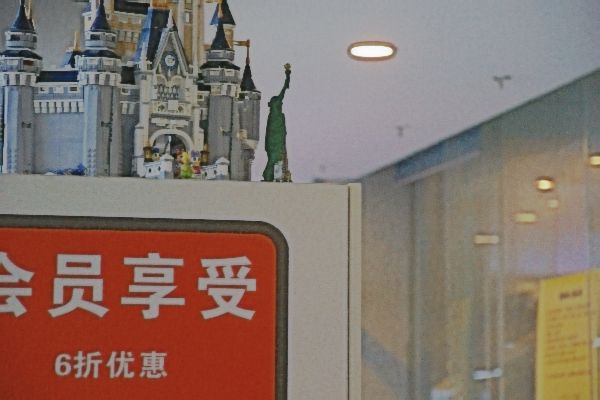}}
\centerline{\includegraphics[width=\textwidth]{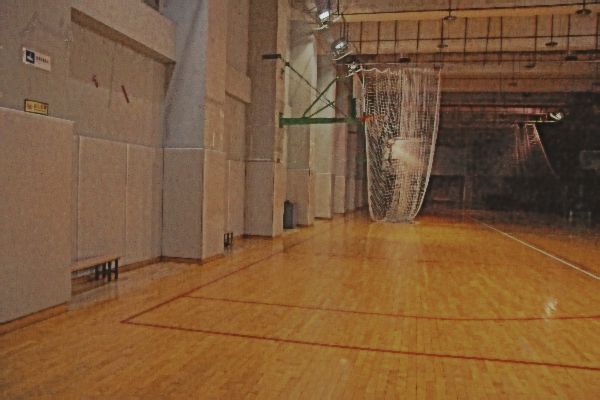}}
\centerline{\includegraphics[width=\textwidth]{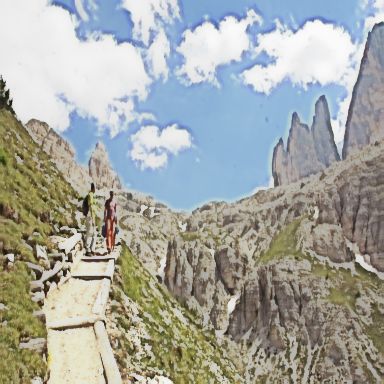}}
\centerline{\includegraphics[width=\textwidth]{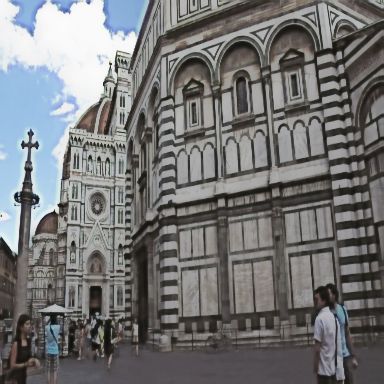}}
\centerline{\includegraphics[width=\textwidth]{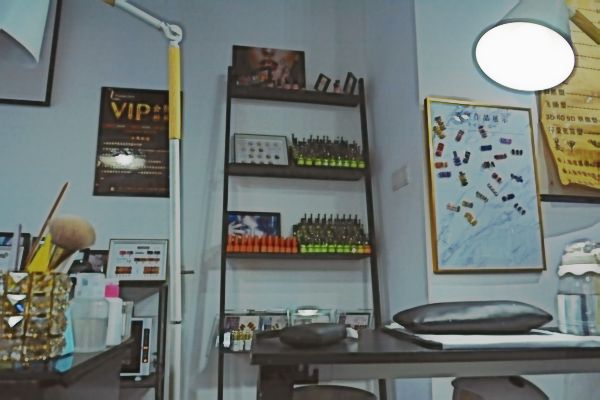}}
\centerline{\includegraphics[width=\textwidth]{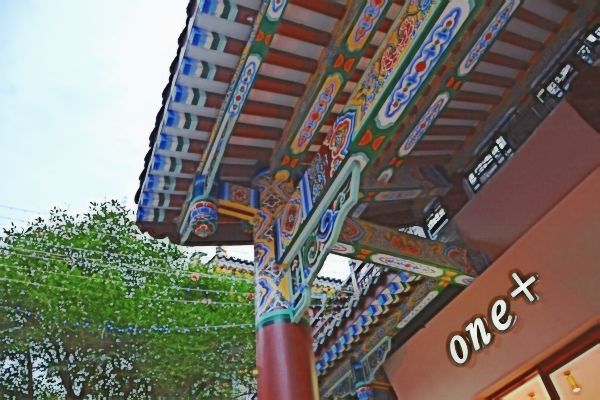}}
\centerline{\scriptsize\textbf{(g)}}
\end{minipage}%
\begin{minipage}[htbp]{0.1\linewidth}
\centerline{\includegraphics[width=\textwidth]{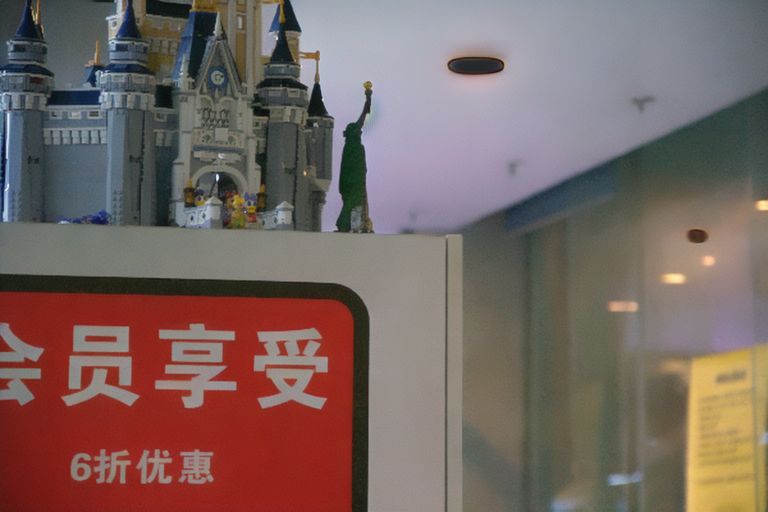}}
\centerline{\includegraphics[width=\textwidth]{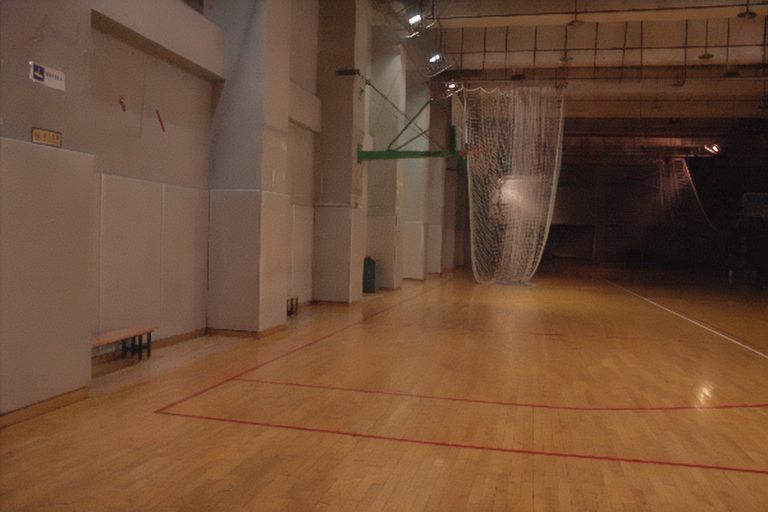}}
\centerline{\includegraphics[width=\textwidth]{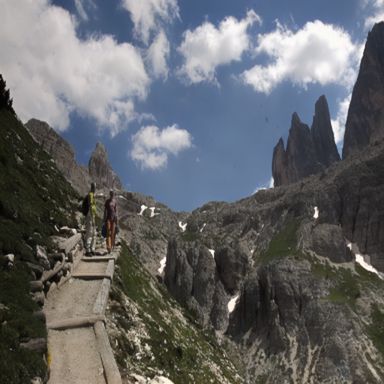}}
\centerline{\includegraphics[width=\textwidth]{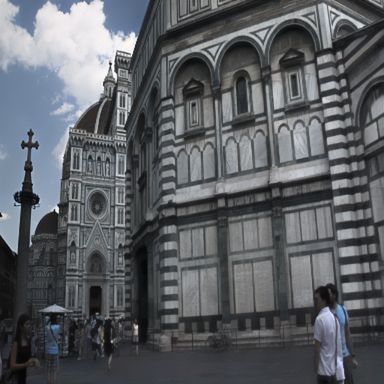}}
\centerline{\includegraphics[width=\textwidth]{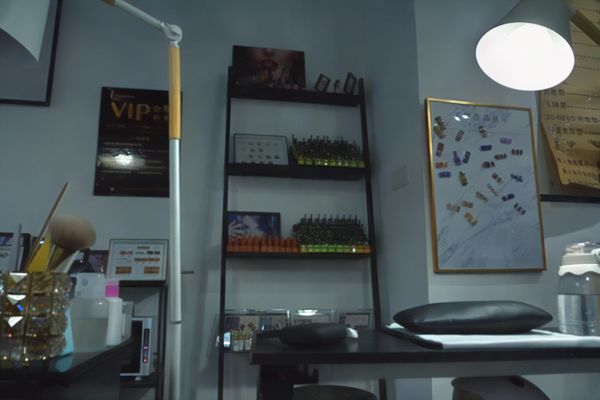}}
\centerline{\includegraphics[width=\textwidth]{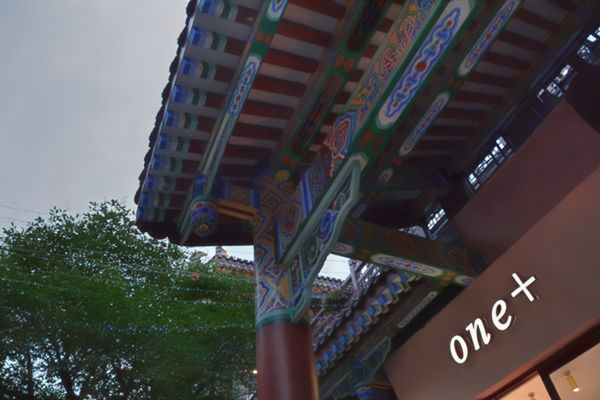}}
\centerline{\scriptsize\textbf{(h)}}
\end{minipage}%
\begin{minipage}[htbp]{0.1\linewidth}
\centerline{\includegraphics[width=\textwidth]{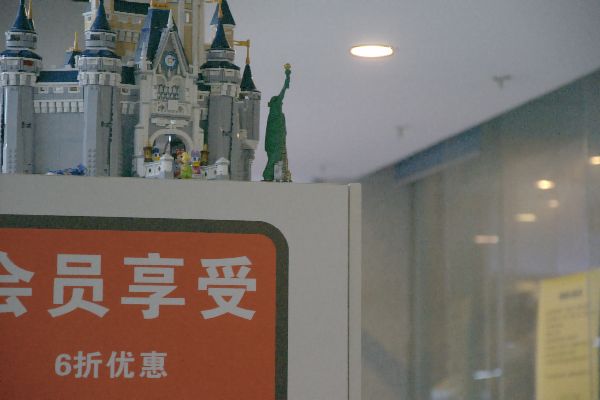}}
\centerline{\includegraphics[width=\textwidth]{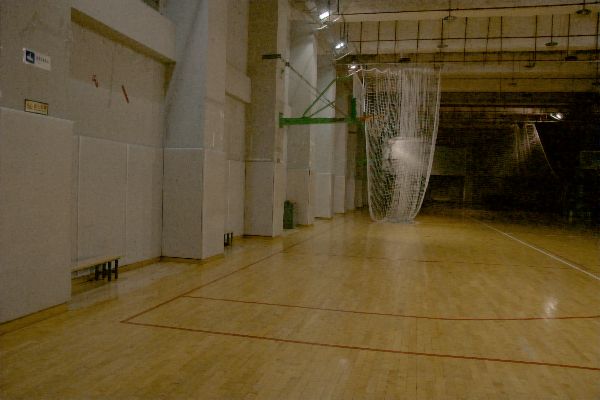}}
\centerline{\includegraphics[width=\textwidth]{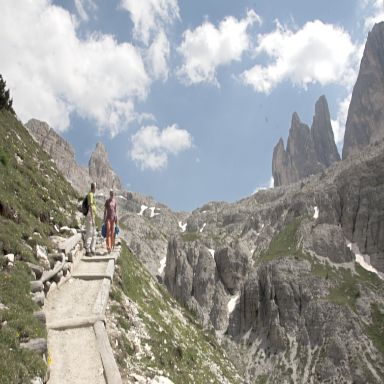}}
\centerline{\includegraphics[width=\textwidth]{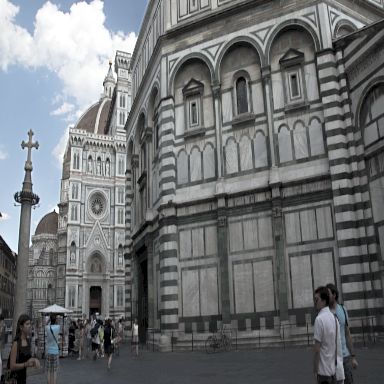}}
\centerline{\includegraphics[width=\textwidth]{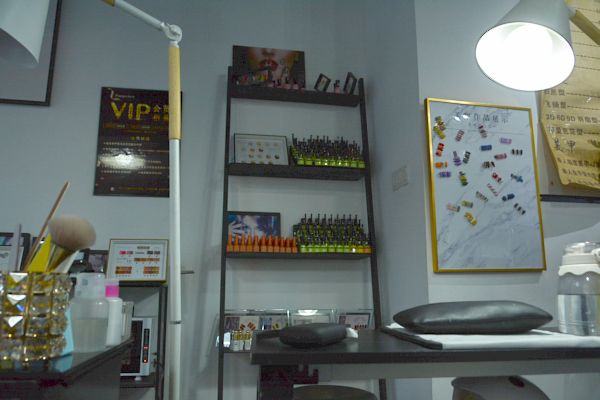}}
\centerline{\includegraphics[width=\textwidth]{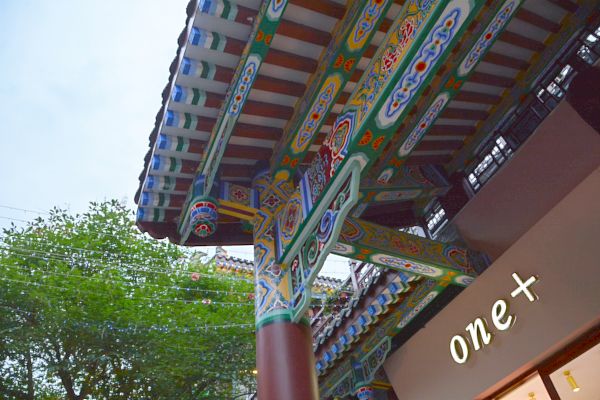}}
\centerline{\scriptsize\textbf{(i)}}
\end{minipage}%
\begin{minipage}[htbp]{0.1\linewidth}
\centerline{\includegraphics[width=\textwidth]{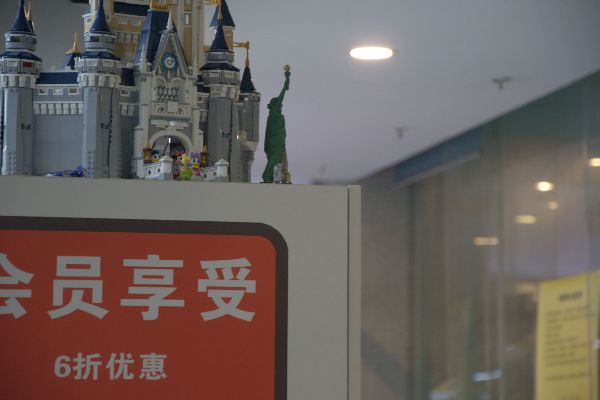}}
\centerline{\includegraphics[width=\textwidth]{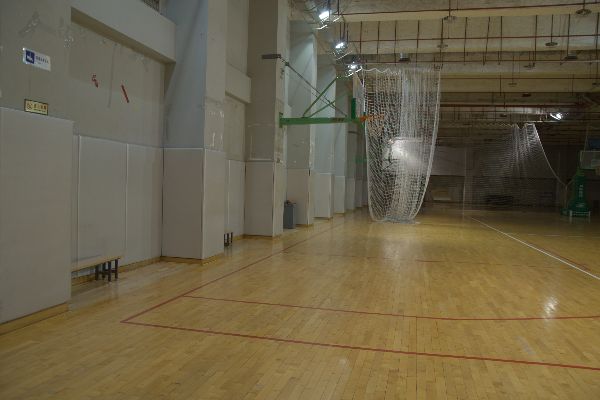}}
\centerline{\includegraphics[width=\textwidth]{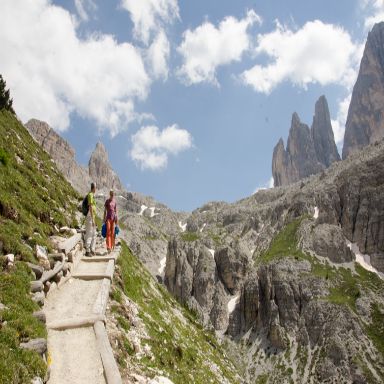}}
\centerline{\includegraphics[width=\textwidth]{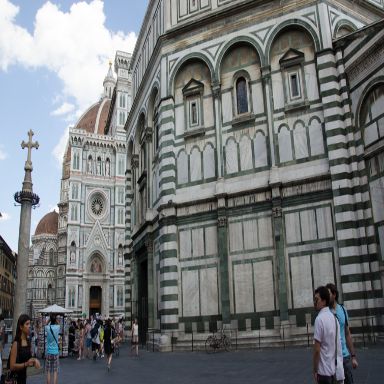}}
\centerline{\includegraphics[width=\textwidth]{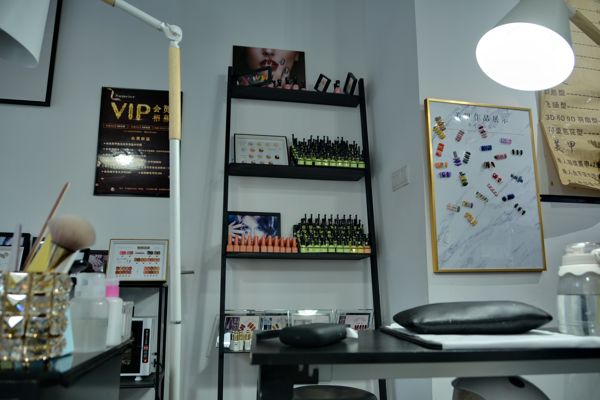}}
\centerline{\includegraphics[width=\textwidth]{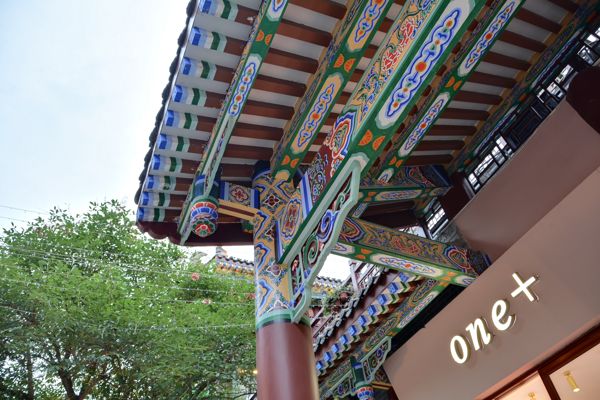}}
\centerline{\scriptsize\textbf{(j)}}
\end{minipage}%
\caption{\label{fig2}Visualization comparison examples of various methods. (a) Input (b) CLIP-LIT~\cite{ref35} (c) SCI-LOL~\cite{ref34} (d) RUAS-LOL~\cite{ref33} (e) RDHCE~\cite{ref32} (f) Zero-DCE~\cite{ref30} (g) PairLIE~\cite{ref37} (h) ZR-PQR~\cite{ref38} (i) Ours (j) Ground Truth. Please zoom in for a better view.}
\end{figure*}

\begin{figure*}[b]
\centering
 \begin{minipage}[htbp]{0.11111111\linewidth}
\centerline{\includegraphics[width=\textwidth]{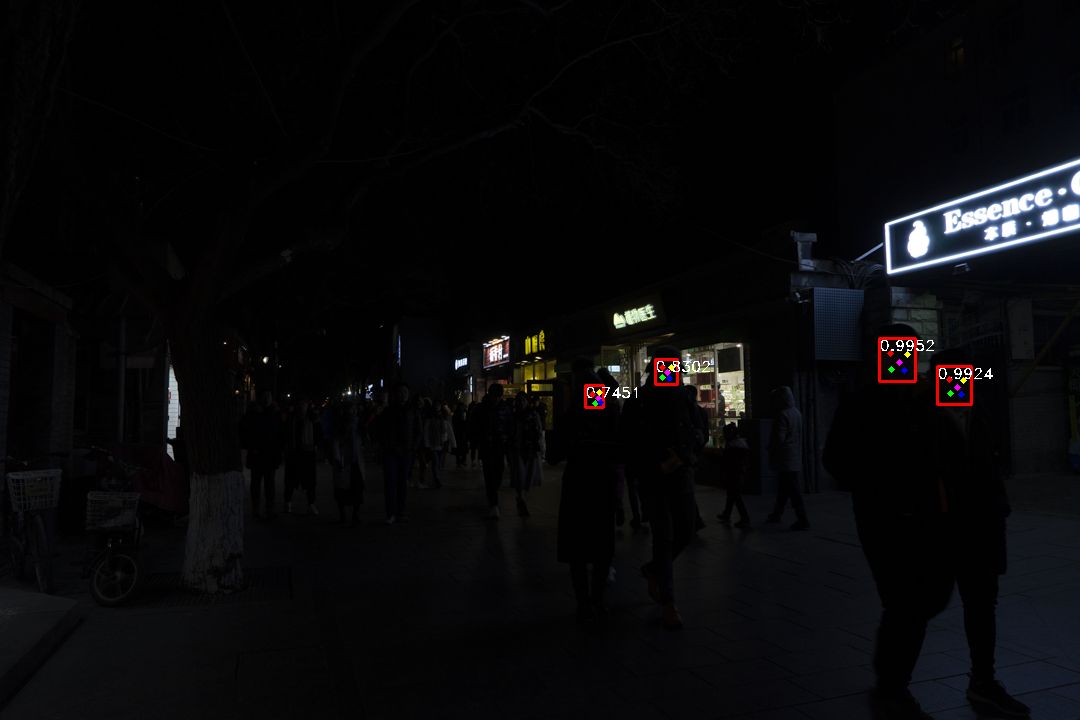}}
\centerline{\includegraphics[width=\textwidth]{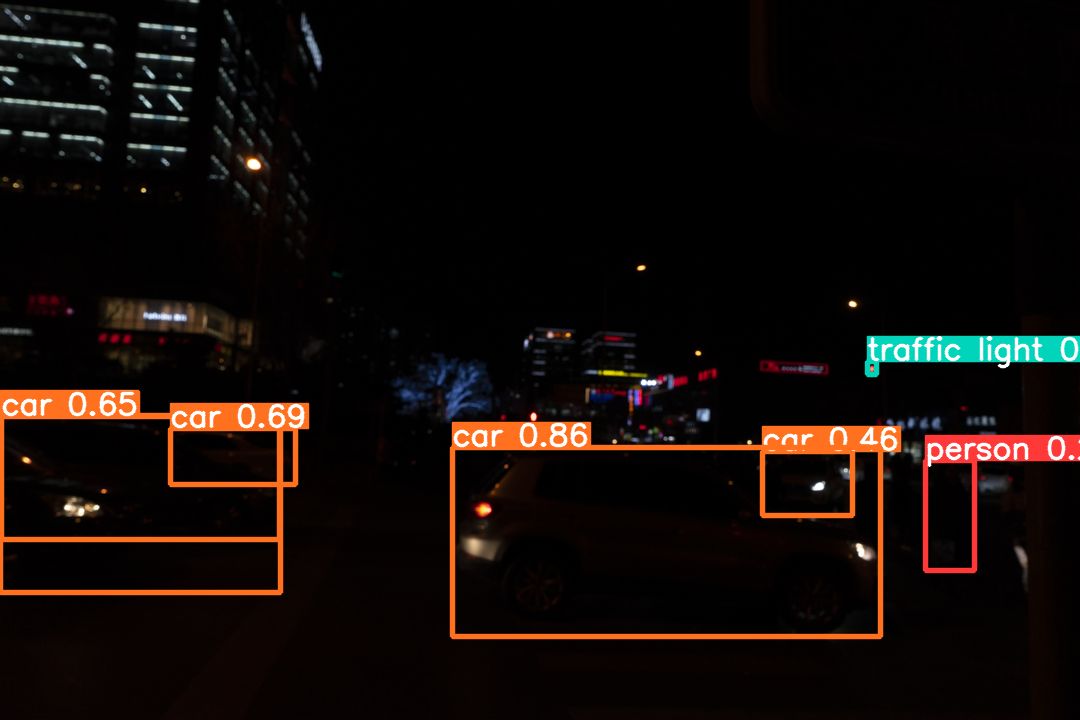}}
\centerline{\includegraphics[width=\textwidth]{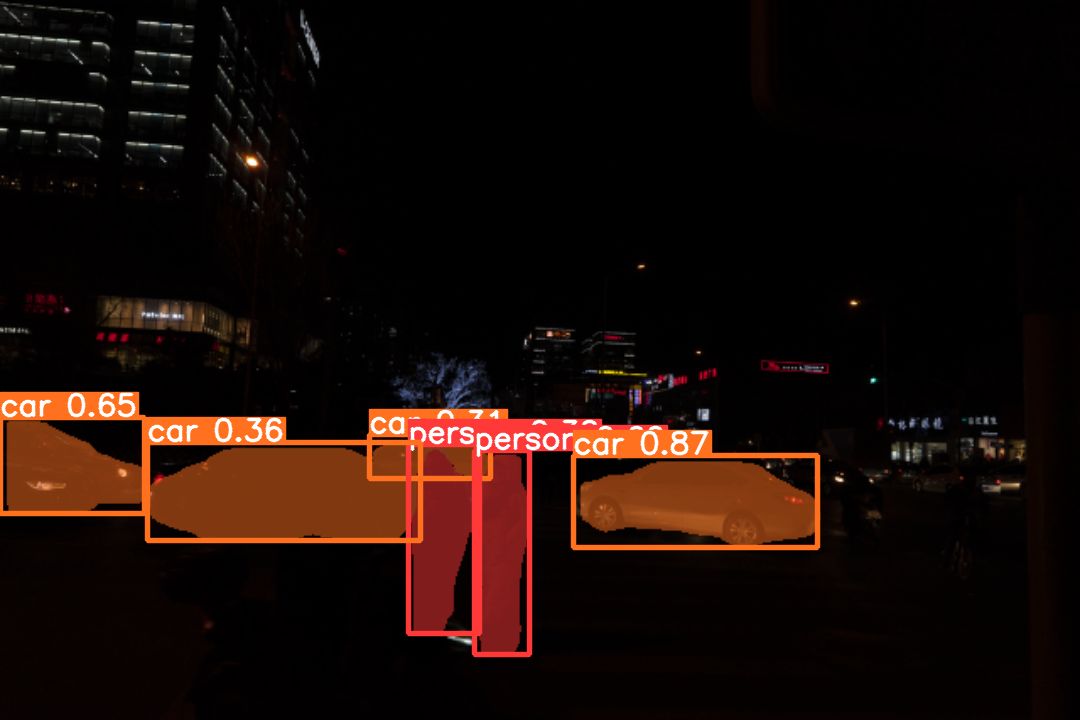}}
\centerline{\scriptsize\textbf{(a)}}
\end{minipage}%
\begin{minipage}[htbp]{0.11111111\linewidth}
\centerline{\includegraphics[width=\textwidth]{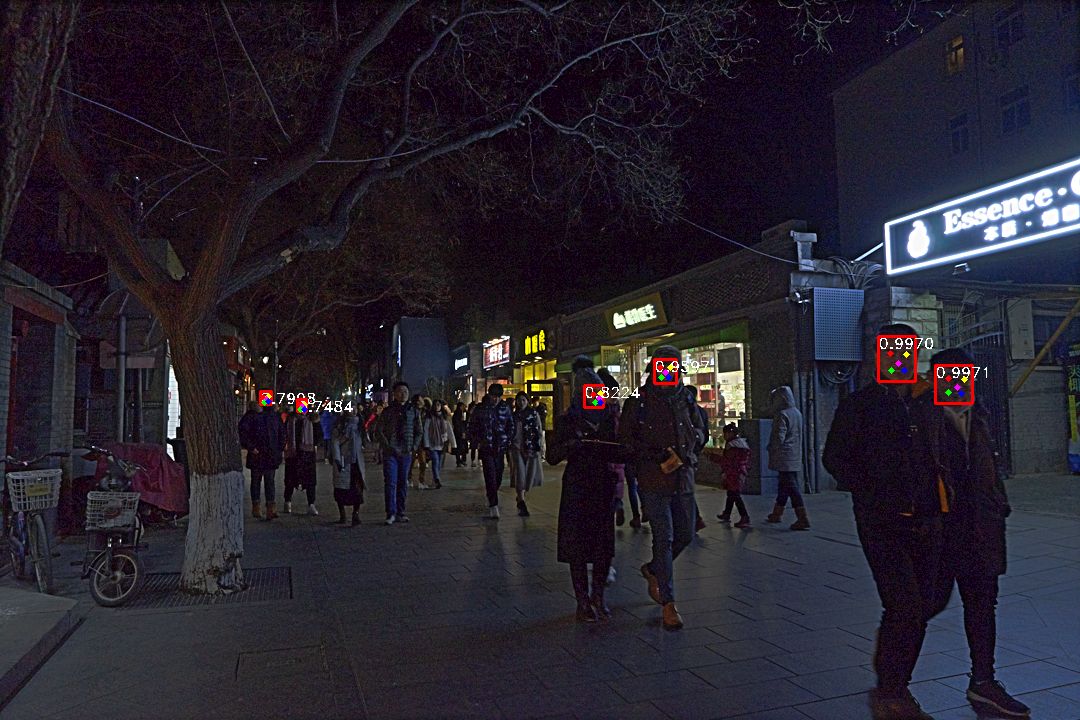}}
\centerline{\includegraphics[width=\textwidth]{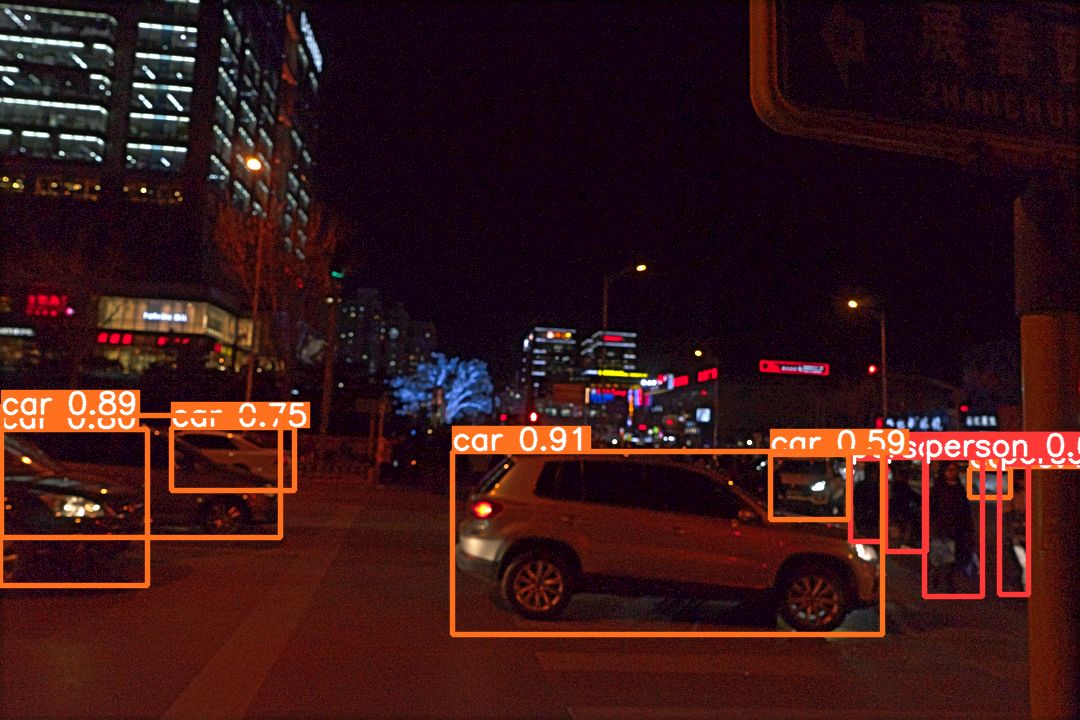}}
\centerline{\includegraphics[width=\textwidth]{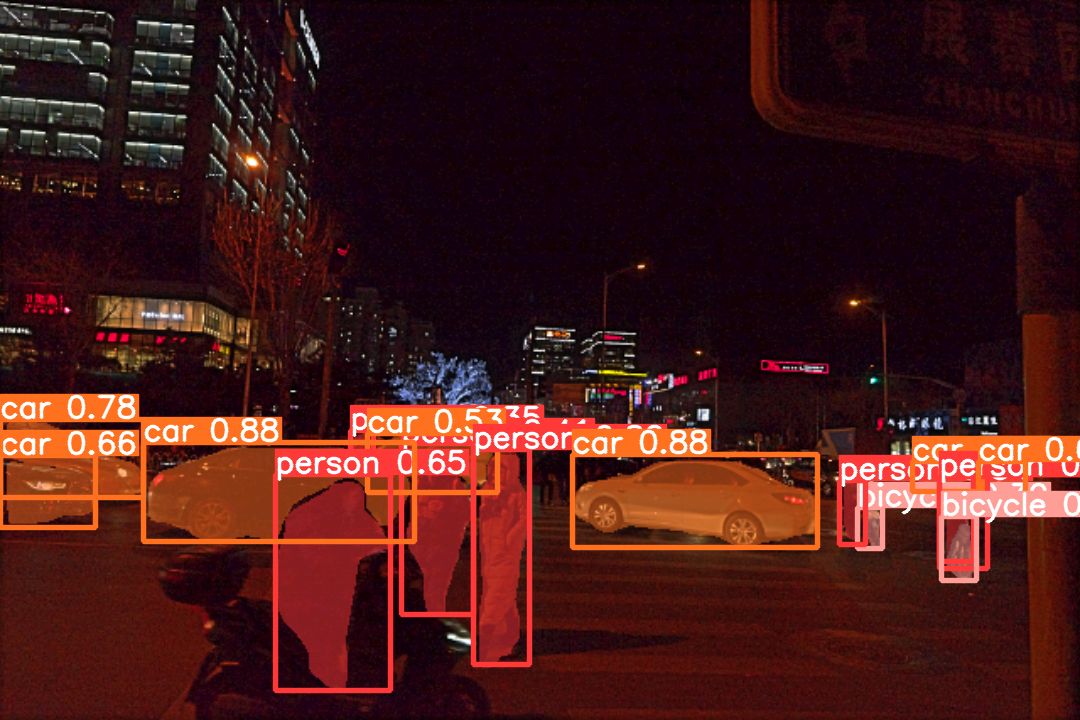}}
\centerline{\scriptsize\textbf{(b)}}
\end{minipage}%
\begin{minipage}[htbp]{0.11111111\linewidth}
\centerline{\includegraphics[width=\textwidth]{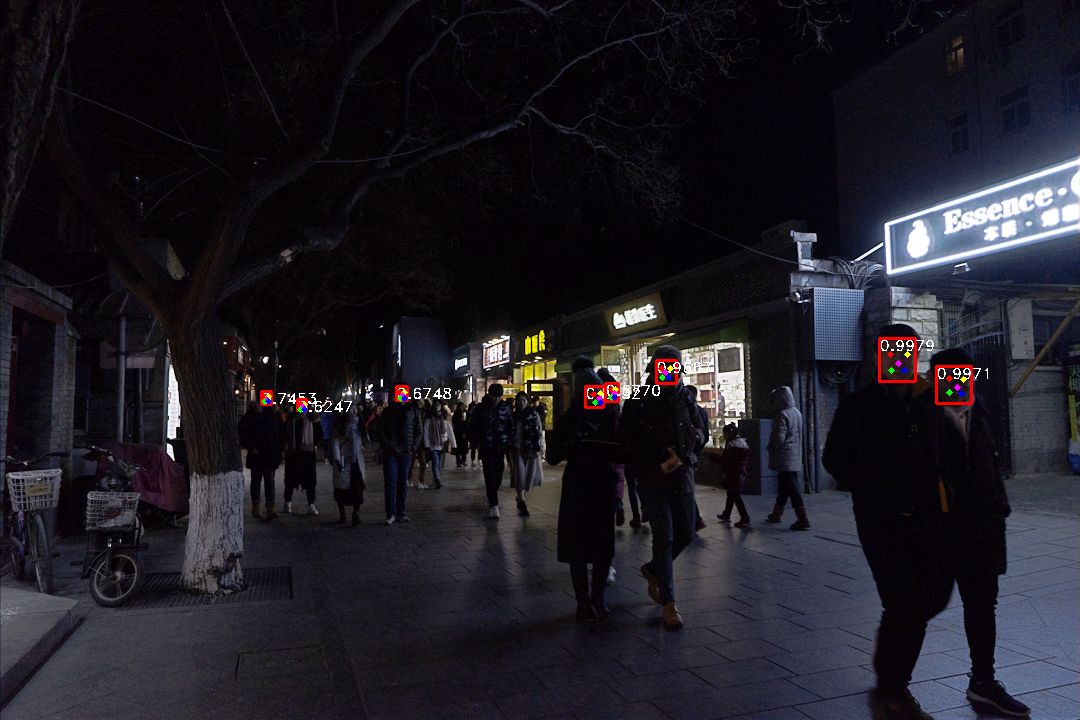}}
\centerline{\includegraphics[width=\textwidth]{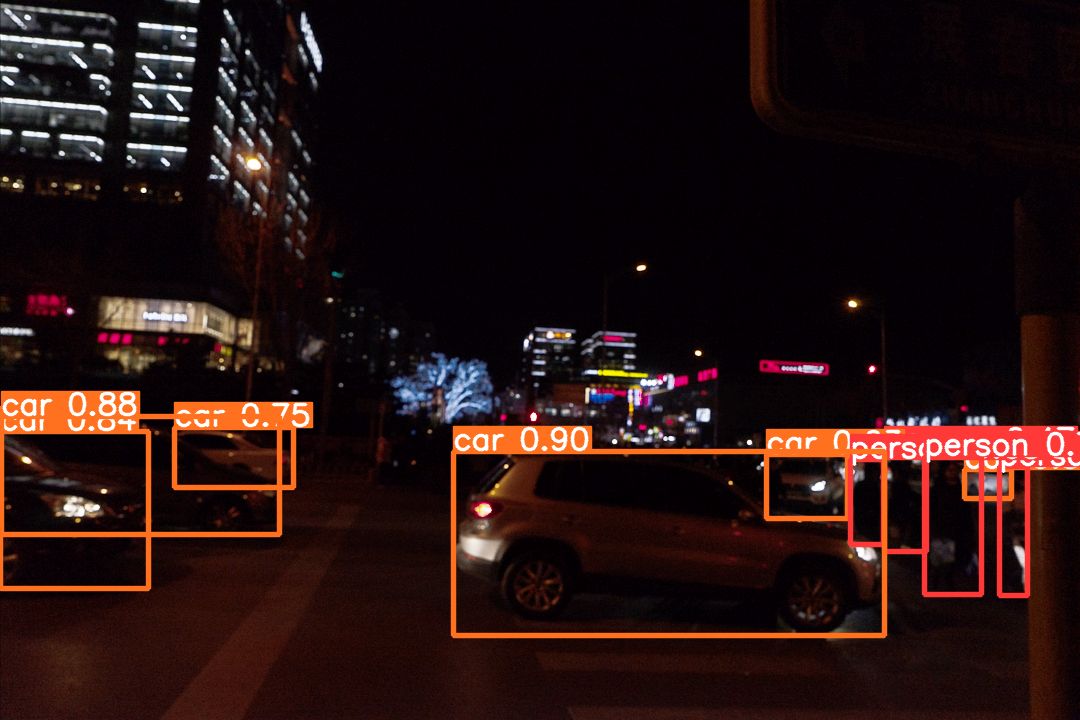}}
\centerline{\includegraphics[width=\textwidth]{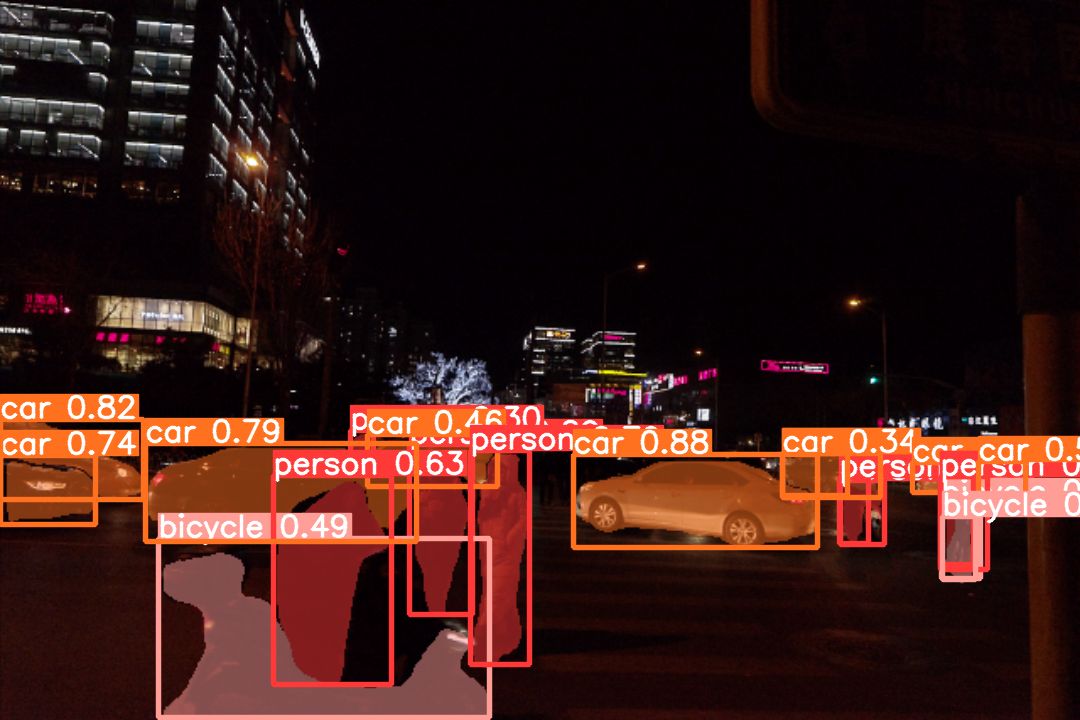}}
\centerline{\scriptsize\textbf{(c)}}
\end{minipage}%
\begin{minipage}[htbp]{0.11111111\linewidth}
\centerline{\includegraphics[width=\textwidth]{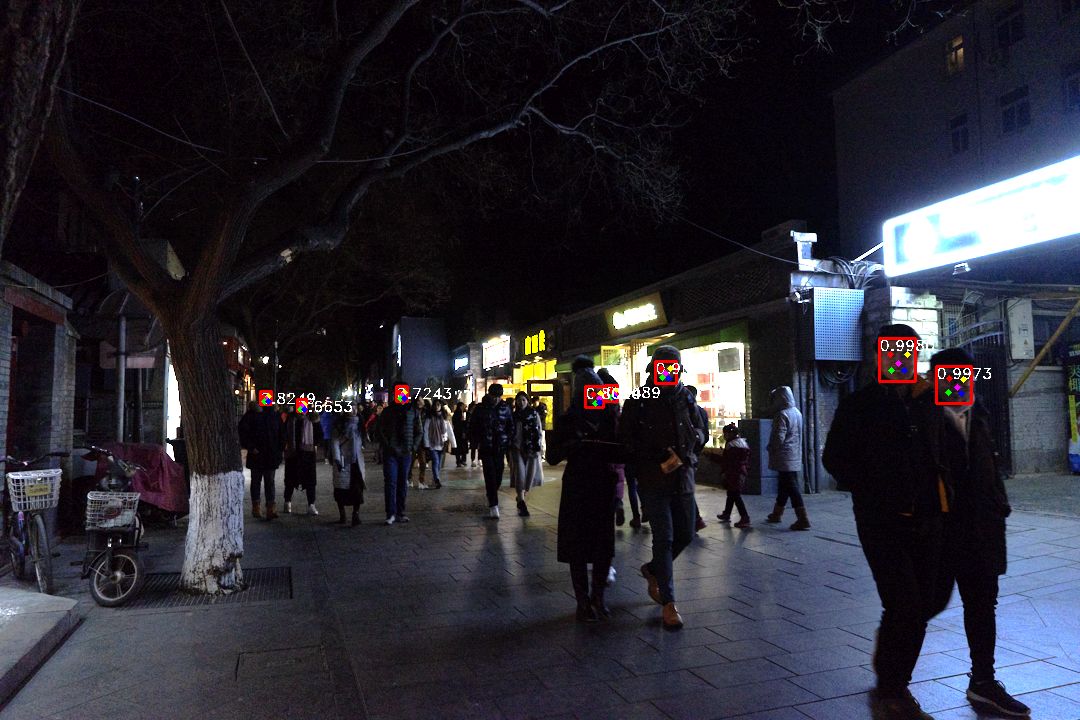}}
\centerline{\includegraphics[width=\textwidth]{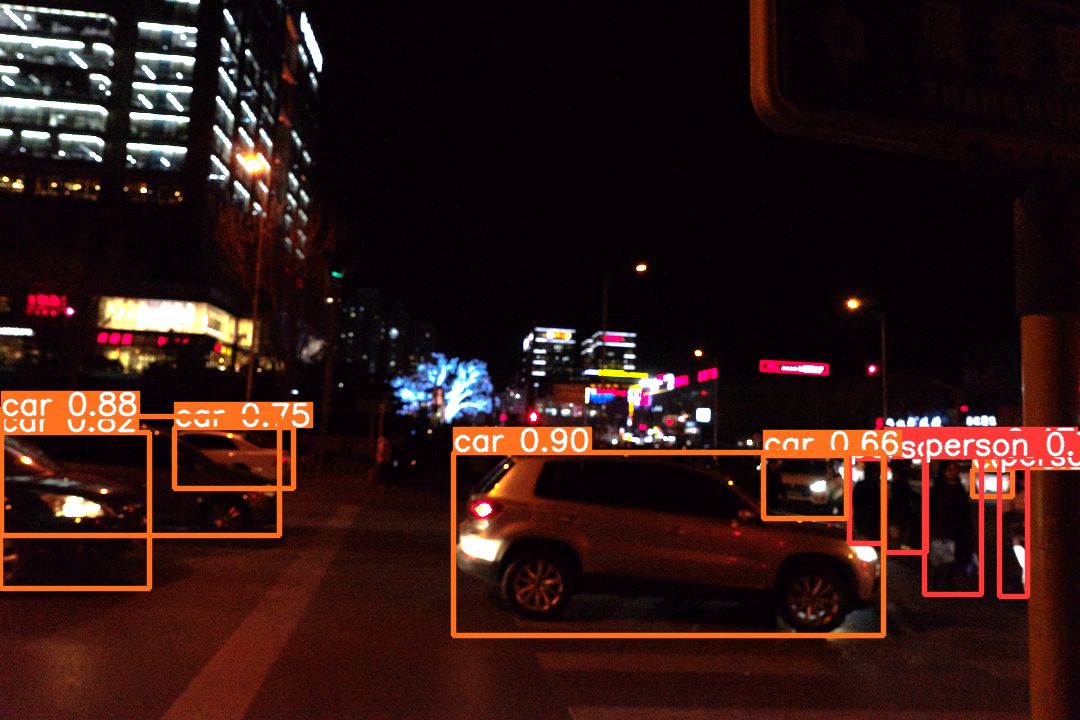}}
\centerline{\includegraphics[width=\textwidth]{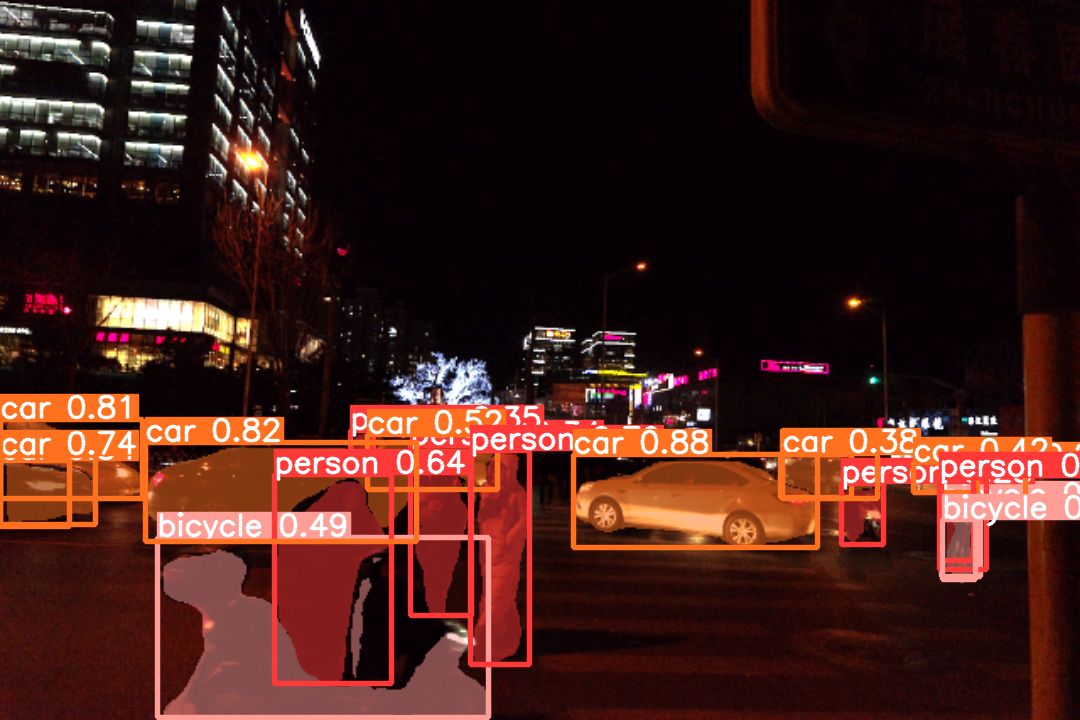}}
\centerline{\scriptsize\textbf{(d)}}
\end{minipage}%
\begin{minipage}[htbp]{0.11111111\linewidth}
\centerline{\includegraphics[width=\textwidth]{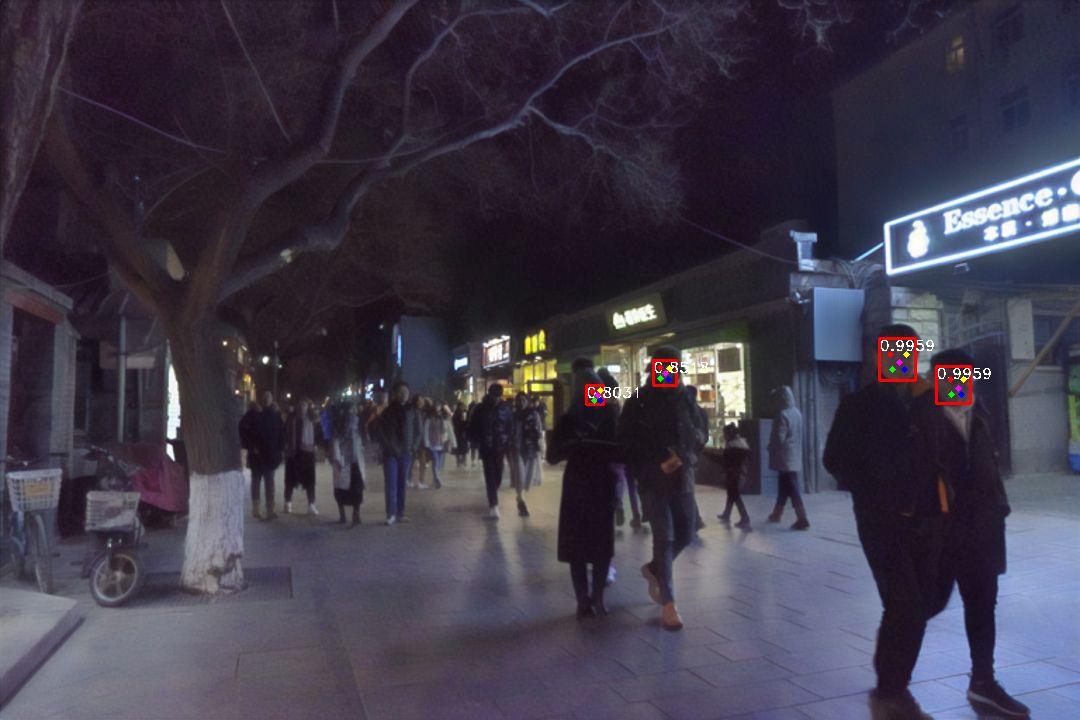}}
\centerline{\includegraphics[width=\textwidth]{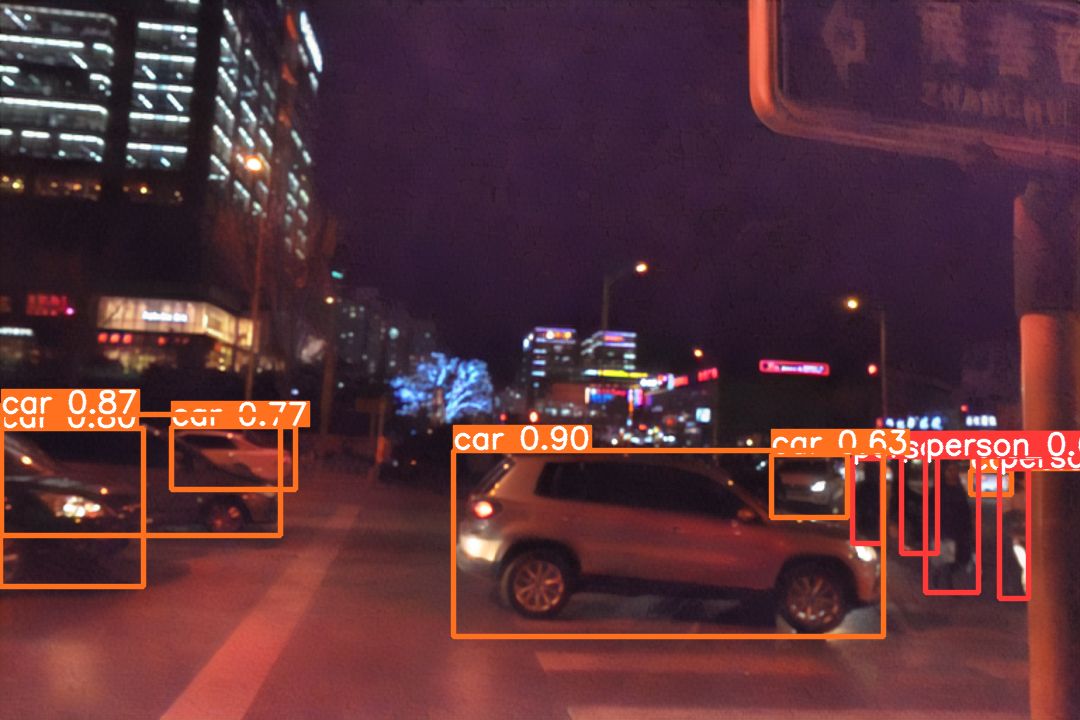}}
\centerline{\includegraphics[width=\textwidth]{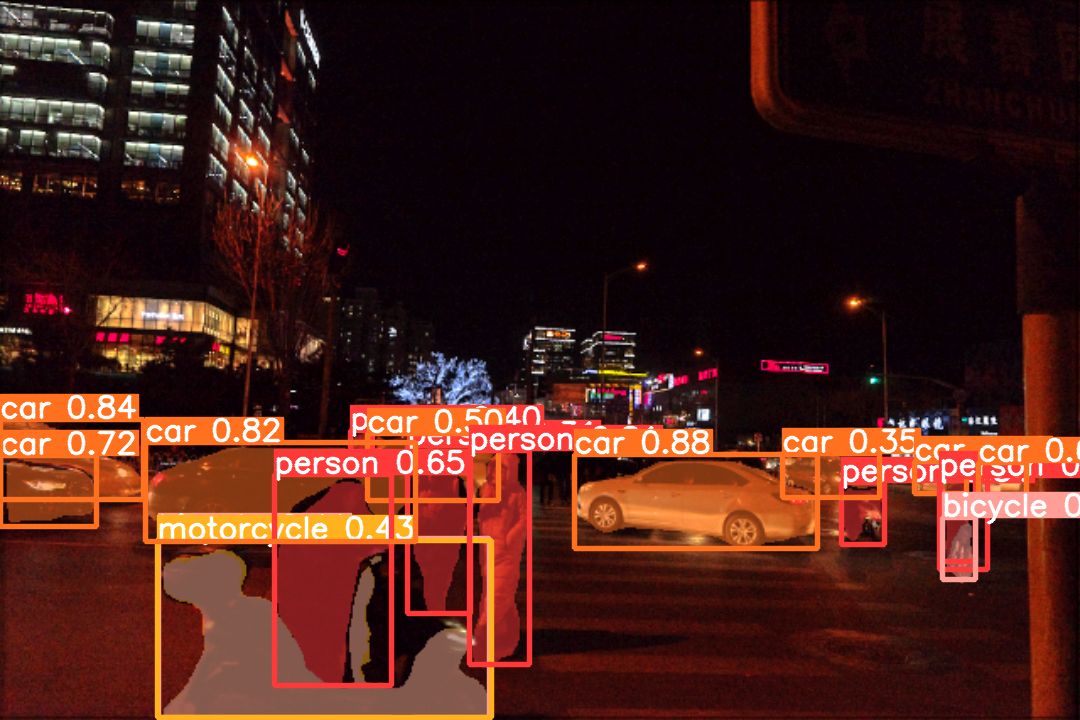}}
\centerline{\scriptsize\textbf{(e)}}
\end{minipage}%
\begin{minipage}[htbp]{0.11111111\linewidth}
\centerline{\includegraphics[width=\textwidth]{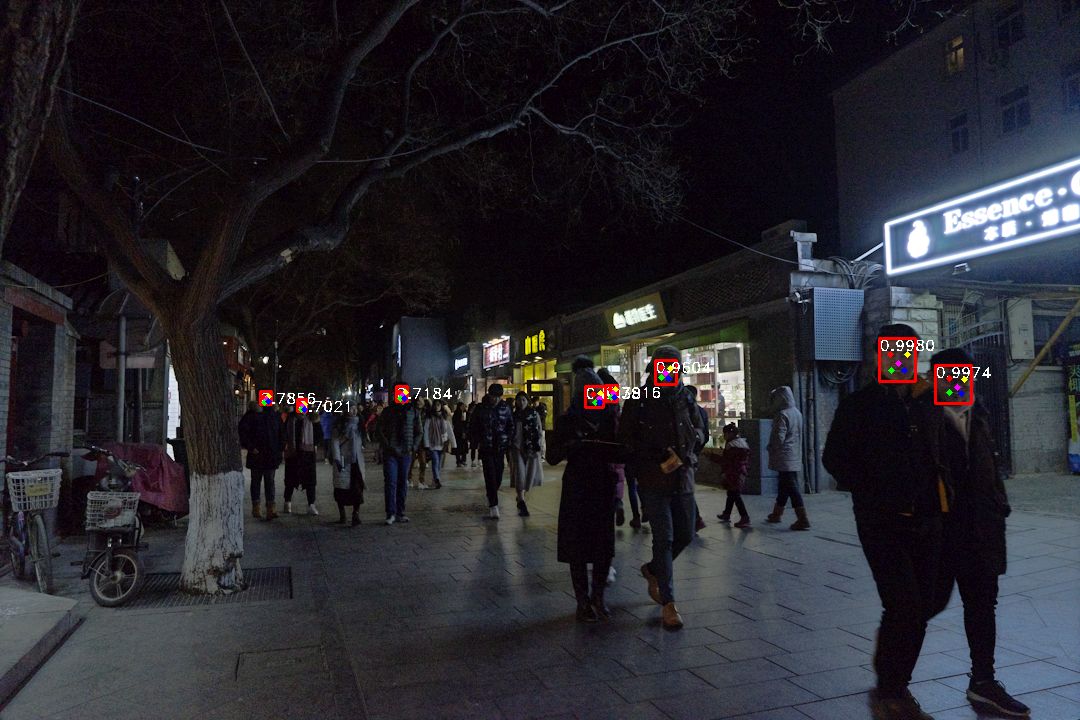}}
\centerline{\includegraphics[width=\textwidth]{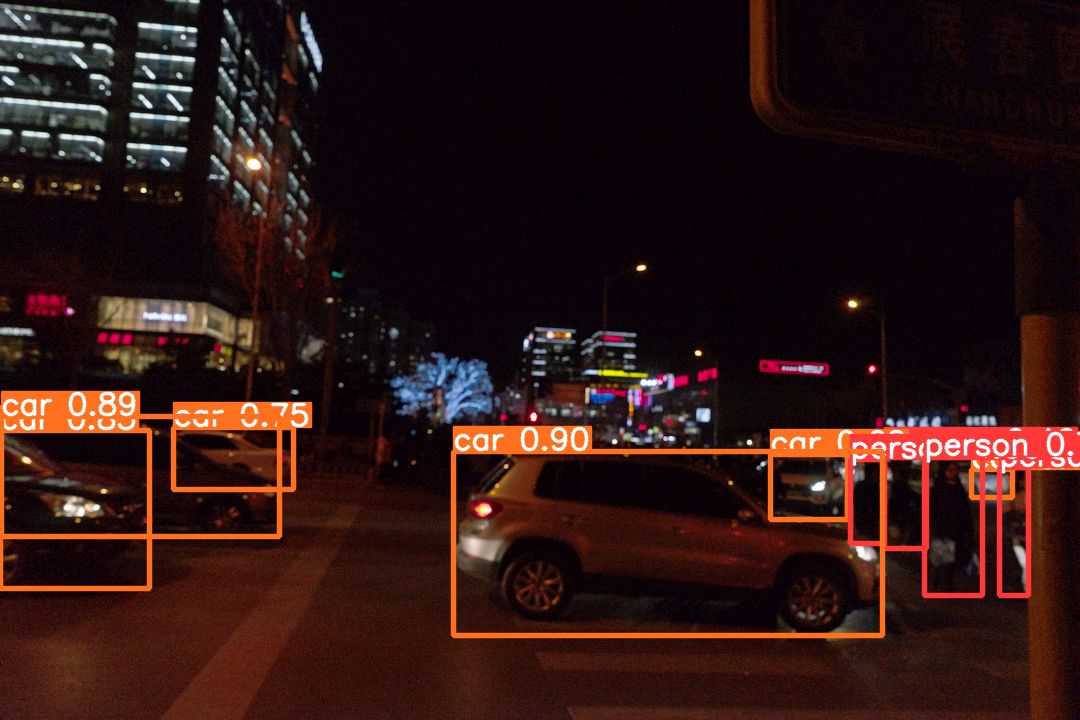}}
\centerline{\includegraphics[width=\textwidth]{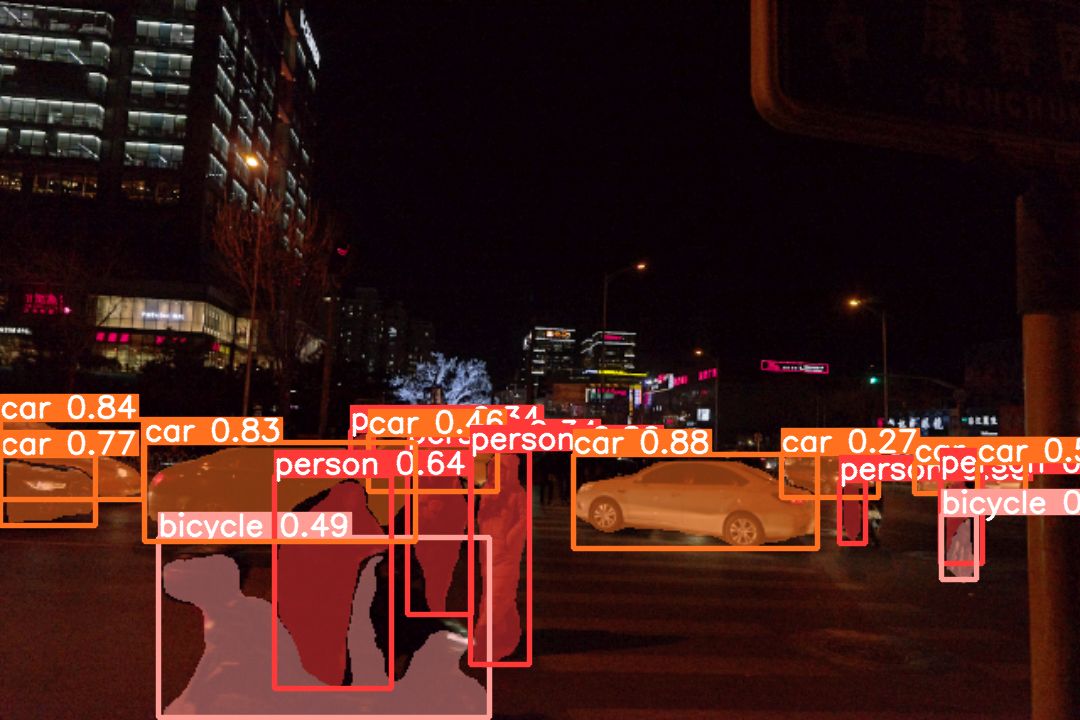}}
\centerline{\scriptsize\textbf{(f)}}
\end{minipage}%
\begin{minipage}[htbp]{0.11111111\linewidth}
\centerline{\includegraphics[width=\textwidth]{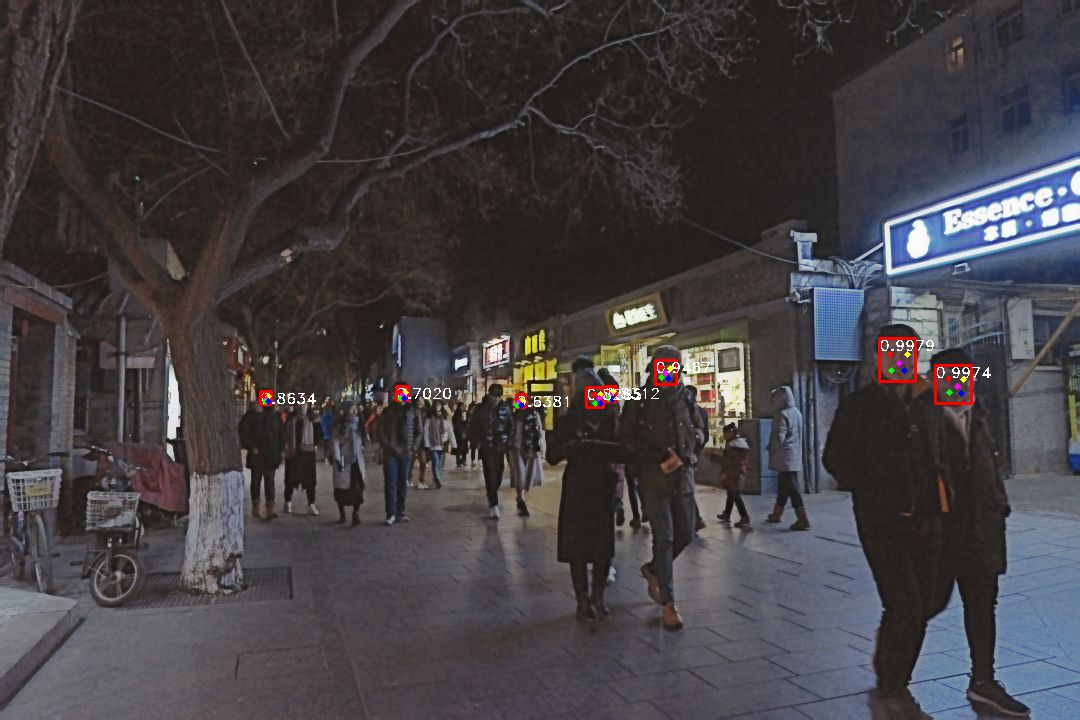}}
\centerline{\includegraphics[width=\textwidth]{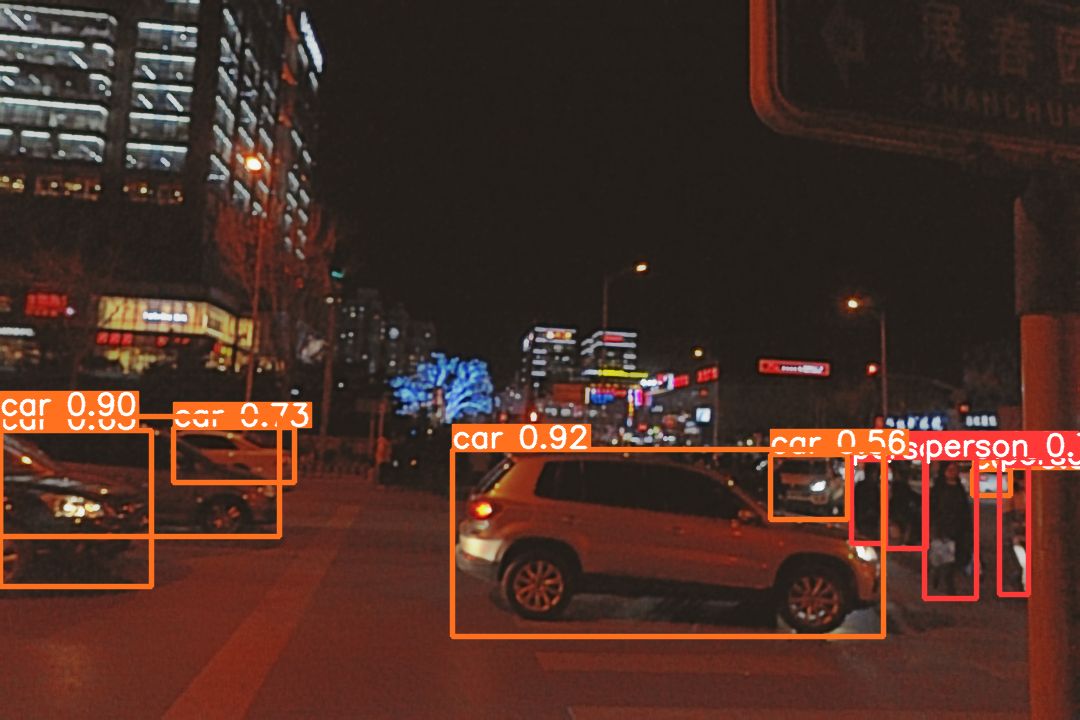}}
\centerline{\includegraphics[width=\textwidth]{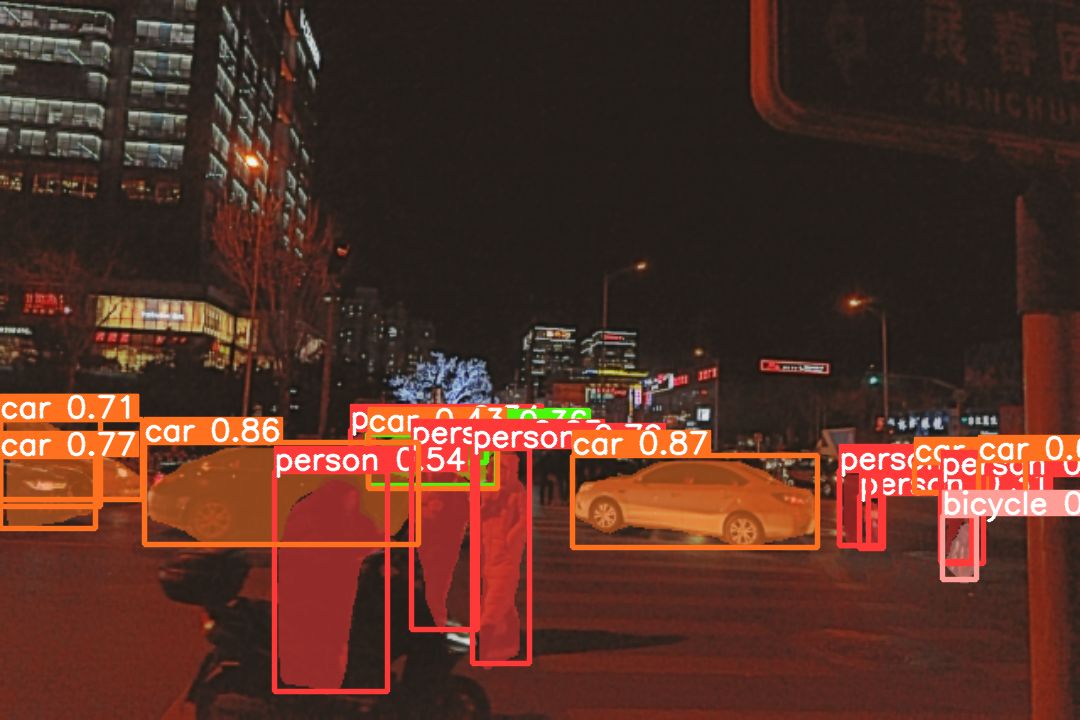}}
\centerline{\scriptsize\textbf{(g)}}
\end{minipage}%
\begin{minipage}[htbp]{0.11111111\linewidth}
\centerline{\includegraphics[width=\textwidth]{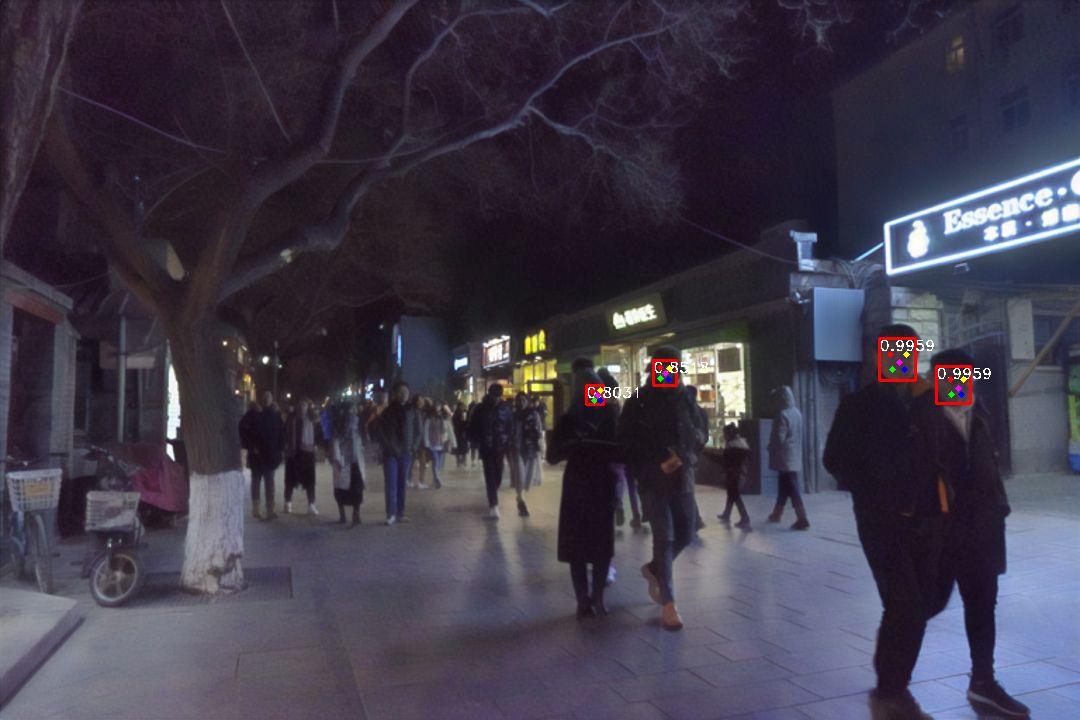}}
\centerline{\includegraphics[width=\textwidth]{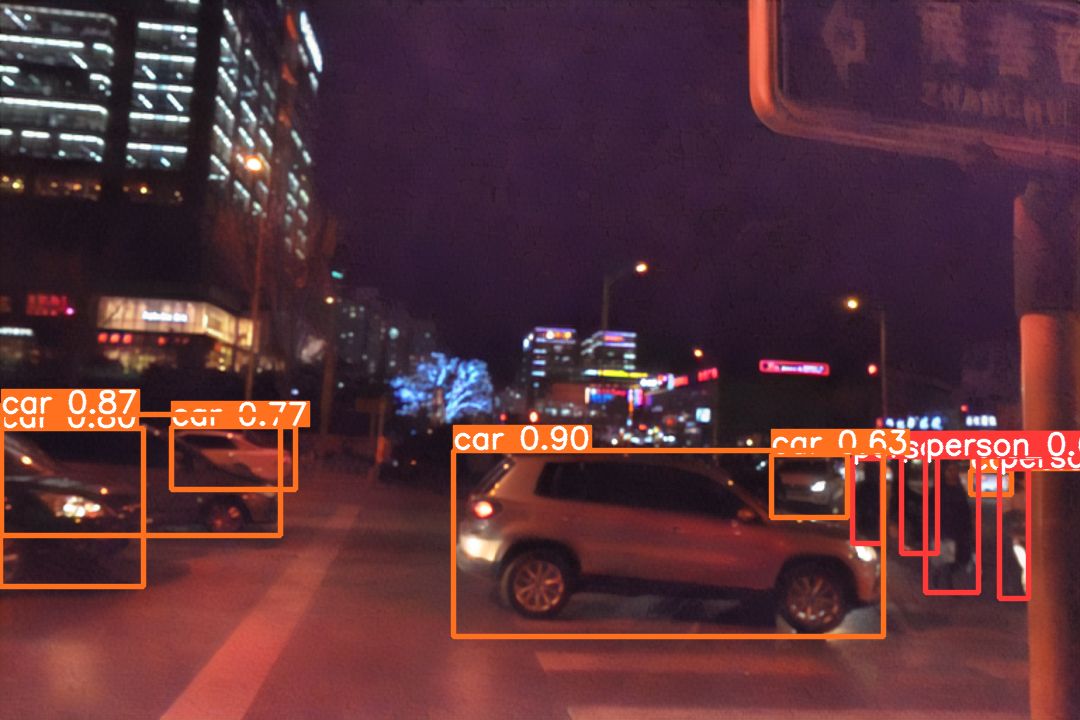}}
\centerline{\includegraphics[width=\textwidth]{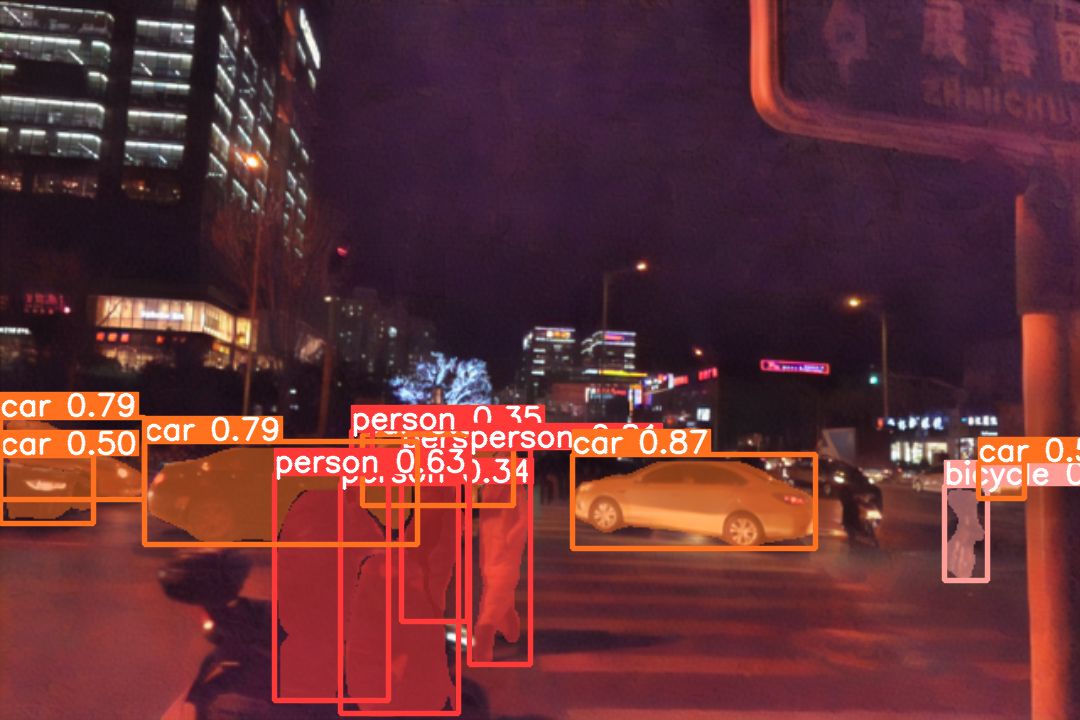}}
\centerline{\scriptsize\textbf{(h)}}
\end{minipage}%
\begin{minipage}[htbp]{0.11111111\linewidth}
\centerline{\includegraphics[width=\textwidth]{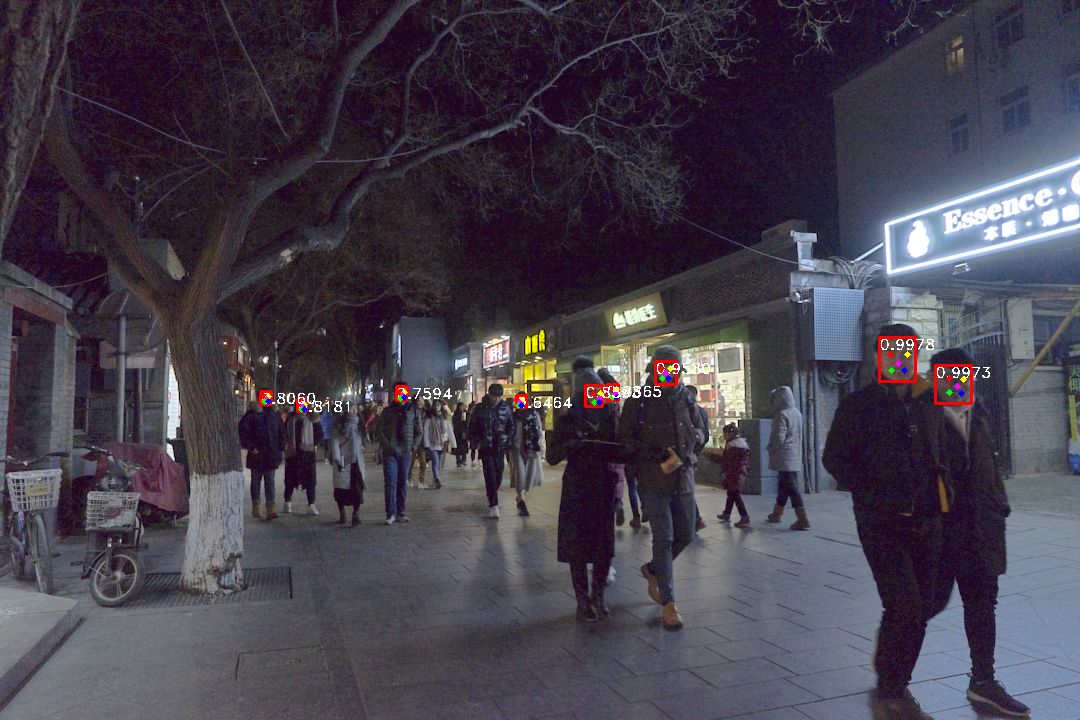}}
\centerline{\includegraphics[width=\textwidth]{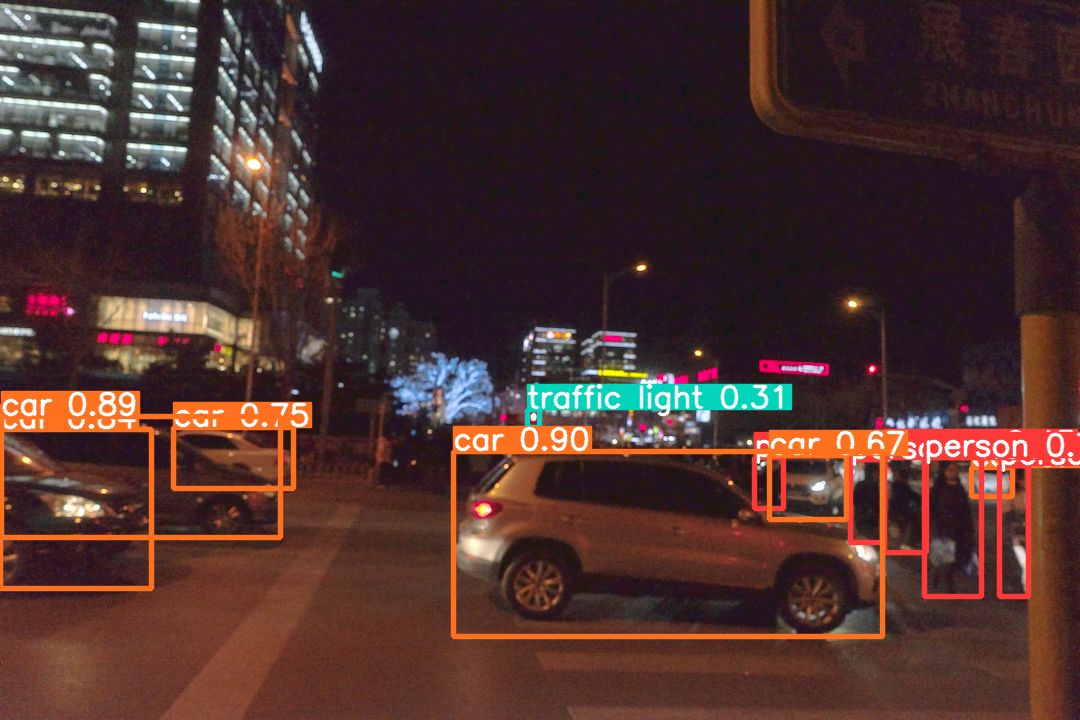}}
\centerline{\includegraphics[width=\textwidth]{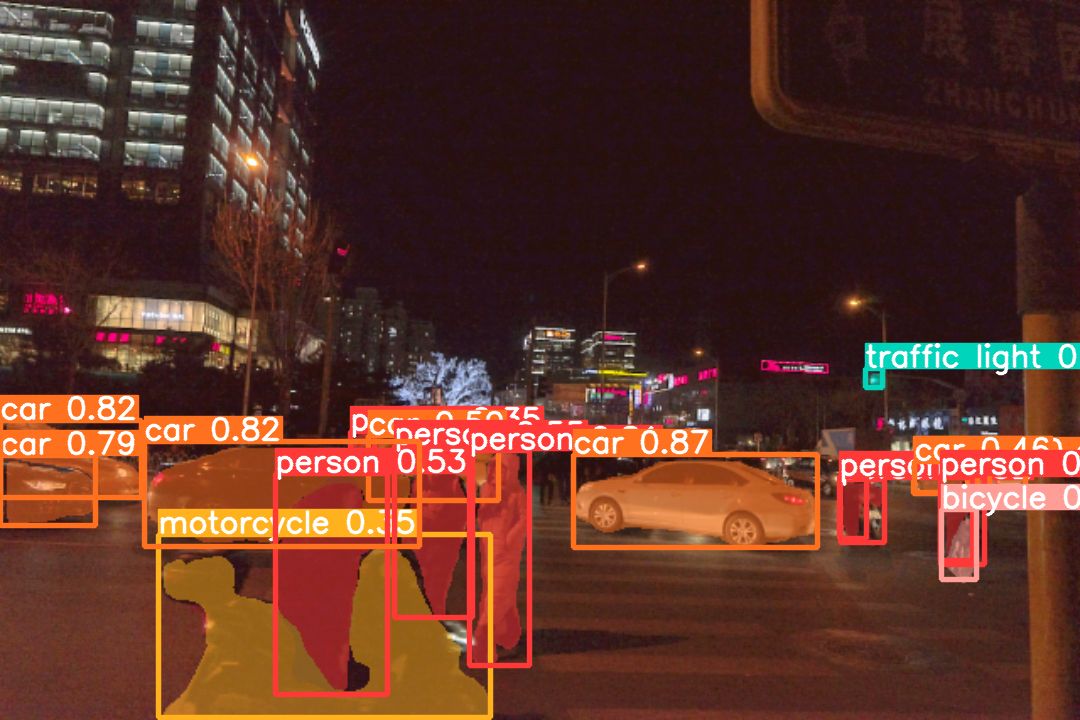}}
\centerline{\scriptsize\textbf{(i)}}
\end{minipage}%
\caption{\label{fig3}Visualization comparison examples of higher-level tasks on Dark Face. (a) Input (b) CLIP-LIT~\cite{ref35} (c) SCI-LOL~\cite{ref34} (d) RUAS-LOL~\cite{ref33} (e) RDHCE~\cite{ref32} (f) Zero-DCE~\cite{ref30} (g) PairLIE~\cite{ref37} (h) ZR-PQR~\cite{ref38} (i) Ours. Please zoom in for a better view.}
\end{figure*}

\begin{figure*}[htbp]
\centering
\begin{minipage}[htbp]{0.32\linewidth}
\centerline{\includegraphics[width=\textwidth]{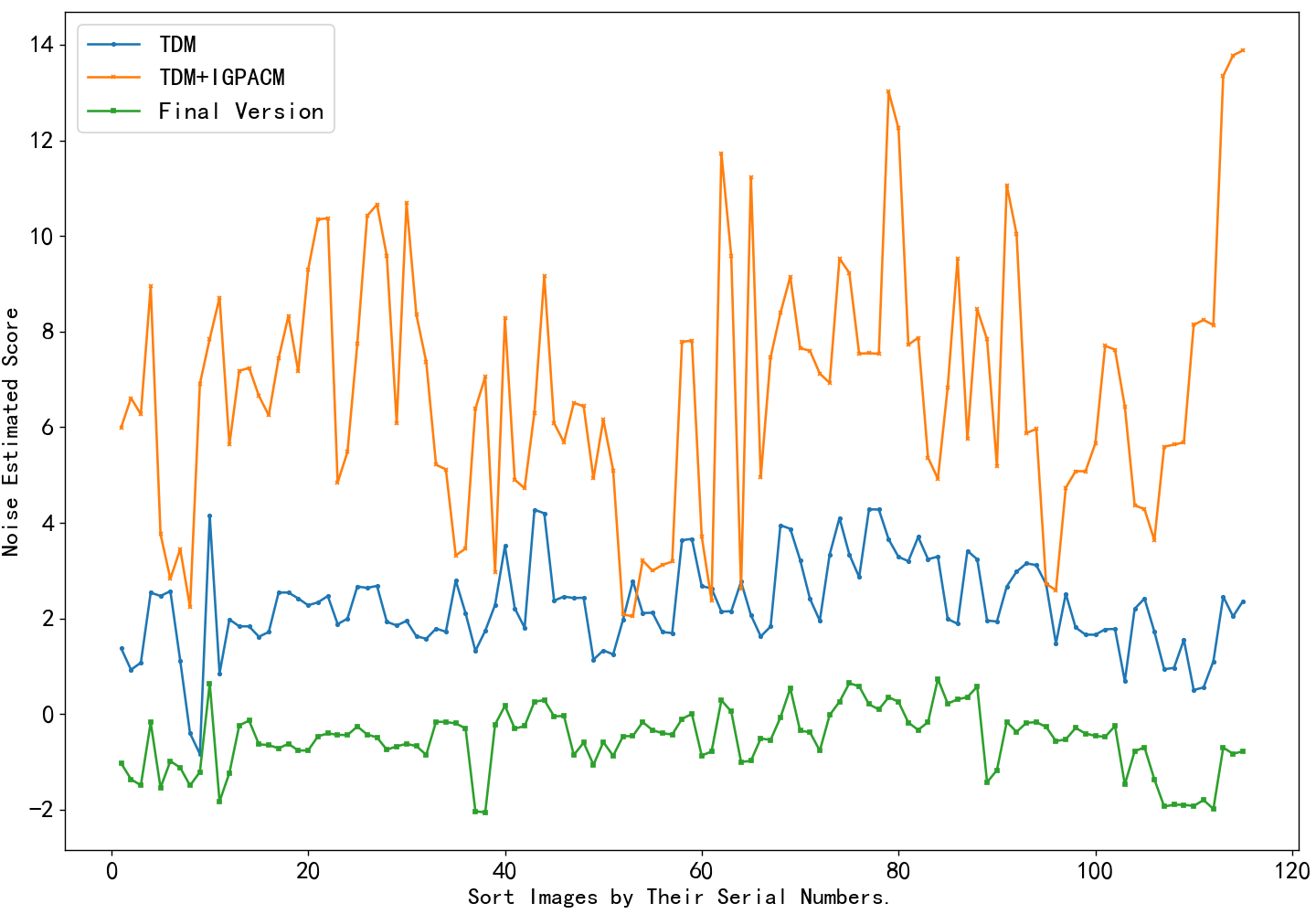}}
\centerline{\scriptsize\textbf{(a). LOLv1+LOLv2-Real}}
\end{minipage}%
\hspace{0.01\linewidth} 
\begin{minipage}[htbp]{0.32\linewidth}
\centerline{\includegraphics[width=\textwidth]{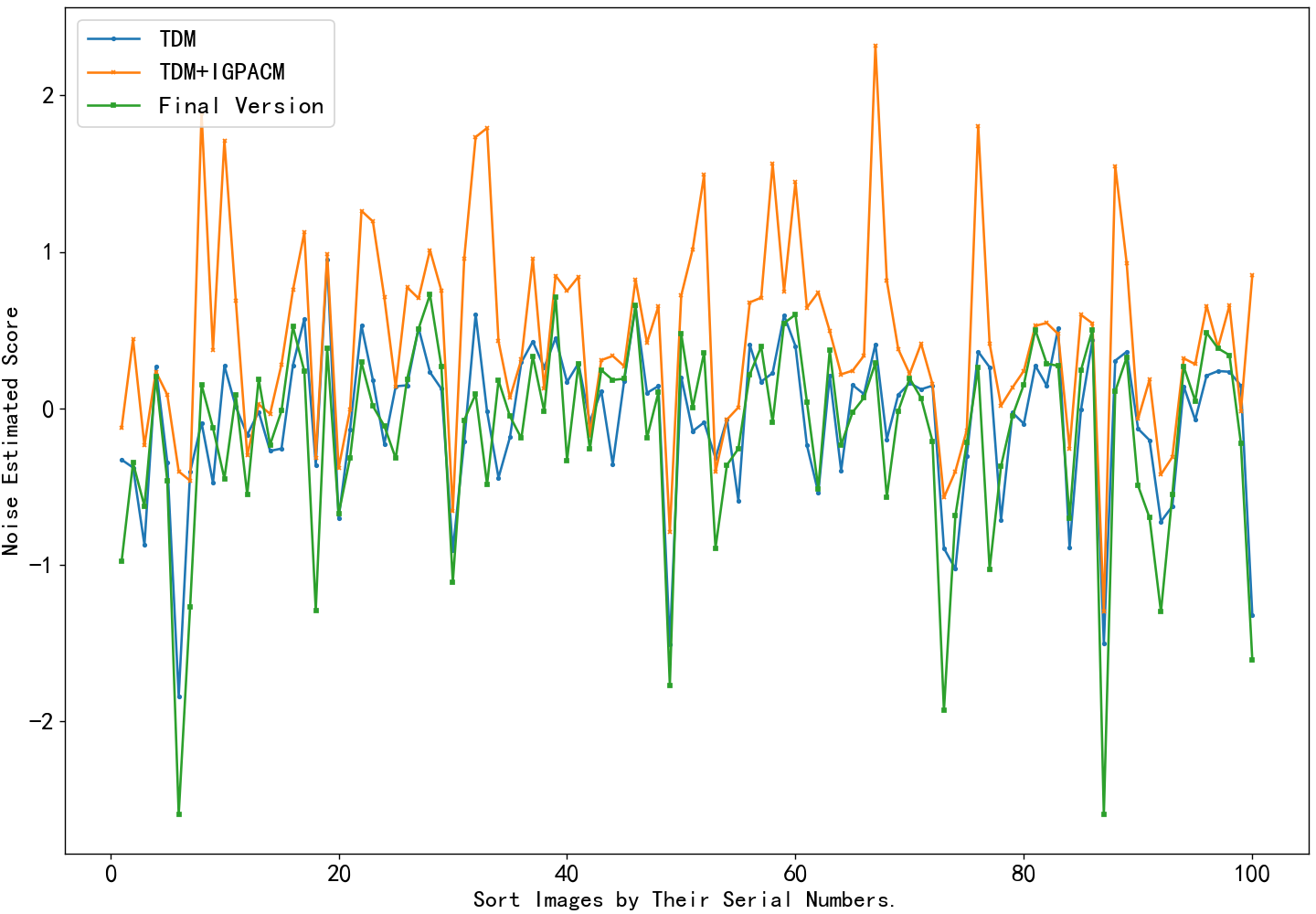}}
\centerline{\scriptsize\textbf{(b). LOLv2-Synthetic}}
\end{minipage}%
\hspace{0.01\linewidth} 
\begin{minipage}[htbp]{0.32\linewidth}
\centerline{\includegraphics[width=\textwidth]{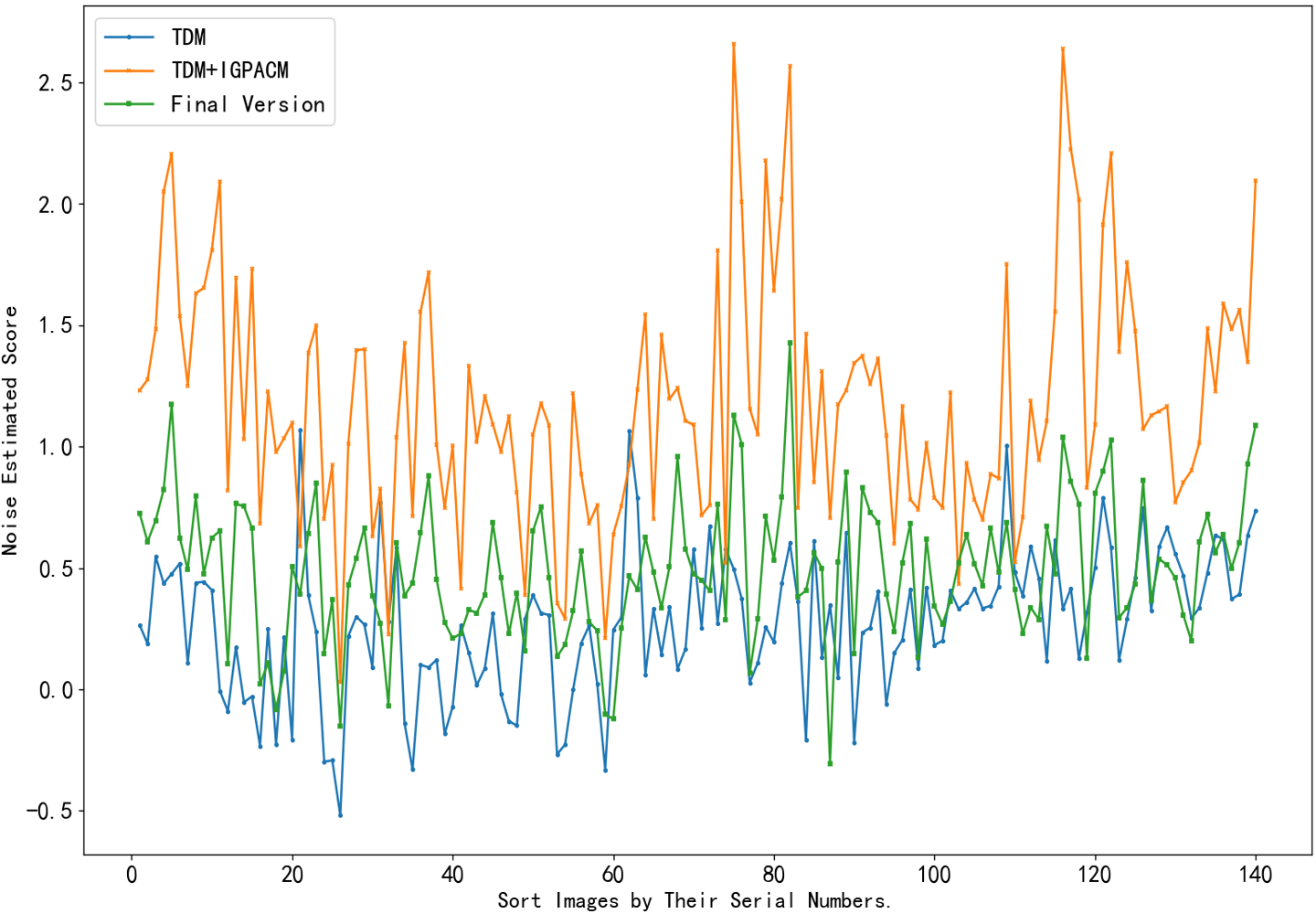}}
\centerline{\scriptsize\textbf{(c). NTIRE LLE}}
\end{minipage}
\caption{\label{fig4}Comparing $NES$ among diverse components on datasets. Where $NES$ = $NES_{E}-NES_{GT}$; $NES_{E}$ and $NES_{GT}$ represent the noise estimated score of the enhanced image and Ground Truth, respectively. Please zoom in for a better view.}
\end{figure*}

\begin{figure*}[htbp]
\centering
\begin{minipage}[htbp]{0.19\linewidth}
\centerline{\includegraphics[width=\textwidth]{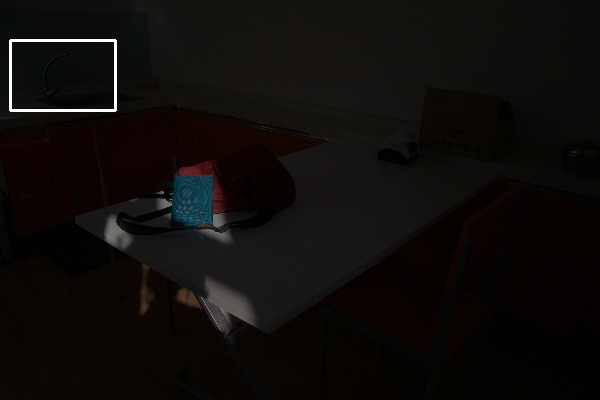}}
\vspace{2pt}
\centerline{\includegraphics[width=\textwidth]{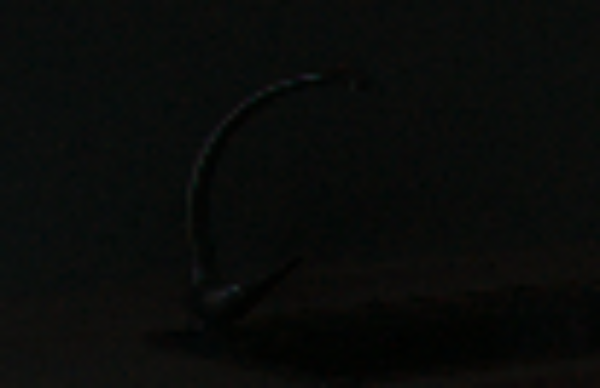}}
\centerline{\scriptsize\textbf{(a). NES: 1.1134}} 
\centerline{\includegraphics[width=\textwidth]{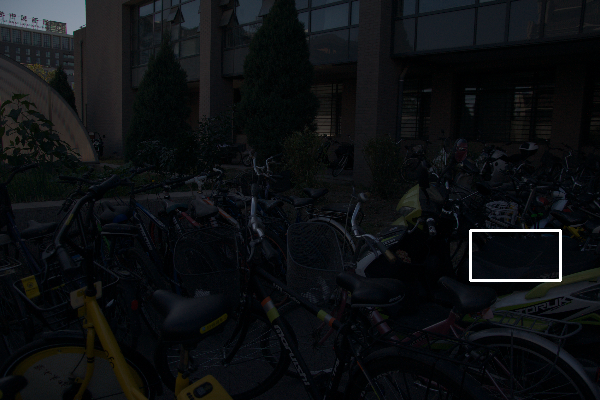}}
\vspace{2pt}
\centerline{\includegraphics[width=\textwidth]{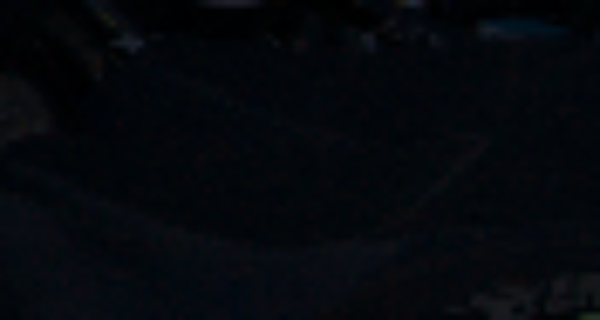}}
\centerline{\scriptsize\textbf{(a). NES: 2.0772}}
\end{minipage}%
\hspace{1pt}
\begin{minipage}[htbp]{0.19\linewidth}
\centerline{\includegraphics[width=\textwidth]{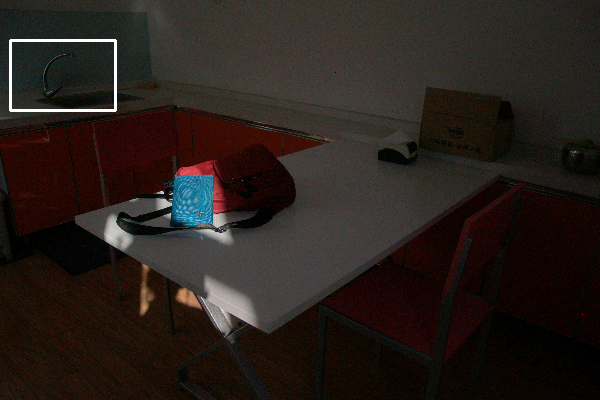}}
\vspace{2pt}
\centerline{\includegraphics[width=\textwidth]{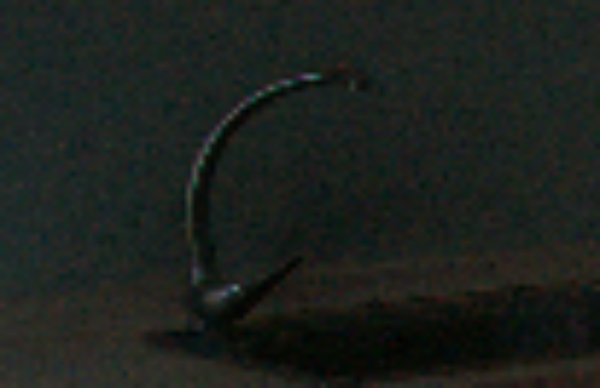}}
\centerline{\scriptsize\textbf{(b). NES: 3.2296}} 
\centerline{\includegraphics[width=\textwidth]{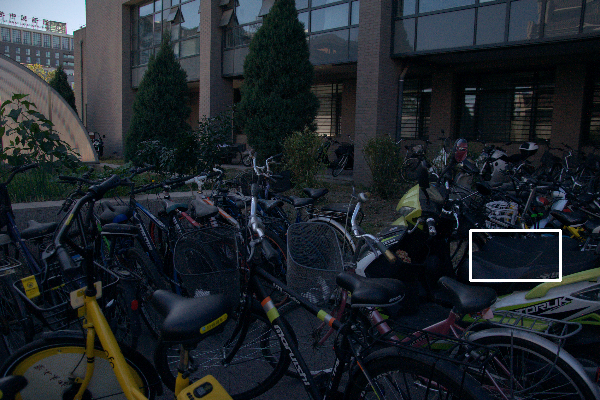}}
\vspace{2pt}
\centerline{\includegraphics[width=\textwidth]{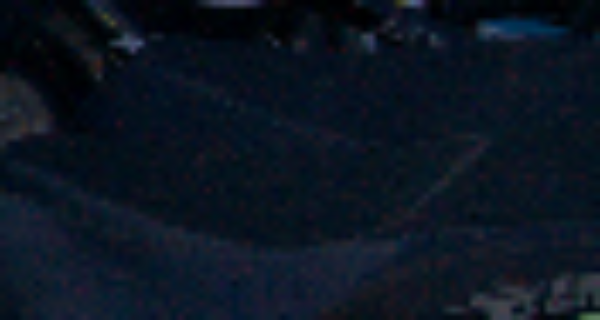}}
\centerline{\scriptsize\textbf{(b). NES: 5.0932}} 
\end{minipage}%
\hspace{1pt}
\begin{minipage}[htbp]{0.19\linewidth}
\centerline{\includegraphics[width=\textwidth]{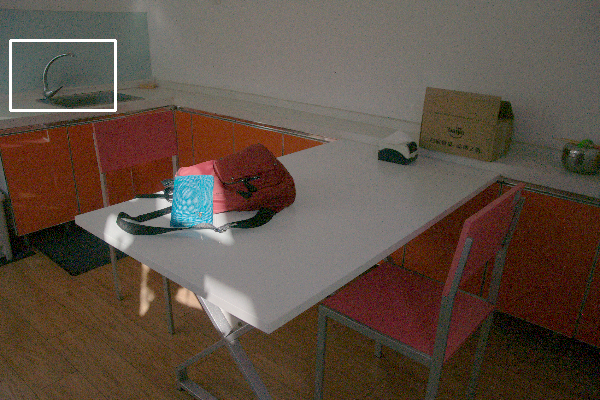}}
\vspace{2pt}
\centerline{\includegraphics[width=\textwidth]{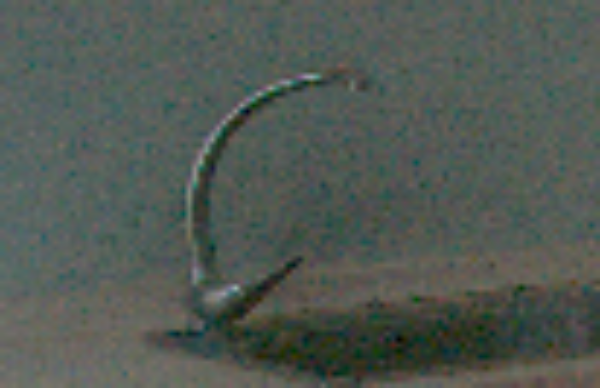}}
\centerline{\scriptsize\textbf{(c). NES: 9.9532}} 
\centerline{\includegraphics[width=\textwidth]{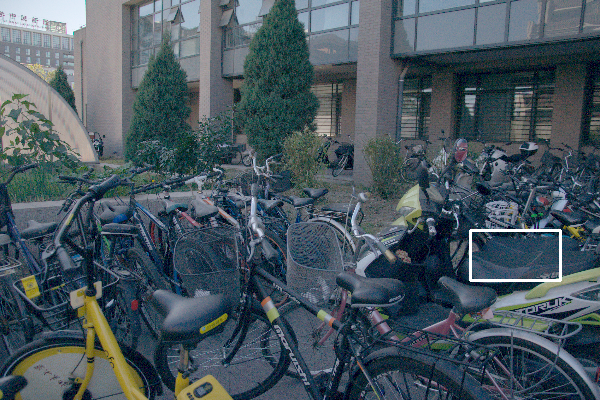}}
\vspace{2pt}
\centerline{\includegraphics[width=\textwidth]{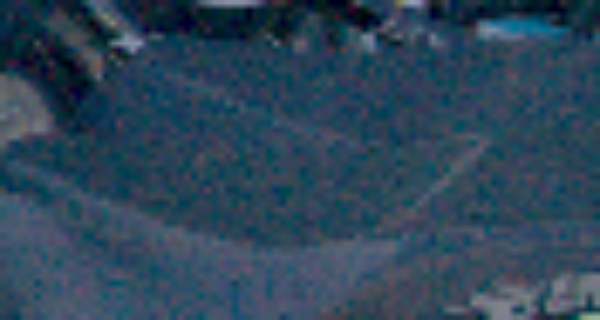}}
\centerline{\scriptsize\textbf{(c). NES: 10.1598}}
\end{minipage}%
\hspace{1pt}
\begin{minipage}[htbp]{0.19\linewidth}
\centerline{\includegraphics[width=\textwidth]{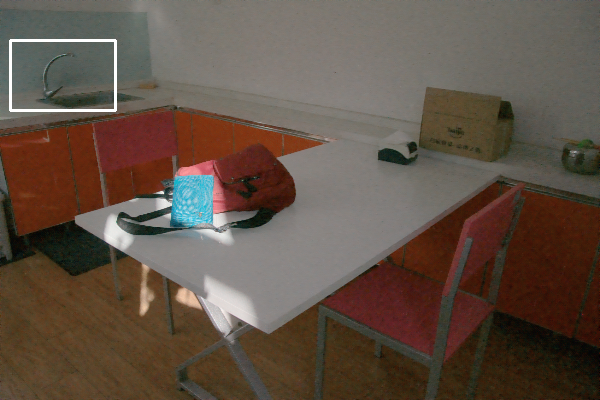}}
\vspace{2pt}
\centerline{\includegraphics[width=\textwidth]{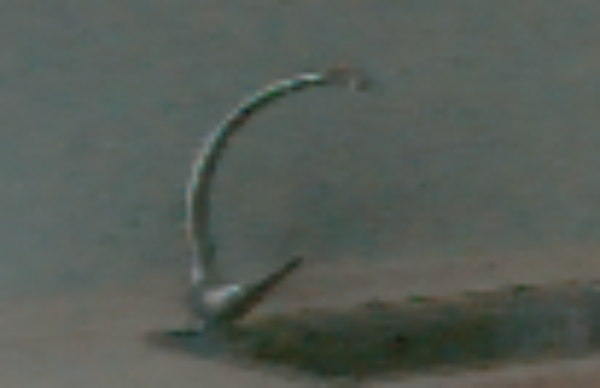}}
\centerline{\scriptsize\textbf{\textcolor{red}{(d). NES: 0.9325}}} 
\centerline{\includegraphics[width=\textwidth]{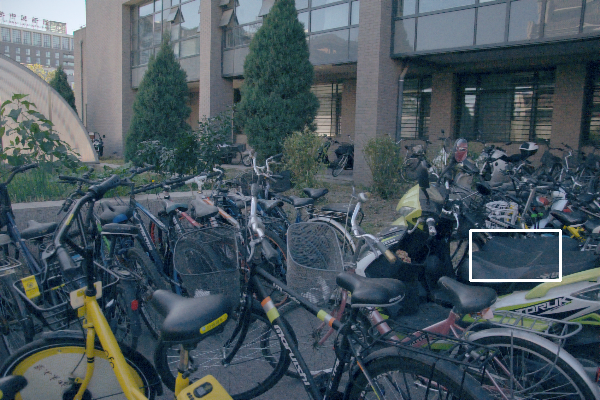}}
\vspace{2pt}
\centerline{\includegraphics[width=\textwidth]{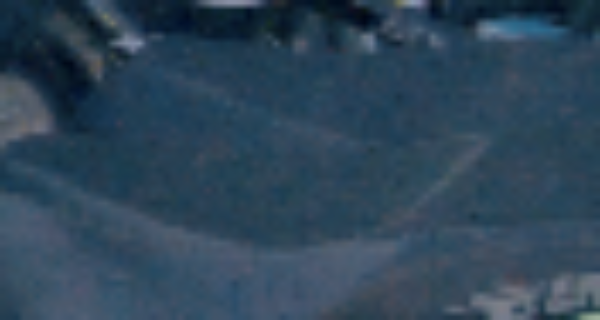}}
\centerline{\scriptsize\textbf{\textcolor{red}{(d). NES: 1.7380}}}
\end{minipage}%
\hspace{1pt}
\begin{minipage}[htbp]{0.19\linewidth}
\centerline{\includegraphics[width=\textwidth]{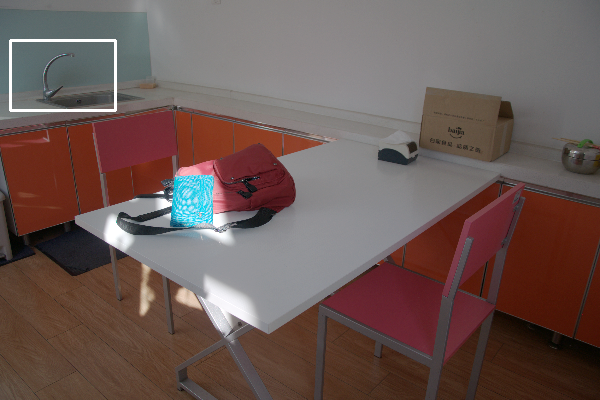}}
\vspace{2pt}
\centerline{\includegraphics[width=\textwidth]{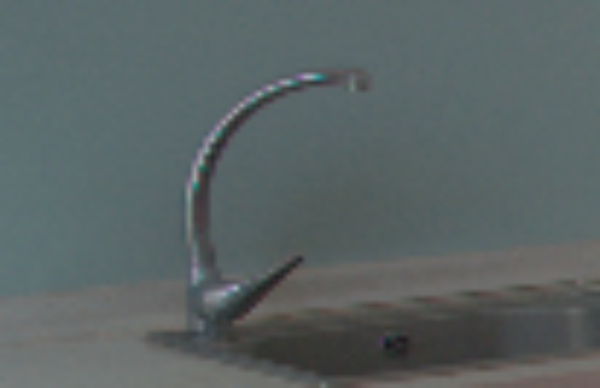}}
\centerline{\scriptsize\textbf{(e)}}
\centerline{\includegraphics[width=\textwidth]{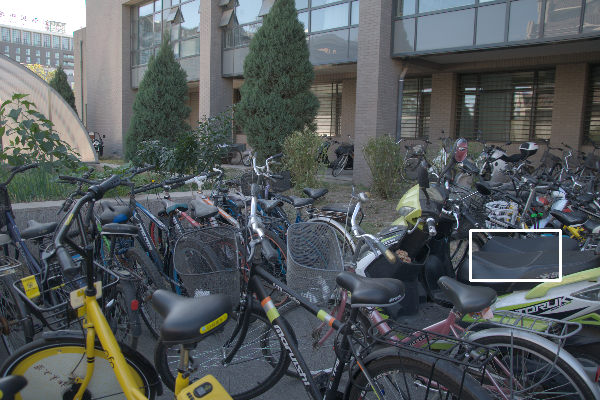}}
\vspace{2pt}
\centerline{\includegraphics[width=\textwidth]{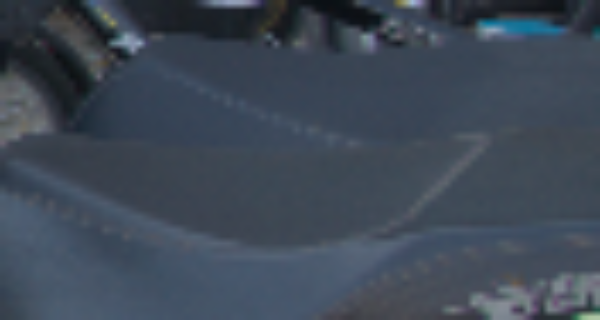}}
\centerline{\scriptsize\textbf{(e)}}
\end{minipage}%
\caption{\label{fig5}Visual comparisons of diverse components. (a) Input  (b) TDM (c) TDM + IGPACM (d) Ours (e) Ground Truth. NES~\cite{chen2015efficient} represents the noise estimation score. Please zoom in for a better view.}
\end{figure*}

\subsection{Quantitative and Visual Comparisons}

To comprehensively evaluate the effectiveness of IGDNet, we conduct comparative analyses against existing methods in terms of both quantitative metrics and visual quality.

\textbf{Numerical indicators.} As shown in Table~\ref{tab:2}, IGDNet outperforms most unsupervised methods, achieving either the best or second-best results across five evaluation metrics on three datasets. It also significantly narrows the performance gap with supervised approaches.

Supervised methods often overfit to training labels, limiting their generalization to unseen scenarios. Variations in lighting or content frequently degrade performance; for instance, RetinexFormer performs well on LOL but underperforms on other datasets, revealing its dataset dependency.

While unsupervised methods avoid labeled data, many still rely on prior guidance or are biased toward specific data distributions, limiting their adaptability. For example, ZR-PQR depends on a pre-trained diffusion model, and SCI trained on alternate datasets performs poorly on LOL.

In contrast, the zero-shot design of IGDNet offers clear advantages. It enhances underexposed images without any exposure-specific training data, demonstrating strong flexibility and generalization. This adaptability enables robust performance across diverse and previously unseen lighting conditions. By eliminating dataset dependence and integrating novel architectural designs, IGDNet consistently delivers high-quality enhancement, making it a promising solution for a wide range of low-light applications.

\textbf{Visual effects.} Fig.~\ref{fig2} presents visual comparisons between IGDNet and state-of-the-art methods covering diverse scenes and lighting conditions. In low overall brightness scenarios (first two rows), CLIP-LIT and Zero-DCE show limited control over global illumination, resulting in under-enhanced and dim outputs. RUAS improves brightness but tends to cause global overexposure, leading to detail loss in highlight areas. PairLIE and ZR-PQR exhibit noticeable color deviations, indicating poor color preservation. In contrast, IGDNet effectively suppresses noise and preserves structural details, resulting in balanced and visually appealing enhancement.

In scenarios with low local brightness (last four rows of Fig.~\ref{fig2}), a similar trend is observed. While SCI and RUAS achieve notable brightness improvement on the LOLv1, they often cause global overexposure on unseen datasets, likely due to overfitting to specific training data. RDHCE occasionally introduces artifacts that degrade image quality. In contrast, IGDNet consistently delivers results that better align with human visual perception, demonstrating superior robustness and naturalness in handling non-uniform illumination.

\textbf{Visual Tasks in Dimly Lit Environments.} Enhancing under-exposed images frequently supports higher-level visual tasks like target detection and instance segmentation. We employ the face detection model~\cite{ref53} and YOLOv8~\cite{ref54} for object detection and instance segmentation to assess the proposed method on the Dark Face dataset.

In Fig. \ref{fig3} (i), IGDNet surpasses comparative approaches in terms of both the quantity and classification of detected faces. In the instance segmentation task, IGDNet demonstrates superior precision and more refined boundary delineation. Furthermore, the proposed method exhibits a significant improvement in detection confidence, indicating that IGDNet not only enhances the visual quality of images but also boosts the performance of downstream tasks.

The observed improvements can be attributed to IGDNet's ability to deliver clearer and more detailed enhancement effects when processing underexposed images. The enhanced images facilitate the extraction of critical features by object detection and instance segmentation models, thereby achieving higher accuracy and confidence in these tasks. This transition from performance metrics to the underlying cause provides a cohesive understanding of how IGDNet's advancements contribute to overall task effectiveness.

\begin{table}[h]
\centering
\caption{Quantitative results of ablation experiment. The best results are marked in bold.}
\resizebox{\columnwidth}{!}{
\begin{tabular}{ccc|cc|c|ccccc}
\hline
\multicolumn{3}{c|}{TDM} & \multicolumn{3}{c|}{TDEM} & \multicolumn{5}{c}{LOLv2-Synthetic} \\ 
\hline
$\mathcal{L}_{recon}$ & $\mathcal{L}_{tv}$ & \multicolumn{1}{c|}{$\mathcal{L}_{noise}$} & $\mathcal{L}_{sr}$ & $\mathcal{L}_{sc}$& \multicolumn{1}{c|}{IGPACM} & PSNR$\uparrow$ & SSIM$\uparrow$ & LPIPS$\downarrow$ & LOE$\downarrow$ & MAE$\downarrow$ \\ 
\hline
$\surd$ &  &  & $\surd$ & $\surd$ & $\surd$ & 18.33 & 0.808 & 0.176 & 0.135 & 0.123 \\ 
$\surd$ & $\surd$ &  & $\surd$  & $\surd$ & $\surd$ & 19.22 & 0.850 & 0.164 & 0.135 & 0.119 \\ 
$\surd$ &  & $\surd$ & $\surd$ & $\surd$ & $\surd$ & 18.65 & 0.809 & 0.176 & 0.139 & 0.111 \\ 
\hline
$\surd$ & $\surd$ & $\surd$ &  &  &  & 14.63 & 0.652 & 0.251 & \textbf{0.117} & 0.243 \\ 
$\surd$ & $\surd$ & $\surd$ & $\surd$ &  & $\surd$ & 19.46 & 0.853 & 0.160 & 0.128 & 0.108 \\ 
$\surd$ & $\surd$ & $\surd$ &  & $\surd$ &  & 17.25 & 0.826 & 0.257 & 0.138 & 0.135 \\ 
$\surd$ & $\surd$ & $\surd$ &  &  & $\surd$ & 19.73 & 0.859 & 0.155 & 0.118 & \textbf{0.104} \\ 
\hline
$\surd$ & $\surd$ & $\surd$ & $\surd$ & $\surd$ & $\surd$ & \textbf{19.75} & \textbf{0.860} & \textbf{0.157} & 0.119 & \textbf{0.104} \\ 
\hline
\end{tabular}}
\label{tab:3}
\end{table}

\begin{figure}[htbp]
\centering
\begin{minipage}[htbp]{0.333\linewidth}
\centerline{\includegraphics[width=\textwidth]{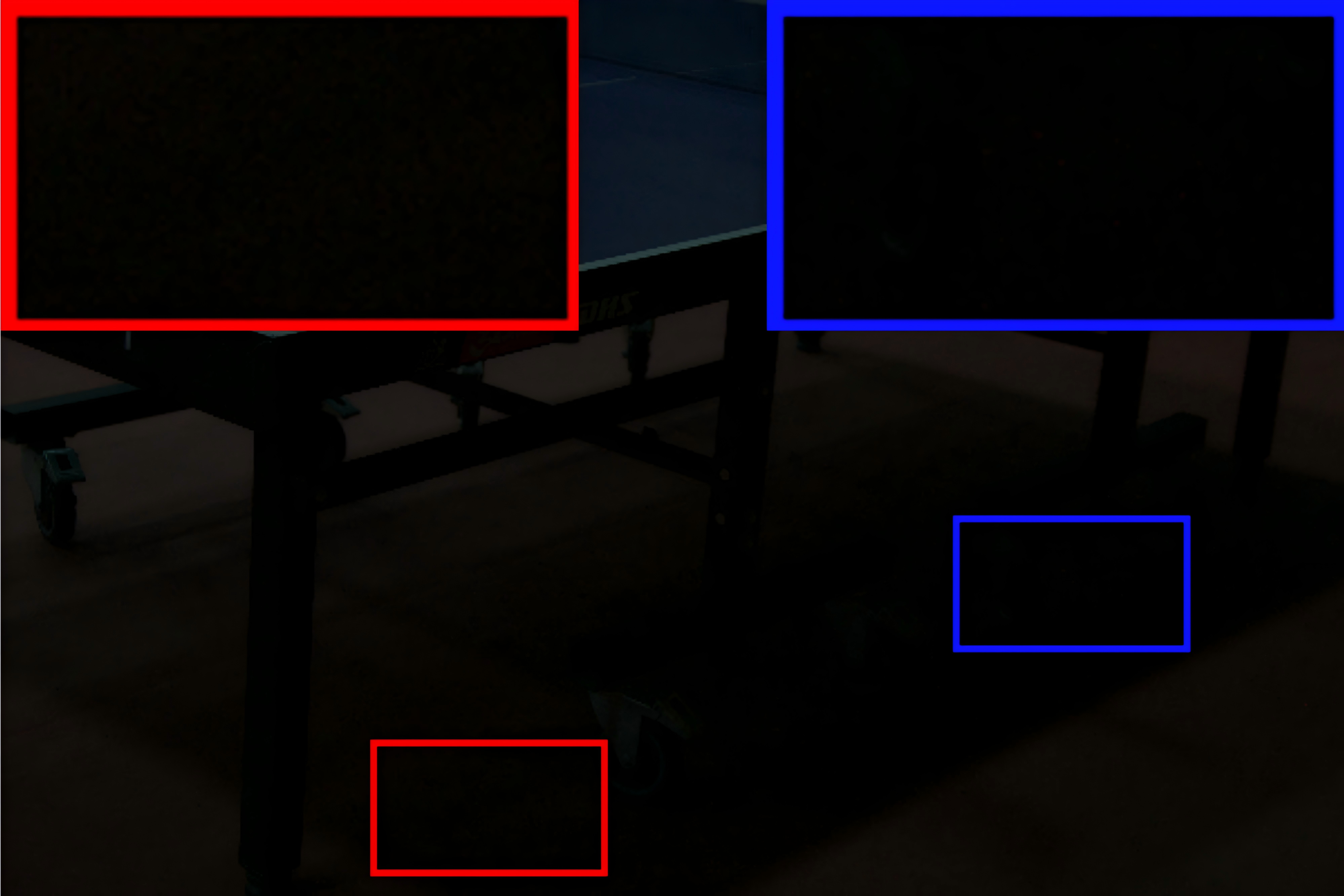}}
\centerline{\scriptsize\textbf{(a). Input, $I$}}
\centerline{\includegraphics[width=\textwidth]{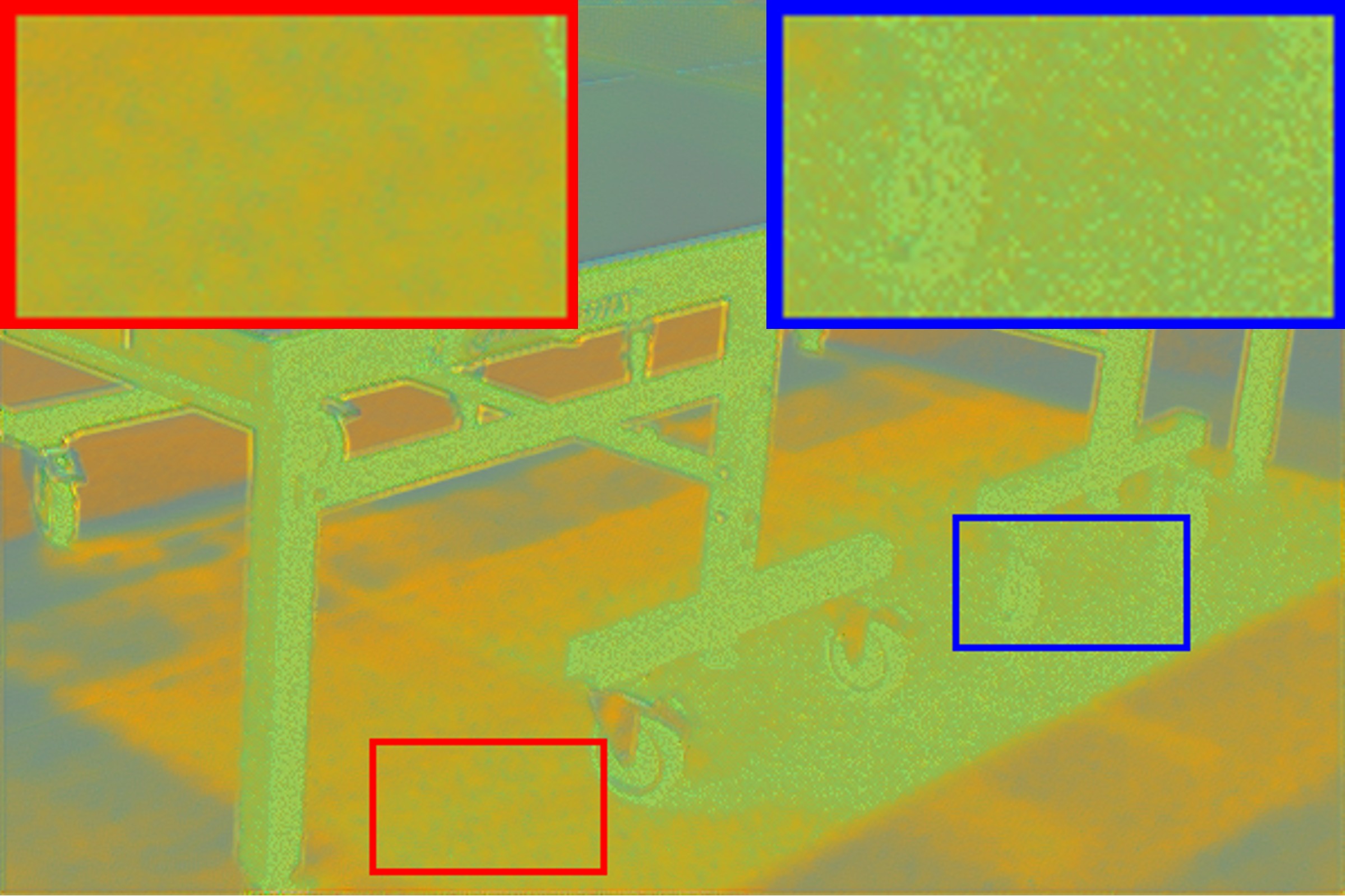}}
\centerline{\scriptsize\textbf{(d). Noise, $N$}}
\end{minipage}%
\begin{minipage}[htbp]{0.333\linewidth}
\centerline{\includegraphics[width=\textwidth]{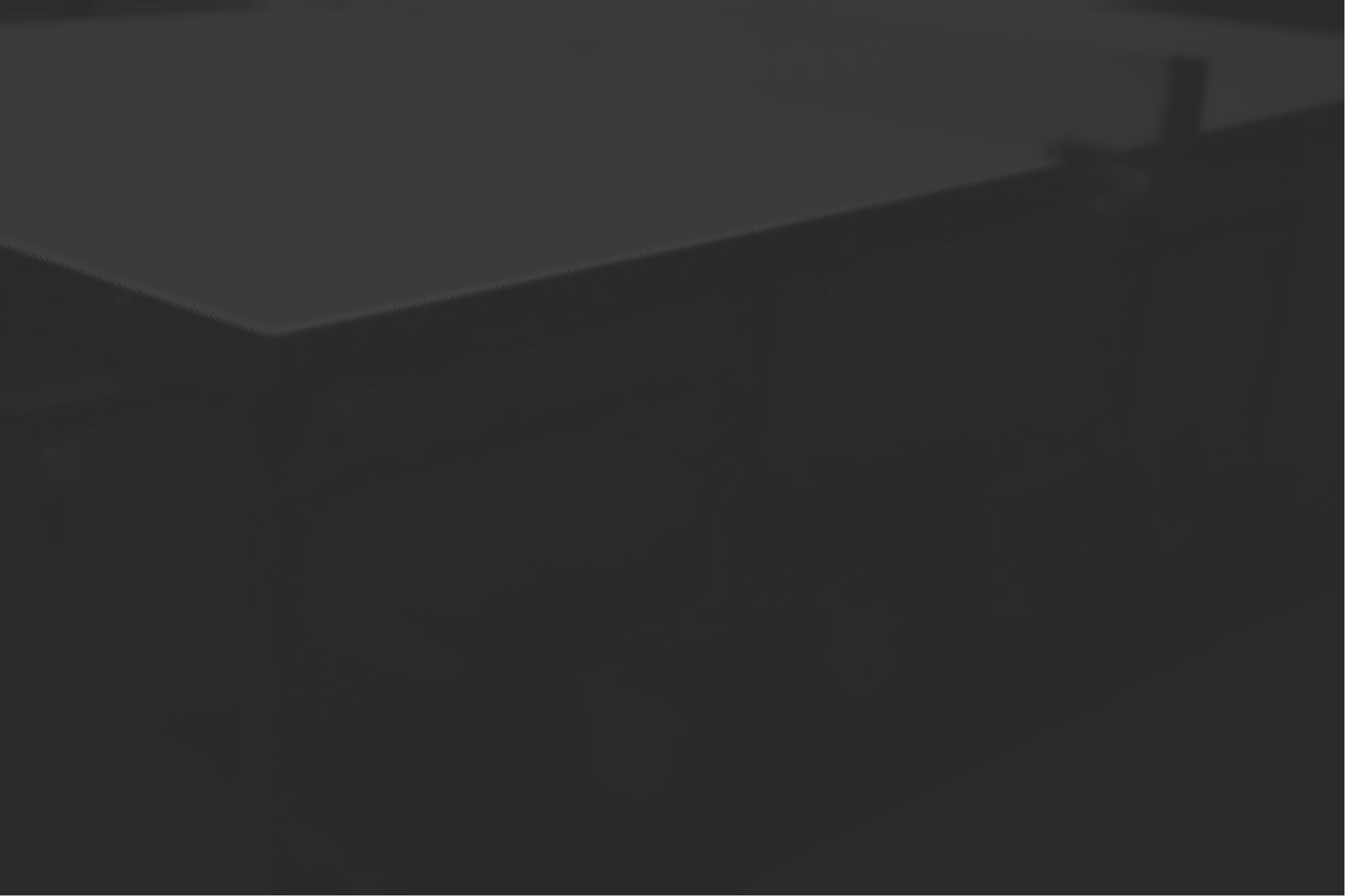}}
\centerline{\scriptsize\textbf{(b). illumination, $L_{1}$}}
\centerline{\includegraphics[width=\textwidth]{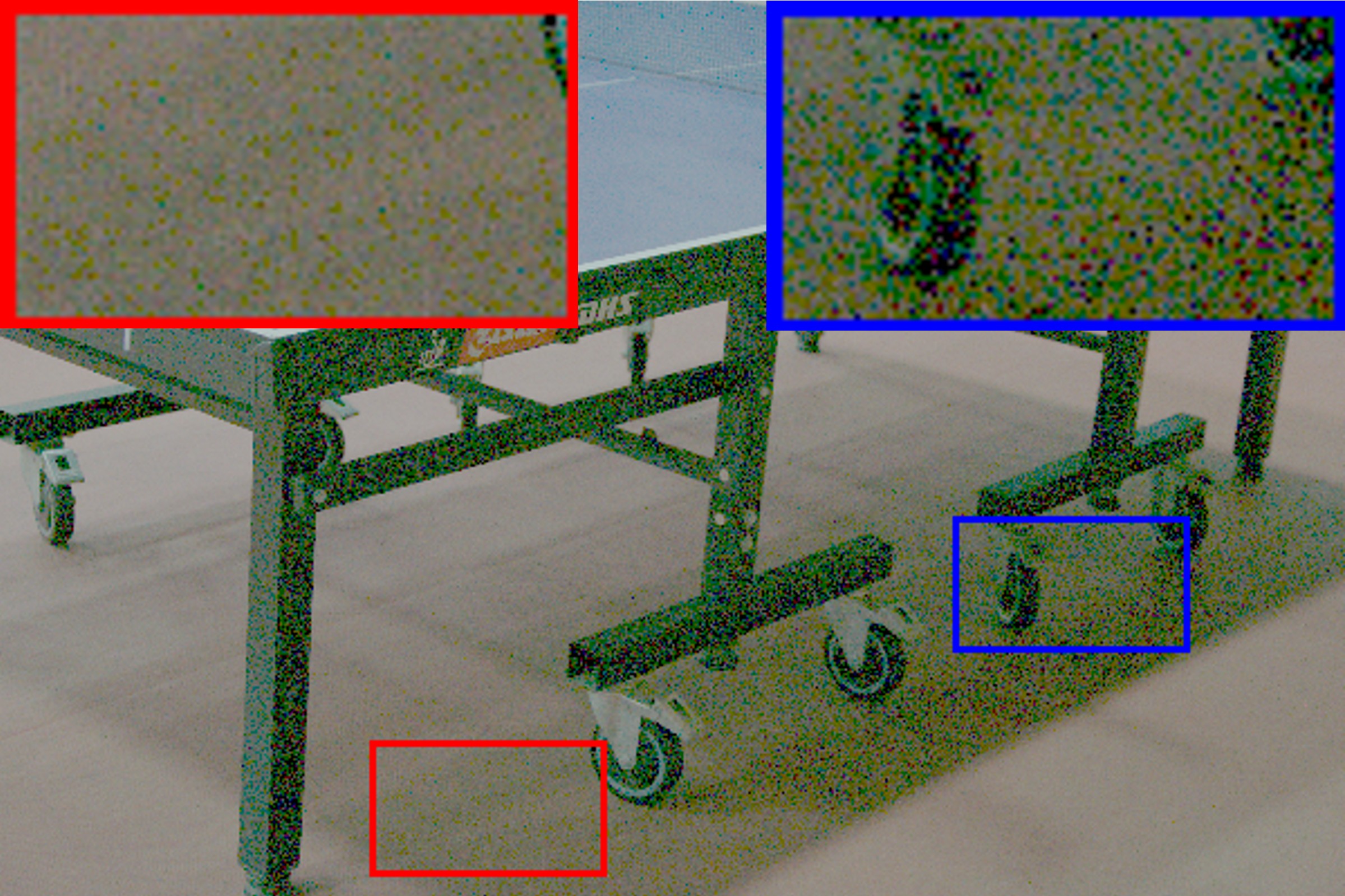}}
\centerline{\scriptsize\textbf{(e). Gamma ($I$)}}
\end{minipage}%
\begin{minipage}[htbp]{0.333\linewidth}
\centerline{\includegraphics[width=\textwidth]{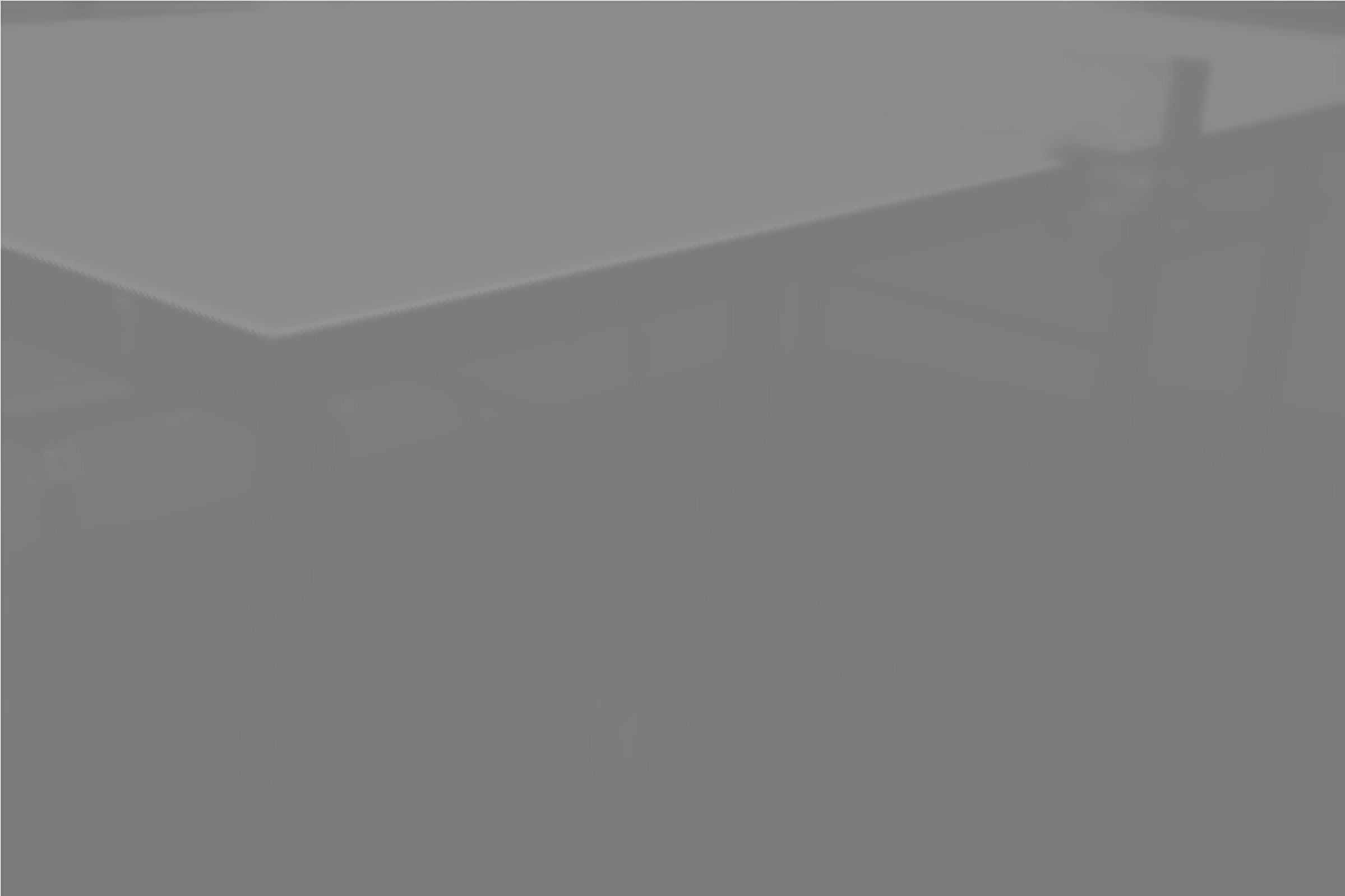}}
\centerline{\scriptsize\textbf{(c). Gamma($L_{1}$) = $L_{2}$}}
\centerline{\includegraphics[width=\textwidth]{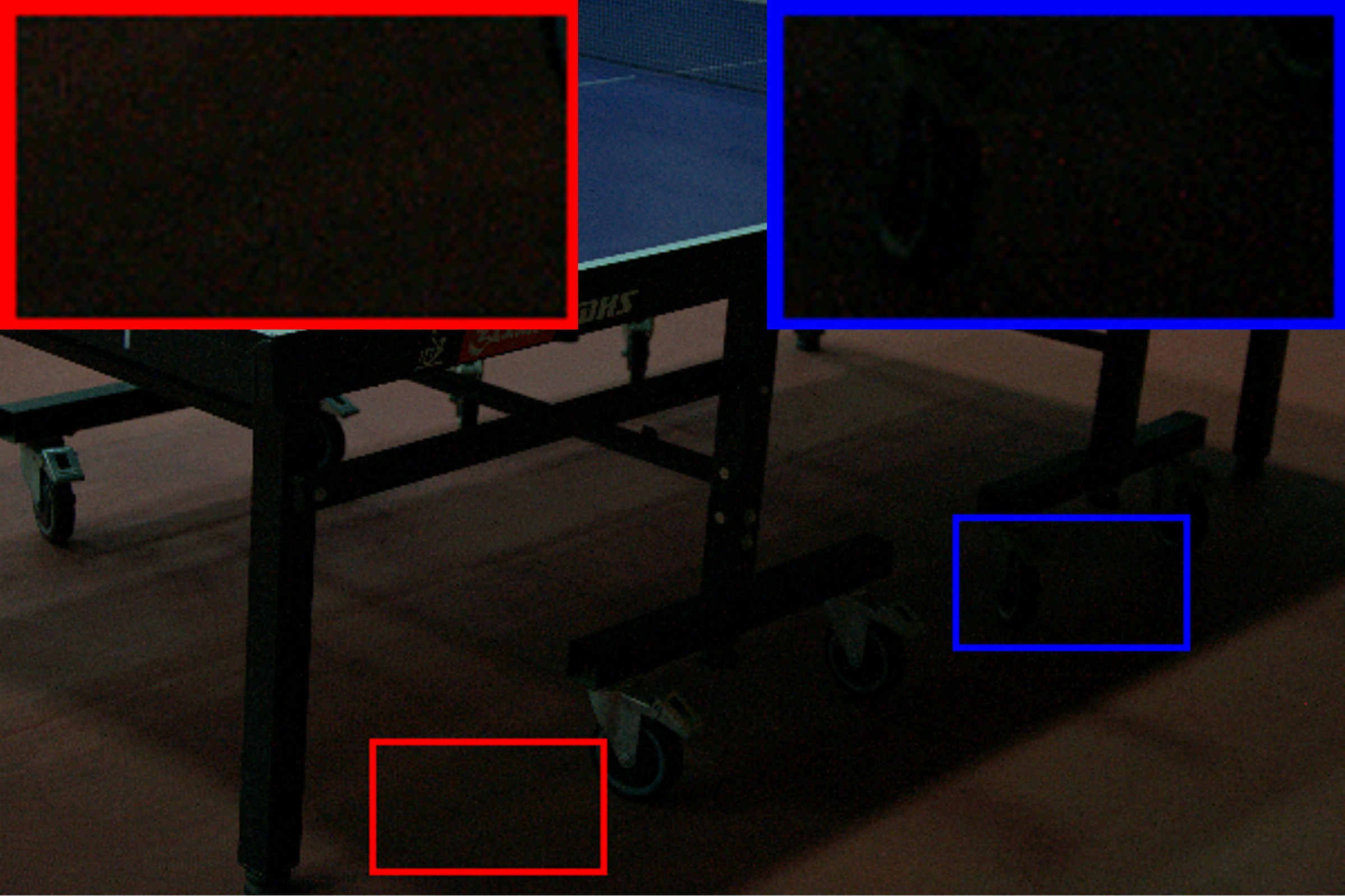}}
\centerline{\scriptsize\textbf{(f). Coarse enhanced result, $\tilde{I}$}}
\end{minipage}%
\caption{\label{fig6}Visualization of different components in Decomposition Module. Gamma represents gamma correction. Please zoom in for a better view.}
\end{figure}

\subsection{Ablation Study}
\textbf{Quantitative ablation.} We conducted a comprehensive ablation study on the LOLv2-Synthetic, integrating various loss components and quantitatively evaluating their contributions, as illustrated in Table \ref{tab:3}. TDM, TDEM, and IGPACM denote the decomposition module, the denoising module, and the illumination-guided pixel-adaptive correction module, respectively. Specifically, the results from employing only TDM revealed lower scores, underscoring the efficacy of TDEM. Overall, the performance enhancement of the complete version validates the importance of each loss component.

\textbf{Qualitative ablation.} To further elucidate the pivotal role of the denoising module in enhancing underexposed images, we initially utilized NES to estimate the noise levels associated with each module, as depicted in Fig. \ref{fig4}. An examination of Fig.\ref{fig4} reveals that despite the noise amplification observed during the illumination correction process~\cite{lv2021attention,li2021low}, the denoising module effectively mitigated the noise levels in the enhanced images, thereby improving visual quality (Fig. \ref{fig5}).

Fig. \ref{fig6} presents the visual outputs of each component in the decomposition module, offering a clear depiction of the pipeline’s intermediate stages. The raw illumination component ($L_{1}$, Fig. \ref{fig6}(b)) is first gamma-corrected to obtain a coarse illumination ($L_{2}$, Fig. \ref{fig6}(c)), enhancing the global brightness of the underexposed image. This avoids the artifacts, such as local over-enhancement and amplified noise, caused by directly applying gamma correction to the original input (Fig. \ref{fig6}(e)). Concurrently, the noise component ($N$, Fig. \ref{fig6}(d)) captures structured noise and is fused with $L_{2}$ via Eq. (\ref{(2)}), yielding a denoised and coarsely enhanced result ($\tilde{I}$, Fig. \ref{fig6}(f)) with improved brightness and structural clarity. Based on this, IGPACM refines local regions, enhancing contrast while mitigating over-enhancement (Fig. \ref{fig5}(c)). Finally, the denoising module suppresses residual noise, producing a clean, natural output (Fig. \ref{fig5}(d)). These visualizations validate the effectiveness of IGDNet’s staged strategy.

\textbf{Hyper-factor ablation.}
Two Hyper-factor require determination: the gamma value $\gamma$ used for illumination correction in the decomposition module, and the factor $\lambda_{n}$ for the $\mathcal{L}_{noise}$.

\begin{figure}[tbp]
\centering
 \begin{minipage}[htbp]{0.5\linewidth}
\centerline{\includegraphics[width=\textwidth]{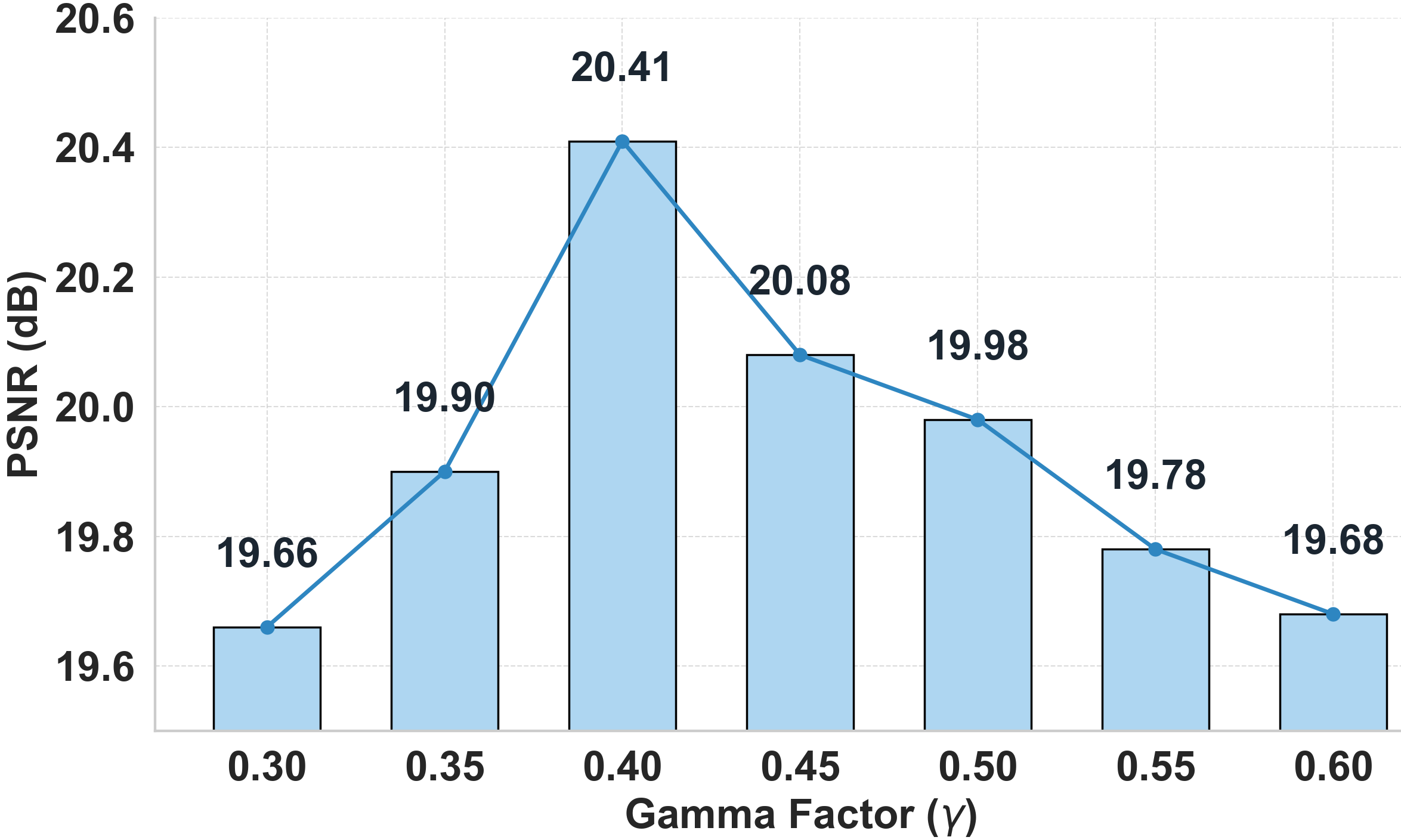}}
\centerline{\scriptsize\textbf{(a)}}
\end{minipage}%
\begin{minipage}[htbp]{0.5\linewidth}
\centerline{\includegraphics[width=\textwidth]{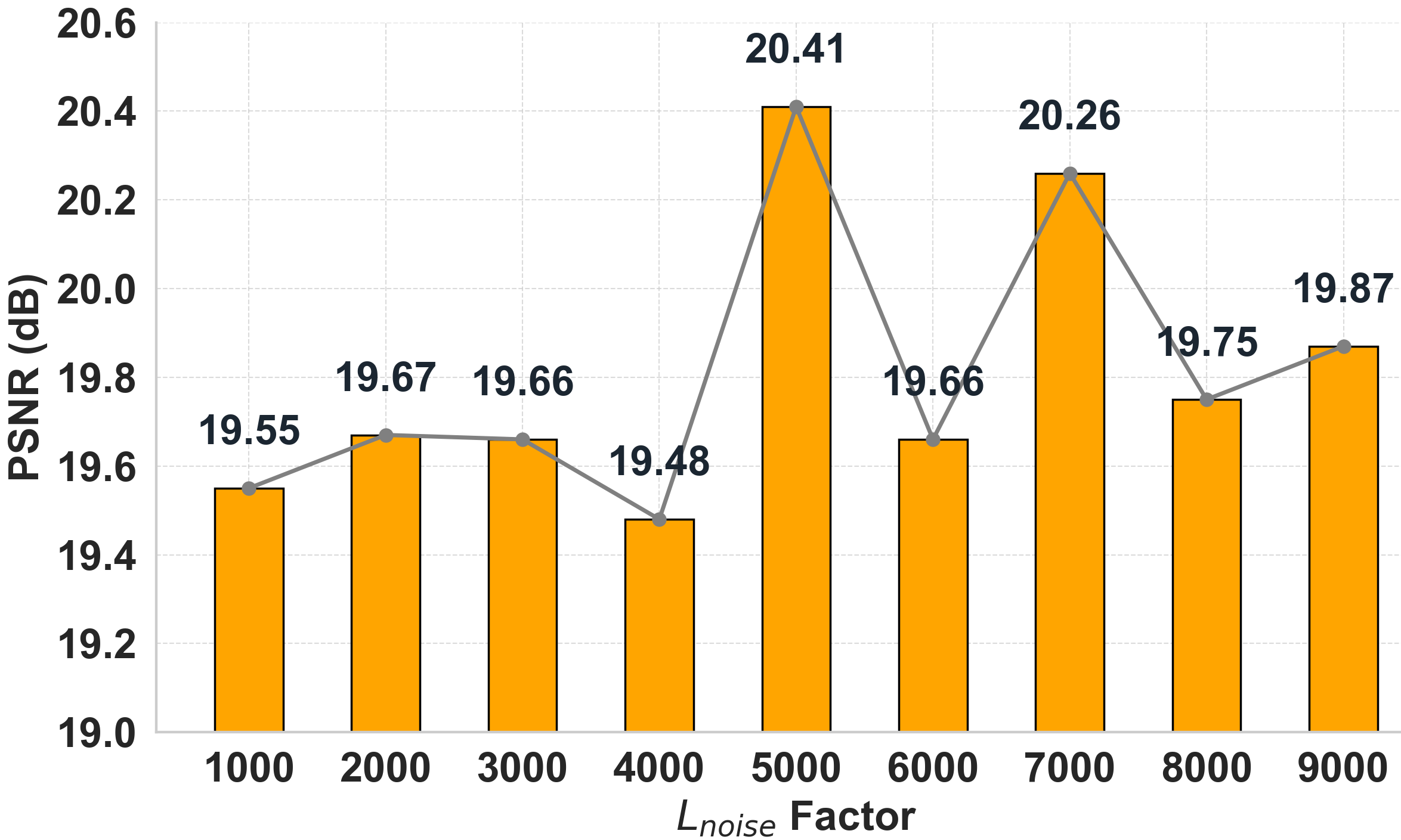}}
\centerline{\scriptsize\textbf{(b)}}
\end{minipage}%
\caption{\label{fig7}Impact of gamma and $\mathcal{L}_{noise}$ factors on PSNR performance.}
\end{figure}

For $\gamma$, since the initial illumination $L_{1}$ undergoes gamma correction for coarse enhancement, an excessively low value may lead to insufficient brightness, while a high value may cause local overexposure. Therefore, we constrain the search range to [0.3, 0.6]. Experiments on the LOLv1 and LOLv2-Real datasets were conducted to evaluate PSNR across different $\gamma$ values. As shown in Fig. \ref{fig7}(a), the best PSNR is achieved when $\gamma$=0.40, which is adopted as the final setting.

For $\lambda_{n}$, according to Eq. (\ref{(10)}), $\mathcal{L}_{noise}$ is guided by the illumination component to suppress potential noise that may be amplified alongside brightness in underexposed regions. However, since the illumination map has a value range of [0, 1]—typically much lower in dark areas—and the noise map, though bounded by Tanh in [-1, 1], is highly sparse with most values close to zero, the resulting loss is numerically small. This results in insufficient gradient flow during backpropagation, necessitating a higher loss weight to maintain balanced optimization with other terms.

Moreover, increasing the weight helps prevent ``pseudo-structure learning.'' Without sufficient constraint, the network may mistakenly encode texture and edge details, originally belonging to the reflectance component, into the noise map, resulting in structural confusion and semantic inconsistency. A higher weight mitigates this issue by enforcing a clearer separation between reflectance and noise. To identify the optimal setting, we conducted a grid search over $\lambda_{n}$. As shown in Fig. \ref{fig7}(b), the best PSNR performance is achieved when $\lambda_{n}$=5,000. Lower values fail to suppress noise adequately, while excessively high values risk over-smoothing image details. Therefore, $\lambda_{n}$=5,000 offers the best trade-off between noise suppression and detail preservation.

\begin{table}[htbp]
  \centering
  \caption{Evaluation of iterative image enhancement methods.}
\resizebox{1\columnwidth}{!}{
    \begin{tabular}{ccccc}
    \hline
    Method & Venue & PSNR  & SSIM & Time (s) \\
    \hline
    RetinexDIP \cite{zhao2021retinexdip} & TCSVT’21 & 9.44 & 0.322 & 39.29 \\
    DUNP \cite{liang2022self} & TCSVT’22 & 15.49 & 0.654 & 153.27 \\
    ZR-PQR \cite{ref38}& CVPR’24 & \textcolor{blue}{20.31} & \textcolor{red}{0.808} & 29.78 \\
    FourierDiff \cite{lv2024fourier}& CVPR’24 & 18.67 & 0.602 & 32.01 \\
    SOFG \cite{gu2024seed}& TCSVT’25 & 18.10 & 0.750 & \textcolor{red}{20.30} \\
    OUR & - & \textcolor{red}{20.41} & \textcolor{blue}{0.803} & \textcolor{blue}{26.67} \\
    \hline
    \end{tabular}}
  \label{tab:4}
\end{table}

\textbf{Efficiency ablation.} The primary computational cost of IGDNet arises from its image-specific iterative optimization, which aims to maximize performance through decomposition for each input. Unlike pretrained models, IGDNet operates without training data, demonstrating strong generalization and independence from data distribution, which is particularly advantageous in scenarios with limited or biased data.

While IGDNet has higher latency than pretrained models, it delivers competitive performance. As shown in Table \ref{tab:4}, it achieves leading PSNR (20.41) and SSIM (0.803) scores, with significantly lower latency than most iterative methods, such as DUNP (153.27s) and RetinexDIP (39.29s). This balance between quality and efficiency underscores its practical value.

Significantly, IGDNet is not positioned in opposition to pretrained models; rather, it serves as a practical complement in cases where training data is insufficient or of low quality. Its untrained network prior not only performs well in image enhancement but also exhibits promising potential for other low-level vision tasks, thereby extending the applicability of existing approaches to a broader range of scenarios.

\begin{figure}[htbp]
\centering
 \begin{minipage}[htbp]{0.3333\linewidth}
\centerline{\includegraphics[width=\textwidth]{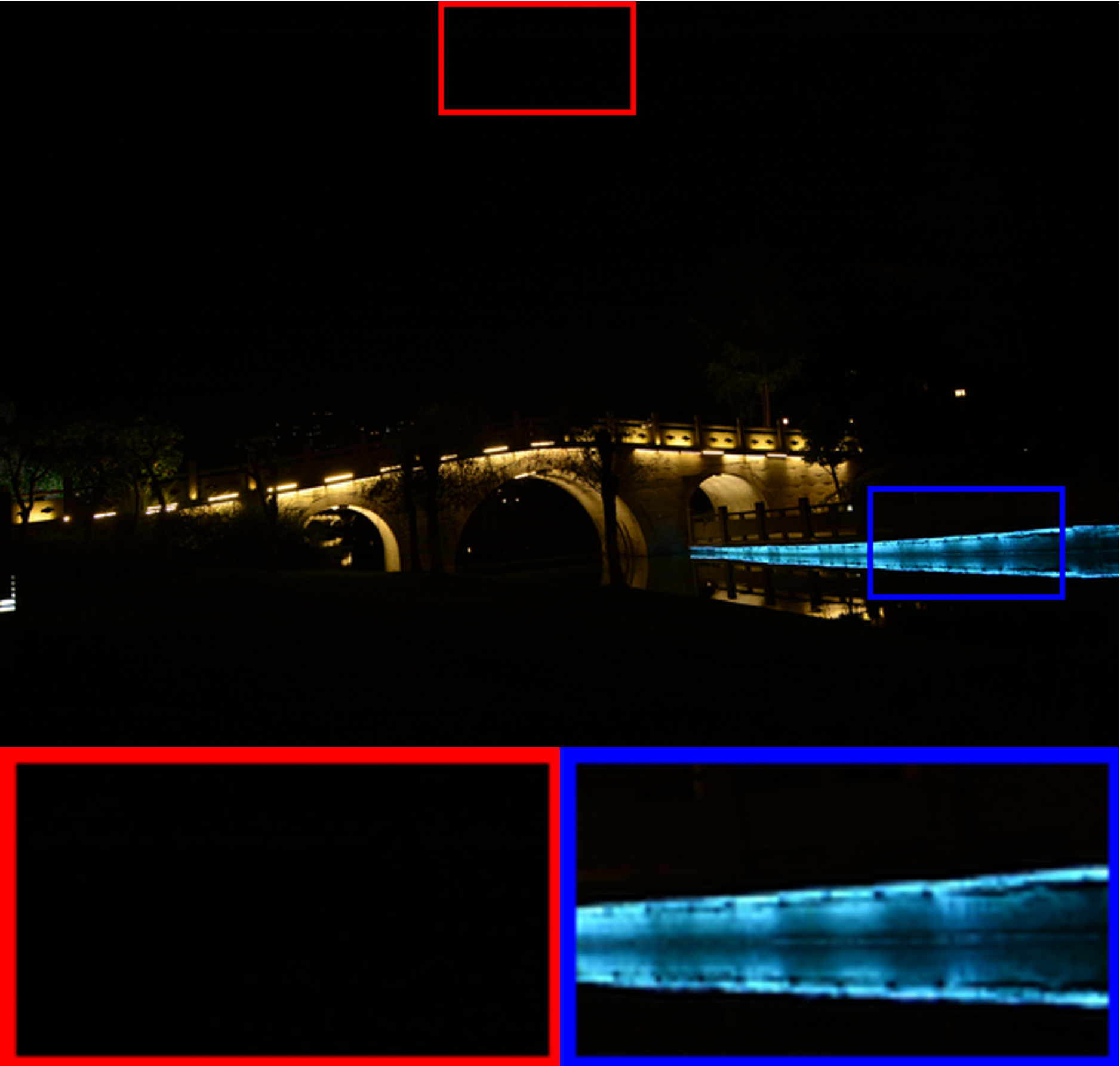}}
\centerline{\scriptsize\textbf{(a). Input}}
\end{minipage}%
\begin{minipage}[htbp]{0.3333\linewidth}
\centerline{\includegraphics[width=\textwidth]{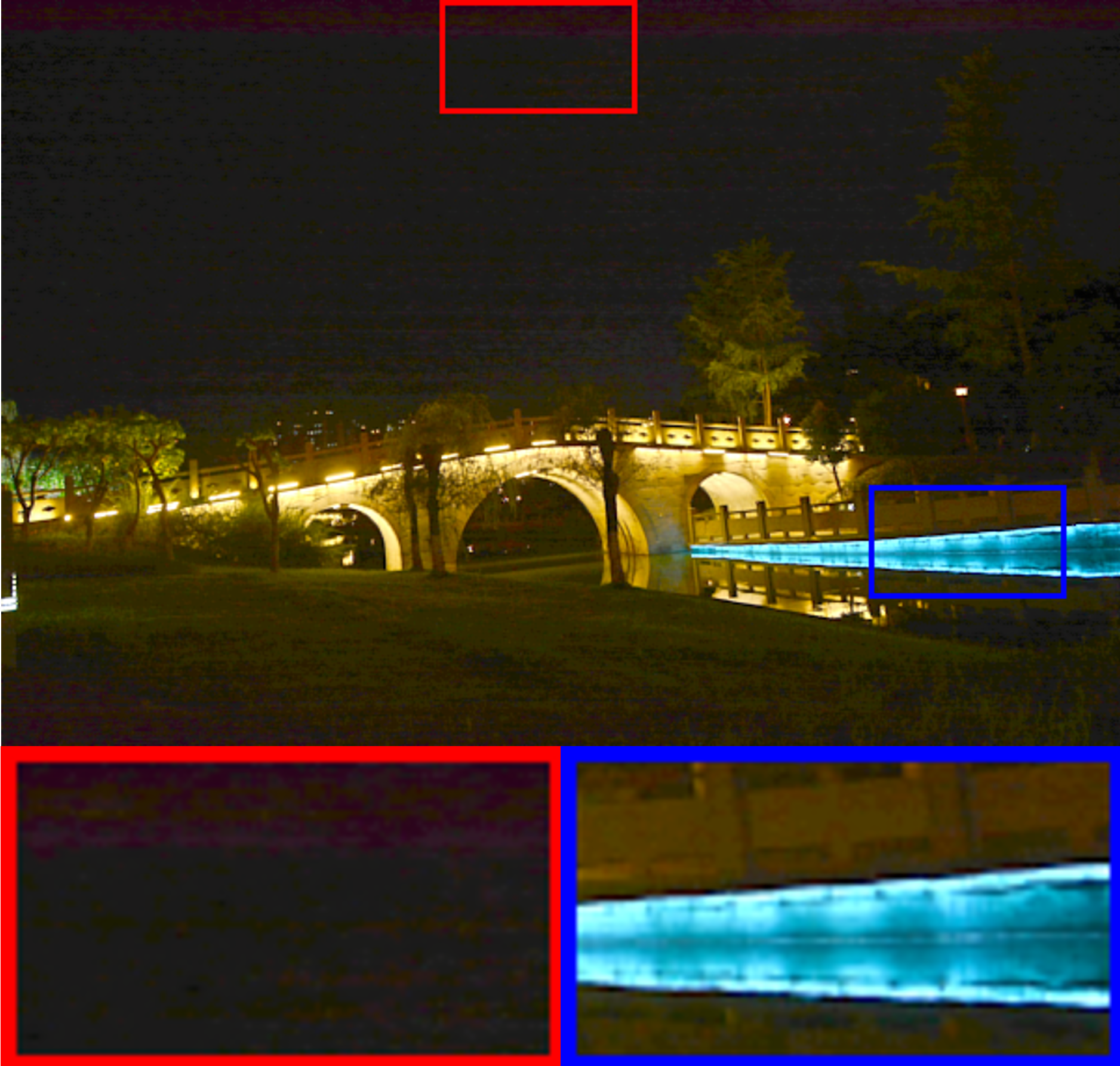}}
\centerline{\scriptsize\textbf{(b). Ours}}
\end{minipage}%
\begin{minipage}[htbp]{0.3333\linewidth}
\centerline{\includegraphics[width=\textwidth]{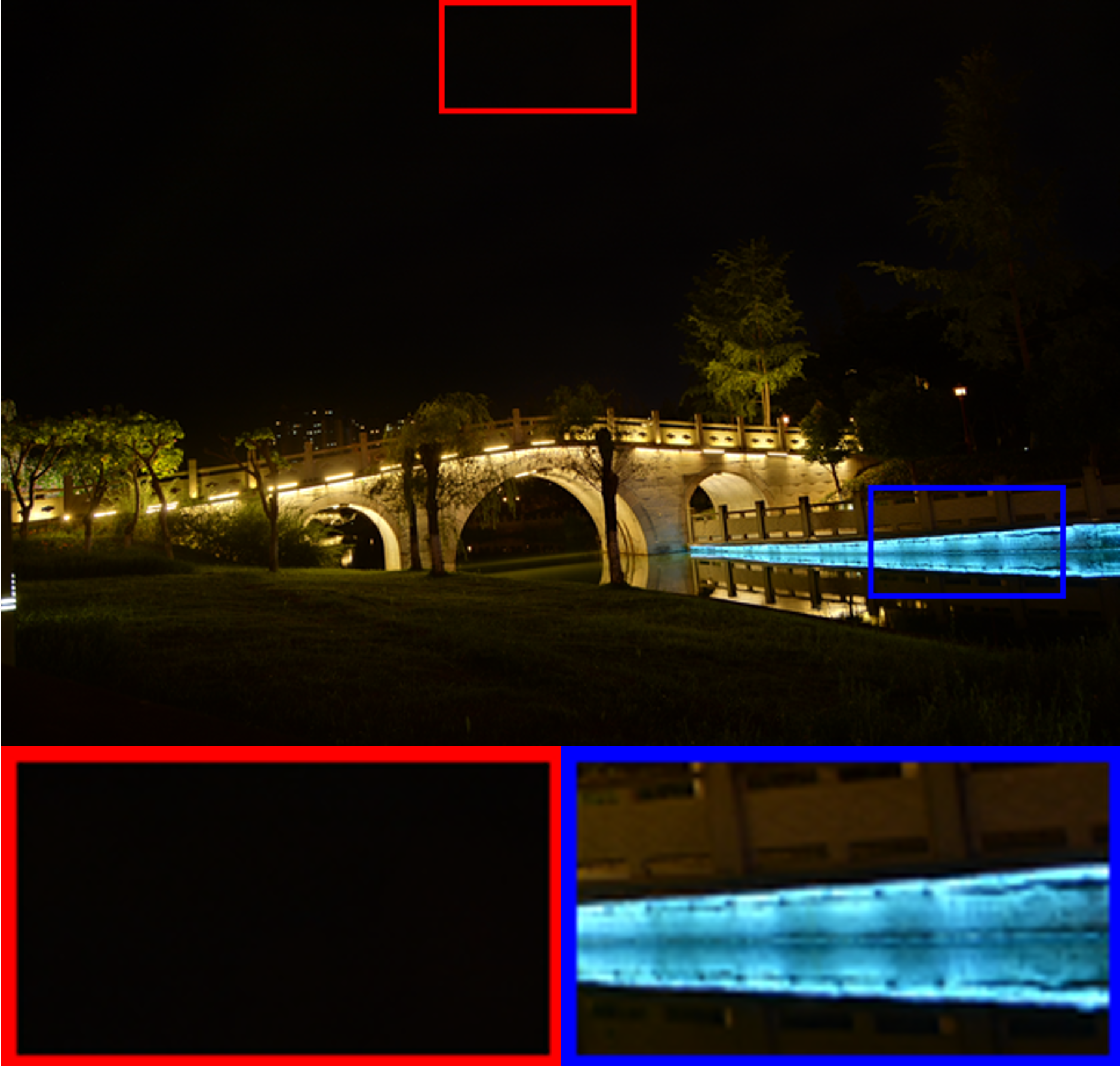}}
\centerline{\scriptsize\textbf{(c). Ground Truth}}
\end{minipage}%
\caption{\label{fig8}Failure case in extremely dark regions: IGDNet introduces artifacts in signal-void areas such as the night sky (\textcolor{red}{red box}), despite accurate enhancement in textured regions (\textcolor{blue}{blue box}).  Please zoom in for a better view.}
\end{figure}

\subsection{Limitation Analysis}

While IGDNet effectively enhances brightness and structural clarity in most underexposure scenarios, it struggles in extremely dark areas. As shown in Fig. \ref{fig8}, the model introduces artifacts when processing nearly information-less regions (e.g, night skies).

In the red box area, the input contains almost no visible information. Since the Retinex-based decomposition relies on luminance, the illumination component approaches zero in such regions, making it difficult for the model to extract meaningful structure. To minimize reconstruction loss, the network may "hallucinate" textures or brightness, leading to false content, manifesting as color patches, noise, or unrealistic structures, as illustrated in the middle column.

Moreover, current loss functions lack explicit constraints for content-absent regions that compromise perceptual quality. Future work could incorporate dark-region-aware mechanisms or local confidence modulation to adjust enhancement strength, thereby improving robustness and naturalness dynamically.

\section{Conclusion}
This paper presents IGDNet, a novel zero-shot approach designed to enhance underexposed images. Unlike supervised learning methods, IGDNet eliminates reliance on large-scale training datasets and effectively suppresses noise while restoring illumination. Extensive experiments validate its robustness across a range of challenging lighting conditions, particularly for images with non-uniform exposure. Compared to existing methods, IGDNet achieves competitive or superior performance. Furthermore, its training-free nature makes it especially suitable for real-world scenarios where clean, unbiased training data are scarce or unavailable.

\bibliographystyle{IEEEtran}
\bibliography{IEEEabrv,egbib}

\begin{thebibliography}{10}
\providecommand{\url}[1]{#1}
\csname url@samestyle\endcsname
\providecommand{\newblock}{\relax}
\providecommand{\bibinfo}[2]{#2}
\providecommand{\BIBentrySTDinterwordspacing}{\spaceskip=0pt\relax}
\providecommand{\BIBentryALTinterwordstretchfactor}{4}
\providecommand{\BIBentryALTinterwordspacing}{\spaceskip=\fontdimen2\font plus
\BIBentryALTinterwordstretchfactor\fontdimen3\font minus \fontdimen4\font\relax}
\providecommand{\BIBforeignlanguage}[2]{{%
\expandafter\ifx\csname l@#1\endcsname\relax
\typeout{** WARNING: IEEEtran.bst: No hyphenation pattern has been}%
\typeout{** loaded for the language `#1'. Using the pattern for}%
\typeout{** the default language instead.}%
\else
\language=\csname l@#1\endcsname
\fi
#2}}
\providecommand{\BIBdecl}{\relax}
\BIBdecl

\bibitem{ref1}
J.~Liu, D.~Xu, W.~Yang, M.~Fan, and H.~Huang, ``Benchmarking low-light image enhancement and beyond,'' \emph{International Journal of Computer Vision}, vol. 129, pp. 1153--1184, 2021.

\bibitem{li2021low}
C.~Li, C.~Guo, L.~Han, J.~Jiang, M.-M. Cheng, J.~Gu, and C.~C. Loy, ``Low-light image and video enhancement using deep learning: A survey,'' \emph{IEEE Transactions on Pattern Analysis and Machine Intelligence}, vol.~44, no.~12, pp. 9396--9416, 2021.

\bibitem{ref27}
Y.~Jiang, X.~Gong, D.~Liu, Y.~Cheng, C.~Fang, X.~Shen, J.~Yang, P.~Zhou, and Z.~Wang, ``Enlightengan: Deep light enhancement without paired supervision,'' \emph{IEEE Transactions on Image Processing}, vol.~30, pp. 2340--2349, 2021.

\bibitem{ref30}
C.~Guo, C.~Li, J.~Guo, C.~C. Loy, J.~Hou, S.~Kwong, and R.~Cong, ``Zero-reference deep curve estimation for low-light image enhancement,'' in \emph{Proceedings of the IEEE/CVF Conference on Computer Vision and Pattern Recognition}, 2020, pp. 1780--1789.

\bibitem{ref2}
C.~Lee, C.~Lee, and C.-S. Kim, ``Contrast enhancement based on layered difference representation of 2d histograms,'' \emph{IEEE Transactions on Image Processing}, vol.~22, no.~12, pp. 5372--5384, 2013.

\bibitem{ref3}
E.~H. Land, ``The retinex theory of color vision,'' \emph{Scientific American}, vol. 237, no.~6, pp. 108--129, 1977.

\bibitem{ref4}
W.~Wang, N.~Sun, and M.~K. Ng, ``A variational gamma correction model for image contrast enhancement,'' \emph{Inverse Problems and Imaging}, vol.~13, no.~3, pp. 461--478, 2019.

\bibitem{rahman2021structure}
Z.~Rahman, Y.-F. Pu, M.~Aamir, and S.~Wali, ``Structure revealing of low-light images using wavelet transform based on fractional-order denoising and multiscale decomposition,'' \emph{The Visual Computer}, vol.~37, no.~5, pp. 865--880, 2021.

\bibitem{rahman2022diverse}
Z.~Rahman, Z.~Ali, I.~Khan, M.~I. Uddin, Y.~Guan, and Z.~Hu, ``Diverse image enhancer for complex underexposed image,'' \emph{Journal of Electronic Imaging}, vol.~31, no.~4, pp. 041\,213--041\,213, 2022.

\bibitem{rahman2023efficient}
Z.~Rahman, M.~Aamir, Z.~Ali, A.~K.~J. Saudagar, A.~AlTameem, and K.~Muhammad, ``Efficient contrast adjustment and fusion method for underexposed images in industrial cyber-physical systems,'' \emph{IEEE Systems Journal}, vol.~17, no.~4, pp. 5085--5096, 2023.

\bibitem{ref10}
C.~Wei, W.~Wang, W.~Yang, and J.~Liu, ``Deep retinex decomposition for low-light enhancement,'' in \emph{British Machine Vision Conference}.\hskip 1em plus 0.5em minus 0.4em\relax British Machine Vision Association, 2018.

\bibitem{ref19}
W.~Wu, J.~Weng, P.~Zhang, X.~Wang, W.~Yang, and J.~Jiang, ``Uretinex-net: Retinex-based deep unfolding network for low-light image enhancement,'' in \emph{Proceedings of the IEEE/CVF Conference on Computer Vision and Pattern Recognition}, 2022, pp. 5901--5910.

\bibitem{ref20}
Y.~Cai, H.~Bian, J.~Lin, H.~Wang, R.~Timofte, and Y.~Zhang, ``Retinexformer: One-stage retinex-based transformer for low-light image enhancement,'' in \emph{Proceedings of the IEEE/CVF International Conference on Computer Vision}, 2023, pp. 12\,504--12\,513.

\bibitem{ref34}
L.~Ma, T.~Ma, R.~Liu, X.~Fan, and Z.~Luo, ``Toward fast, flexible, and robust low-light image enhancement,'' in \emph{Proceedings of the IEEE/CVF Conference on Computer Vision and Pattern Recognition}, 2022, pp. 5637--5646.

\bibitem{ref5}
X.~Guo, Y.~Li, and H.~Ling, ``Lime: Low-light image enhancement via illumination map estimation,'' \emph{IEEE Transactions on Image Processing}, vol.~26, no.~2, pp. 982--993, 2016.

\bibitem{ref29}
A.~Zhu, L.~Zhang, Y.~Shen, Y.~Ma, S.~Zhao, and Y.~Zhou, ``Zero-shot restoration of underexposed images via robust retinex decomposition,'' in \emph{2020 IEEE International Conference on Multimedia and Expo (ICME)}.\hskip 1em plus 0.5em minus 0.4em\relax IEEE, 2020, pp. 1--6.

\bibitem{ref6}
M.~Li, J.~Liu, W.~Yang, X.~Sun, and Z.~Guo, ``Structure-revealing low-light image enhancement via robust retinex model,'' \emph{IEEE Transactions on Image Processing}, vol.~27, no.~6, pp. 2828--2841, 2018.

\bibitem{ref7}
X.~Ren, W.~Yang, W.-H. Cheng, and J.~Liu, ``Lr3m: Robust low-light enhancement via low-rank regularized retinex model,'' \emph{IEEE Transactions on Image Processing}, vol.~29, pp. 5862--5876, 2020.

\bibitem{ref9}
J.~Xu, Y.~Hou, D.~Ren, L.~Liu, F.~Zhu, M.~Yu, H.~Wang, and L.~Shao, ``Star: A structure and texture aware retinex model,'' \emph{IEEE Transactions on Image Processing}, vol.~29, pp. 5022--5037, 2020.

\bibitem{yan2025mobileie}
H.~Yan, A.~Li, X.~Zhang, Z.~Liu, Z.~Shi, C.~Zhu, and L.~Zhang, ``Mobileie: An extremely lightweight and effective convnet for real-time image enhancement on mobile devices,'' in \emph{Proceedings of the IEEE/CVF International Conference on Computer Vision}, 2025.

\bibitem{rahman2020efficient}
Z.~Rahman, P.~Yi-Fei, M.~Aamir, S.~Wali, and Y.~Guan, ``Efficient image enhancement model for correcting uneven illumination images,'' \emph{IEEE Access}, vol.~8, pp. 109\,038--109\,053, 2020.

\bibitem{ref11}
Y.~Wang, Y.~Cao, Z.-J. Zha, J.~Zhang, Z.~Xiong, W.~Zhang, and F.~Wu, ``Progressive retinex: Mutually reinforced illumination-noise perception network for low-light image enhancement,'' in \emph{Proceedings of the 27th ACM International Conference on Multimedia}, 2019, pp. 2015--2023.

\bibitem{ref12}
Y.~Zhang, J.~Zhang, and X.~Guo, ``Kindling the darkness: A practical low-light image enhancer,'' in \emph{Proceedings of the 27th ACM International Conference on Multimedia}, 2019, pp. 1632--1640.

\bibitem{ref13}
Y.~Zhang, X.~Guo, J.~Ma, W.~Liu, and J.~Zhang, ``Beyond brightening low-light images,'' \emph{International Journal of Computer Vision}, vol. 129, pp. 1013--1037, 2021.

\bibitem{ref14}
R.~Wang, Q.~Zhang, C.-W. Fu, X.~Shen, W.-S. Zheng, and J.~Jia, ``Underexposed photo enhancement using deep illumination estimation,'' in \emph{Proceedings of the IEEE/CVF Conference on Computer Vision and Pattern Recognition}, 2019, pp. 6849--6857.

\bibitem{ref16}
M.~Fan, W.~Wang, W.~Yang, and J.~Liu, ``Integrating semantic segmentation and retinex model for low-light image enhancement,'' in \emph{Proceedings of the 28th ACM International Conference on Multimedia}, 2020, pp. 2317--2325.

\bibitem{ref17}
L.~Ma, R.~Liu, J.~Zhang, X.~Fan, and Z.~Luo, ``Learning deep context-sensitive decomposition for low-light image enhancement,'' \emph{IEEE Transactions on Neural Networks and Learning Systems}, vol.~33, no.~10, pp. 5666--5680, 2021.

\bibitem{ref18}
Y.~Wang, R.~Wan, W.~Yang, H.~Li, L.-P. Chau, and A.~Kot, ``Low-light image enhancement with normalizing flow,'' in \emph{Proceedings of the AAAI Conference on Artificial Intelligence}, vol.~36, no.~3, 2022, pp. 2604--2612.

\bibitem{ref25}
H.~Jiang, A.~Luo, H.~Fan, S.~Han, and S.~Liu, ``Low-light image enhancement with wavelet-based diffusion models,'' \emph{ACM Transactions on Graphics (TOG)}, vol.~42, no.~6, pp. 1--14, 2023.

\bibitem{ref28}
S.~Yang, M.~Ding, Y.~Wu, Z.~Li, and J.~Zhang, ``Implicit neural representation for cooperative low-light image enhancement,'' in \emph{Proceedings of the IEEE/CVF International Conference on Computer Vision}, 2023, pp. 12\,918--12\,927.

\bibitem{kandula2023illumination}
P.~Kandula, M.~Suin, and A.~Rajagopalan, ``Illumination-adaptive unpaired low-light enhancement,'' \emph{IEEE Transactions on Circuits and Systems for Video Technology}, vol.~33, no.~8, pp. 3726--3736, 2023.

\bibitem{ref31}
C.~Li, C.~Guo, and C.~C. Loy, ``Learning to enhance low-light image via zero-reference deep curve estimation,'' \emph{IEEE Transactions on Pattern Analysis and Machine Intelligence}, vol.~44, no.~8, pp. 4225--4238, 2021.

\bibitem{ref32}
Y.~Xia, F.~Xu, and Q.~Zheng, ``Zero-shot adaptive low light enhancement with retinex decomposition and hybrid curve estimation,'' in \emph{2023 International Joint Conference on Neural Networks (IJCNN)}.\hskip 1em plus 0.5em minus 0.4em\relax IEEE, 2023, pp. 1--8.

\bibitem{ref33}
R.~Liu, L.~Ma, J.~Zhang, X.~Fan, and Z.~Luo, ``Retinex-inspired unrolling with cooperative prior architecture search for low-light image enhancement,'' in \emph{Proceedings of the IEEE/CVF Conference on Computer Vision and Pattern Recognition}, 2021, pp. 10\,561--10\,570.

\bibitem{ref35}
Z.~Liang, C.~Li, S.~Zhou, R.~Feng, and C.~C. Loy, ``Iterative prompt learning for unsupervised backlit image enhancement,'' in \emph{Proceedings of the IEEE/CVF International Conference on Computer Vision}, 2023, pp. 8094--8103.

\bibitem{ref36}
S.~Zheng and G.~Gupta, ``Semantic-guided zero-shot learning for low-light image/video enhancement,'' in \emph{Proceedings of the IEEE/CVF Winter Conference on Applications of Computer Vision}, 2022, pp. 581--590.

\bibitem{ref37}
Z.~Fu, Y.~Yang, X.~Tu, Y.~Huang, X.~Ding, and K.-K. Ma, ``Learning a simple low-light image enhancer from paired low-light instances,'' in \emph{Proceedings of the IEEE/CVF Conference on Computer Vision and Pattern Recognition}, 2023, pp. 22\,252--22\,261.

\bibitem{ref38}
W.~Wang, H.~Yang, J.~Fu, and J.~Liu, ``Zero-reference low-light enhancement via physical quadruple priors,'' in \emph{Proceedings of the IEEE/CVF Conference on Computer Vision and Pattern Recognition}, 2024.

\bibitem{zhao2021retinexdip}
Z.~Zhao, B.~Xiong, L.~Wang, Q.~Ou, L.~Yu, and F.~Kuang, ``Retinexdip: A unified deep framework for low-light image enhancement,'' \emph{IEEE Transactions on Circuits and Systems for Video Technology}, vol.~32, no.~3, pp. 1076--1088, 2021.

\bibitem{ref40}
T.~Wu, W.~Wu, Y.~Yang, F.-L. Fan, and T.~Zeng, ``Retinex image enhancement based on sequential decomposition with a plug-and-play framework,'' \emph{IEEE Transactions on Neural Networks and Learning Systems}, 2023.

\bibitem{ref39}
S.~Wang, J.~Zheng, H.-M. Hu, and B.~Li, ``Naturalness preserved enhancement algorithm for non-uniform illumination images,'' \emph{IEEE Transactions on Image Processing}, vol.~22, no.~9, pp. 3538--3548, 2013.

\bibitem{ref41}
Z.~Zhang, S.~Zhao, X.~Jin, M.~Xu, Y.~Yang, S.~Yan, and M.~Wang, ``Noise self-regression: A new learning paradigm to enhance low-light images without task-related data,'' \emph{IEEE Transactions on Pattern Analysis and Machine Intelligence}, 2024.

\bibitem{ref42}
Y.~Mansour and R.~Heckel, ``Zero-shot noise2noise: Efficient image denoising without any data,'' in \emph{Proceedings of the IEEE/CVF Conference on Computer Vision and Pattern Recognition}, 2023, pp. 14\,018--14\,027.

\bibitem{ref44}
L.~Xu, Q.~Yan, Y.~Xia, and J.~Jia, ``Structure extraction from texture via relative total variation,'' \emph{ACM Transactions on Graphics (TOG)}, vol.~31, no.~6, pp. 1--10, 2012.

\bibitem{ref45}
K.~Zhang, W.~Zuo, Y.~Chen, D.~Meng, and L.~Zhang, ``Beyond a gaussian denoiser: Residual learning of deep cnn for image denoising,'' \emph{IEEE Transactions on Image Processing}, vol.~26, no.~7, pp. 3142--3155, 2017.

\bibitem{ref46}
X.~Chen and K.~He, ``Exploring simple siamese representation learning,'' in \emph{Proceedings of the IEEE/CVF Conference on Computer Vision and Pattern Recognition}, 2021, pp. 15\,750--15\,758.

\bibitem{ref47}
W.~Yang, S.~Wang, Y.~Fang, Y.~Wang, and J.~Liu, ``From fidelity to perceptual quality: A semi-supervised approach for low-light image enhancement,'' in \emph{Proceedings of the IEEE/CVF Conference on Computer Vision and Pattern Recognition}, 2020, pp. 3063--3072.

\bibitem{ref48}
X.~Liu, Z.~Wu, A.~Li, F.-A. Vasluianu, Y.~Zhang, S.~Gu, L.~Zhang, C.~Zhu, and R.~Timofte, ``Ntire 2024 challenge on low light image enhancement: Methods and results,'' in \emph{Proceedings of the IEEE/CVF Conference on Computer Vision and Pattern Recognition Workshops}, 2024.

\bibitem{ref51}
W.~Yang, Y.~Yuan, W.~Ren, J.~Liu, W.~J. Scheirer, Z.~Wang, T.~Zhang, Q.~Zhong, D.~Xie, S.~Pu \emph{et~al.}, ``Advancing image understanding in poor visibility environments: A collective benchmark study,'' \emph{IEEE Transactions on Image Processing}, vol.~29, pp. 5737--5752, 2020.

\bibitem{ref50}
R.~Zhang, P.~Isola, A.~A. Efros, E.~Shechtman, and O.~Wang, ``The unreasonable effectiveness of deep features as a perceptual metric,'' in \emph{Proceedings of the IEEE conference on Computer Vision and Pattern Recognition}, 2018, pp. 586--595.

\bibitem{ref49}
S.~Wang, J.~Zheng, H.-M. Hu, and B.~Li, ``Naturalness preserved enhancement algorithm for non-uniform illumination images,'' \emph{IEEE Transactions on Image Processing}, vol.~22, no.~9, pp. 3538--3548, 2013.

\bibitem{ref52}
Z.~Zhang, H.~Zheng, R.~Hong, M.~Xu, S.~Yan, and M.~Wang, ``Deep color consistent network for low-light image enhancement,'' in \emph{Proceedings of the IEEE/CVF Conference on Computer Vision and Pattern Recognition}, 2022, pp. 1899--1908.

\bibitem{chen2015efficient}
G.~Chen, F.~Zhu, and P.~Ann~Heng, ``An efficient statistical method for image noise level estimation,'' in \emph{Proceedings of the IEEE International Conference on Computer Vision}, 2015, pp. 477--485.

\bibitem{ref57}
V.~Bychkovsky, S.~Paris, E.~Chan, and F.~Durand, ``Learning photographic global tonal adjustment with a database of input/output image pairs,'' in \emph{CVPR 2011}.\hskip 1em plus 0.5em minus 0.4em\relax IEEE, 2011, pp. 97--104.

\bibitem{ref55}
H.~Jiang, X.~Zhu, Y.~Ren, Y.~Hao, F.~Zou, F.~Lin, and S.~Han, ``R2rnet: Low-light image enhancement via real-low to real-normal network,'' \emph{Journal of Visual Communication and Image Representation}, vol.~90, p. 103712, 2023.

\bibitem{ref56}
T.-Y. Lin, M.~Maire, S.~Belongie, J.~Hays, P.~Perona, D.~Ramanan, P.~Doll{\'a}r, and C.~L. Zitnick, ``Microsoft coco: Common objects in context,'' in \emph{Computer Vision--ECCV 2014: 13th European Conference, Zurich, Switzerland, September 6-12, 2014, Proceedings, Part V 13}.\hskip 1em plus 0.5em minus 0.4em\relax Springer, 2014, pp. 740--755.

\bibitem{chobola2024fast}
T.~Chobola, Y.~Liu, H.~Zhang, J.~A. Schnabel, and T.~Peng, ``Fast context-based low-light image enhancement via neural implicit representations,'' in \emph{Proceedings of the European Conference on Computer Vision (ECCV)}, 2024.

\bibitem{ref53}
J.~Deng, J.~Guo, E.~Ververas, I.~Kotsia, and S.~Zafeiriou, ``Retinaface: Single-shot multi-level face localisation in the wild,'' in \emph{Proceedings of the IEEE/CVF Conference on Computer Vision and Pattern Recognition}, 2020, pp. 5203--5212.

\bibitem{ref54}
\BIBentryALTinterwordspacing
G.~Jocher, A.~Chaurasia, and J.~Qiu, ``{Ultralytics YOLO},'' Jan. 2023. [Online]. Available: \url{https://github.com/ultralytics/ultralytics}
\BIBentrySTDinterwordspacing

\bibitem{lv2021attention}
F.~Lv, Y.~Li, and F.~Lu, ``Attention guided low-light image enhancement with a large scale low-light simulation dataset,'' \emph{International Journal of Computer Vision}, vol. 129, no.~7, pp. 2175--2193, 2021.

\bibitem{liang2022self}
J.~Liang, Y.~Xu, Y.~Quan, B.~Shi, and H.~Ji, ``Self-supervised low-light image enhancement using discrepant untrained network priors,'' \emph{IEEE Transactions on Circuits and Systems for Video Technology}, vol.~32, no.~11, pp. 7332--7345, 2022.

\bibitem{lv2024fourier}
X.~Lv, S.~Zhang, C.~Wang, Y.~Zheng, B.~Zhong, C.~Li, and L.~Nie, ``Fourier priors-guided diffusion for zero-shot joint low-light enhancement and deblurring,'' in \emph{Proceedings of the IEEE/CVF Conference on Computer Vision and Pattern Recognition}, 2024, pp. 25\,378--25\,388.

\bibitem{gu2024seed}
Y.~Gu, Y.~Jin, B.~Wang, Z.~Wei, X.~Ma, H.~Wang, P.~Ling, H.~Chen, and E.~Chen, ``Seed optimization with frozen generator for superior zero-shot low-light image enhancement,'' \emph{IEEE Transactions on Circuits and Systems for Video Technology}, 2024.

\end{thebibliography}
\end{document}